\documentclass[preprint,3p]{elsarticle}
\usepackage{booktabs}
\usepackage{graphicx}
\usepackage{subfig}
\usepackage{float}
\RequirePackage{natbib}

\usepackage{lipsum}
\makeatletter
\def\ps@pprintTitle{%
 \let\@oddhead\@empty
 \let\@evenhead\@empty
 \def\@oddfoot{}%
 \let\@evenfoot\@oddfoot}
\makeatother

\usepackage{amssymb}
\usepackage{amsthm}
\usepackage{amsbsy}
\usepackage{amsmath}
\usepackage{algorithm,algpseudocode}

\usepackage{lineno,hyperref}
\modulolinenumbers[5]

\journal{}

\RequirePackage{hyperref}
\usepackage[utf8]{inputenc} 
\begin{document}

\begin{frontmatter}
 
\title{Fully convolutional networks for structural health monitoring \\ through multivariate time series classification}

\author[1]{Luca Rosafalco}
\ead{luca.rosafalco@polimi.it}
\author[2]{Andrea Manzoni}
\ead{andrea1.manzoni@polimi.it}
\author[1]{Stefano Mariani}
\ead{stefano.mariani@polimi.it}
\author[1]{Alberto Corigliano}
\ead{alberto.corigliano@polimi.it}

\address[1]{Dipartimento di Ingegneria Civile e Ambientale,
Politecnico di Milano, P.zza Leonardo da Vinci 32, 20133, Milano, Italy}
\address[2]{MOX - Dipartimento di Matematica, Politecnico di Milano, P.zza Leonardo da Vinci 32, 20133 Milano, Italy}

\begin{abstract}
 We propose a novel approach to Structural Health Monitoring (SHM), aiming at the automatic identification of damage-sensitive features from data acquired through pervasive sensor systems. Damage detection and localization are formulated as classification problems, and tackled through Fully Convolutional Networks (FCNs). A supervised training of the proposed network architecture is performed on data extracted from numerical simulations of a physics-based  model (playing the role of digital twin of the structure to be monitored) accounting for different damage scenarios. By relying on this simplified model of the structure, several load conditions are considered during the training phase of the FCN, whose architecture has been designed to deal with time series of different length. The training of the neural network is done before the monitoring system starts operating, thus enabling a real time damage classification. The numerical performances of the proposed strategy are assessed on a numerical benchmark case consisting of an eight-story shear building subjected to two load types, one of which modeling random vibrations due to low-energy seismicity. Measurement noise has been added to the responses of the structure to mimic the outputs of a real monitoring system. Extremely good classification capacities are shown: among the nine possible alternatives (represented by the healthy state and by a damage at any floor), damage is correctly classified in up to $95 \%$ of cases, thus showing the strong potential of the proposed approach in view of the application to real-life cases.
\end{abstract}

\begin{keyword}
structural health monitoring \sep fully convolutional networks \sep damage localization \sep time series analysis \sep deep learning 
\end{keyword}

\end{frontmatter}

\section{Introduction}
\label{sec:introduction}
Collapses of civil infrastructures strike public opinion more and more often. They are generally due to either structural deterioration or modified working conditions with respect to the design ones. The main challenge of Structural Health Monitoring (SHM) is to increase the safety level of ageing structures by detecting, locating and quantifying the presence and the development of damages, possibly in real-time \cite{art:review_SHM}. However, visual inspections -- whose frequencies are usually determined by the importance and the age of the structure -- are still the workhorse in this field, even if they are rarely able to provide a quantitative estimate of structural damages. Therefore, it is evident why recent advances in sensing technologies and signal processing, coupled to the increased availability of computing power, are creating huge expectations in the development of robust and continuous SHM systems \cite{art:Azam_Mariani}. 


SHM applications are often treated as \textit{classification problems} \cite{art:Farrar_1} aiming {\em (i)} to distinguish the damage state of a structure from the undamaged state, starting from a set of available recordings of a monitoring sensor system,  and {\em (ii)} to locate and quantify the current damage. In this framework, we have adopted the so-called 
Simulation-Based Classification (SBC) approach \cite{art:Taddei_Patera}, and we have exploited 
for the first time Deep Learning (DL) techniques for the sake of automatic classification. In our procedure, data are displacement and/or acceleration recordings of the structure response, and the classification task consists of recognizing which structural state, among a discrete set, could have most probably 
produced them. These structural states, characterized by the presence of damage in different positions and of different magnitudes, suitably represent different damage scenarios.  \\ 

To highlight the distinctive components of the SBC approach, we recall the general paradigm for a SHM system, according to \cite{art:Farrar_1}. A SHM system consists of four sequential procedures: $(i)$ operational evaluation, $(ii)$ data acquisition, $(iii)$ features extraction and $(iv)$ statistical inference. Operational evaluation defines what the object of the monitoring is and what the most probable damage scenarios are; data acquisition deals instead with the implementation of the sensing system; features extraction specifies how to exploit the acquired signals to derive features, that is, a reduced representation of the initial data, yet containing all their relevant information -- for the case at hand, the onset and propagation of damage in the structure; statistical inference finally sets the criteria under which the classification task is performed.

Focusing on stages $(ii)$ and $(iii)$, the vibration-based approach is nowadays the most common procedure in civil SHM. Its popularity is mainly due to the effective idea that the ongoing damage alters the structure vibration response \cite{art:Farrar_2} and, consequently, the associated modal information. By looking at the displacement and/or the acceleration time recordings acquired at a certain set of points of a building, the vibration-based approach enables the analysis of both global and local structural behaviors. The technology required to build this type of sensor system is mature and can be exploited on massive scale \cite{book:Farrar_Worden}. In most of the cases, features extraction relies on determining the system eigenfrequencies and the modal shapes. On the other hand, it might be necessary to employ more involved outcomes to distinguish between the effect of modified loading conditions and the true effect of damage \cite{art:Farrar_3}, for instance by constructing parametric time series models \cite{art:Alireza_2}. By employing DL, we aim at dealing with these aspects automatically. \\

Two competing approaches are employed in literature to deal with stage $(iv)$, the \textit{a)} model-based and the \textit{b)} data-based approach, both introducing a sort of offline-online decomposition. By this expression, we mean the possibility to split the procedure into two phases: first, the offline phase is performed before the structure starts operating; then, the online phase is carried out during its normal operations. 

The model-based approach builds a physics-based model, initially calibrated to simulate the structure response. The model is updated whenever new observations become available and, accordingly, damage is detected and located. Data assimilation techniques such as Kalman filters have been employed to efficiently deal with model updating \cite{book:Azam}. Model-based approaches are typically ill-conditioned, and many uncertainties related to the proper tuning of model parameters may prevent from a correct damage estimation. 

Hence, data-based approaches are becoming more and more popular; they exploit a collection of structural responses and, either assess any deviation between real and simulated data, or assign to the measured data the relevant class label. The dataset construction can be done either experimentally \cite{art:waves_SHM} or numerically; however,  the latter option is usually preferred, due to the frequent difficulties in reproducing the effects of damage in real-scale civil structures properly. To reduce the computational burden associated with the dataset construction, simplified models (e.g. mass-spring models for the dynamics of tall and slender buildings)  -- still able to catch the correct structural response -- are preferred with respect to more expensive high-fidelity simulations, involving, e.g., the discretization both of structural and non-structural elements. By adopting the SBC method, we rely on a data-driven approach based on synthetic experiments. 

Once a dataset of possible damage scenarios has been constructed, Machine Learning (ML) has proved to be suitable to perform the classification task \cite{book:Farrar_Worden}. The training of the ML classifier could be:
\begin{itemize}
  \item supervised, when a label corresponding to one of the possible outputs of the classification task is associated to each structural response;
  \item unsupervised \cite{art:Alireza_1}, when no labelling is available;
  \item semi-supervised \cite{art:semi-supervised}, when the training data only consist of data referring to a reference condition.
\end{itemize}  
In the SBC framework, a semi-supervised approach was recently explored, e.g., in \cite{art:Bigoni}, leading to great computational savings and robust results when treating the anomaly detection task. 
Despite of their good performances, standard ML techniques, based, e.g., on statistical distributions of the damage classes (as in the so-called decision boundary methods), as well as kernel-based methods (e.g. support vector machines), still rely on heavy data preprocessing, required to compute problem-specific sets of engineered features \cite{proc:time_series_base}. These features can be statistics of the signal, modal properties of the structure, or even more involved measures exploiting different types of signal transformation (e.g. Power Spectral Density, autocorrelation functions, to mention a few) \cite{book:Farrar_Worden}. Some relevant drawbacks arise, since:
\begin{itemize}
	\item pre-computed engineered features are not well suited for non-standard problems, for which setting damage classification criteria can be anything but trivial;
	\item there is no way to assess the optimality of the employed features;
	\item a computationally expensive pre-processing of a huge amount of data is usually required.
\end{itemize}
For these reasons, we rely on deep learning techniques, which that allows both 
data dimensionality reduction and hierarchical pattern recognition at the same time \cite{art:Hinton, book:Bengio}. DL techniques allow us to both:
\begin{itemize}
	\item deal with non-standard problems, especially when different information sources have to be managed (as long as they are in the form of time series), and
	\item detect a set of features, optimized with respect to the classification task, through the training of an artificial neural network.
\end{itemize}
Despite these advantages, the use of DL for the sake of SHM has been quite limited so far \cite{art:SAE2, proc:DL_SHM}. For the first time in the SHM field, we have therefore decided to employ Fully Convolutional Networks (FCNs) \cite{art:multivariate_LSTM_FCN}, a particular Neural Network (NN) architecture, to deal with the Multivariate Time Series (MTS) produced by monitoring sensor systems. To face different information sources, we have applied separate convolutional branches and, at a second stage, performed the data fusion of the extracted information. \\

The remainder of the paper is organized as follows. In Sec.~\ref{sec:methodology} we introduce the proposed DL-based approach to the SHM problem. In Sec.~\ref{sec:FCN} we introduce the employed FCN architecture, detailing the NN hyperparameters tuning. In Sec.~\ref{sec:numerical_results} we describe our benchmark problem and analyse our numerical results. Finally, some concluding remarks are provided in Sec.~\ref{sec:conclusions}.

\section{SHM Methodology}
\label{sec:methodology}

We introduce in this section a detailed explanation of the proposed strategy to deal with the SHM problem exploiting a SBC approach. We exploit a simplified physics-based model of the structure employing $M$ degrees of freedom (dofs), assuming  to record time-dependent signals through a monitoring system employing $N_0 \leq M$ sensors. Our aim is first to train, and then to use, two classifiers $\mathcal{G}_{d}$ and $\mathcal{G}_{l}$ for the sake of damage detection and localization, respectively, where 
\[
\mathcal{G}_{d} :  \mathbb{R}^{N_0 \times L_0} \rightarrow \lbrace 0 , 1 \rbrace ~,  \qquad \qquad 
\mathcal{G}_{l} :  \mathbb{R}^{N_0 \times L_0} \rightarrow \lbrace 0 , 1, \ldots , G \rbrace ~.  
\]
In the former case, labels $0$ and $1$ denote absence or presence of damage, respectively; in the latter, $G>1$ is a priori fixed and denotes the range of possible damage locations -- also in this case, the undamaged state is denoted by $0$.  We have decided to include the undamaged state among the possible outputs of $\mathcal{G}_{l}$ not just to confirm the outcome of $\mathcal{G}_{d}$, but also to observe which damage scenarios, identified by their locations, are more often misclassified with the undamaged state.

The training of $\mathcal{G}_{d}$ and $\mathcal{G}_{l}$ is performed using two datasets $\mathbb{D}^{d}_{train}$ and $\mathbb{D}^{l}_{train}$, respectively. Each of these two datasets (for simplicity we only consider the formation of $\mathbb{D}^l_{train}$, being the process substantially equivalent for $\mathbb{D}^{d}_{train}$) collects $V$ structural responses,  
\begin{equation*}
\mathbb{D}^l_{train} = \lbrace \mathbb{U}_1 , \ldots , \mathbb{U}_{V^{train}} \rbrace ~, 
\end{equation*}
under prescribed damage scenarios and loading conditions. We denote by $\mathbb{U}_i \in \mathbb{R}^{N_0 \times L_0}$, $i=1,\ldots,V^{train}$, a collection of $N_0$ sensor recordings of displacement and/or acceleration time series of length $L_0$, such that 
\begin{equation}
\label{eq:d_l_train}
\mathbb{U}_i = \left[ \boldsymbol{u}_1 \left(\boldsymbol{d}_i,\boldsymbol{l}_i \right) \ | \ \ldots \  | \  \boldsymbol{u}_{N_0} \left(\boldsymbol{d}_i,\boldsymbol{l}_i \right) \right], \qquad   i=1,\ldots,V^{train}  ~; 
\end{equation}
the time series $\boldsymbol{u}_n \left(\boldsymbol{d}_i,\boldsymbol{l}_i \right)$ recorded by the $n$-th sensor depends on the  damage scenario $\boldsymbol{d}_i$ and the loading condition $\boldsymbol{l}_i$, and can be seen as the sampling of a time-dependent signal ${u}_n \left(\boldsymbol{d}_i,\boldsymbol{l}_i \right)$. We assume to deal with recordings acquired at a set of $L_0$ time instants uniformly distributed over the time interval of interest $I$.
The damage scenario $\boldsymbol{d}_i : \mathcal{P}_d \rightarrow \mathbb{R}^M$  is prescribed at each element\footnote{\footnotesize For simplicity, the number $E$ of elements coincides with the number $M$ of degrees of freedom; however, the generalization to the case in which $E \neq M$  is straightforward.} and depends on a set of parameters 
$\boldsymbol{\eta}_d \in \mathcal{P}_d \subset \mathbb{R}^{D}$; the loading condition $\boldsymbol{l}_i : I \times \mathcal{P}_l \rightarrow \mathbb{R}^M$, defined over the time interval $I$, is prescribed at each element, too, and  depends on a set of parameters $\boldsymbol{\eta}_l \in \mathcal{P}_l \subset \mathbb{R}^{L}$. Here, we denote by $\mathcal{P}_d$ and $\mathcal{P}_l$ two sets of parameters, yielding the two sets $\mathcal{C}_d$ and $\mathcal{C}_l$ of admissible damage and loading scenarios, respectively, obtained when sampling $\boldsymbol{\eta}_d \in \mathcal{P}_d$ and $\boldsymbol{\eta}_l \in \mathcal{P}_l$.
During the training procedure, the performances of $\mathcal{G}_d$ and $\mathcal{G}_l$ are tracked by looking at their classification capabilities on two datasets $\mathbb{D}^d_{val}$ and $\mathbb{D}^l_{val}$, each one collecting $V^{val}$ structural responses $\mathbb{U}_i$ (defined as in Eq.~\eqref{eq:d_l_train}), $i=1,\ldots,V^{val}$. 

According to the SBC approach, the datasets $\mathbb{D}^d_{train}$, $\mathbb{D}^l_{train}$, $\mathbb{D}^d_{val}$ and $\mathbb{D}^l_{val}$ are constructed by exploiting a simplified physics-based model of the structure that employs $M$ dofs, and by recording its displacements and/or accelerations by using a virtual monitoring system that uses $N_0$ sensors. For any damage scenario $\boldsymbol{d} \in \mathcal{C}_d$ and loading conditions  $\boldsymbol{l} \in \mathcal{C}_l$ received as inputs, this numerical model -- playing the role of digital twin of the structure to be monitored -- returns a recorded displacement and/or acceleration time series $\boldsymbol{r}_n \left(\boldsymbol{d},\boldsymbol{l} \right)$. Since these latter are deterministic, to make our data more conformal to real measurements $\boldsymbol{u}_n \left(\boldsymbol{d},\boldsymbol{l} \right)$, we assume that each $\boldsymbol{r}_n \left(\boldsymbol{d},\boldsymbol{l} \right)$ is affected by an additive measurement noise $\boldsymbol{\epsilon}_n\sim \mathcal{N} \left( {\bf 0}, \boldsymbol{\Sigma}_{\epsilon} \right)$, so that 
\begin{equation}
\boldsymbol{u}_n = \boldsymbol{r}_n \left(\boldsymbol{d},\boldsymbol{l} \right) + \boldsymbol{\epsilon}_n, \qquad n=1,\ldots,N_0 ~. 
\label{eq:add_noise}
\end{equation}
Here we consider each $\boldsymbol{\epsilon}_n $ normally distributed, with zero mean and covariance matrix $\boldsymbol{\Sigma}_{\epsilon} \in \mathbb{R}^{N_0 \times N_0}$, related to a real monitoring system \cite{art:Capellari_1}. Regarding the auto-correlation of the records ($j=1,\ldots, L_0$) of each sensor ($n=1,\ldots, N_0$) in time, we assume them to be independent and identically distributed. \\  

The classifiers $\mathcal{G}_{d}$ and $\mathcal{G}_{l}$ are based on a fully convolutional neural network architecture (that will be detailed in the following section). The training of the network is supervised, and performed by feeding the FCN with multivariate time series $\lbrace \mathcal{F}^{n}_0 \rbrace_{n=1}^{N_0}$ and associated labels ($0$ or $1$ for $\mathcal{G}_{d}$, $g \in \{0,1,\ldots, G\}$ for $\mathcal{G}_{l}$). In this respect, hereon each multivariate time series $\lbrace \mathcal{F}^{n}_0 \rbrace_{n=1}^{N_0}$ is referred to as instance. In general, $\lbrace \mathcal{F}^{n}_0 \rbrace_{n=1}^{N_0} = \mathbb{U}_i$, however a single instance might be made up to $W$ multivariate time series $\mathbb{U}_{iw}$, $w=1,2,\ldots,W$ of different lengths $L_0^w$ to deal with the case of sensors recording time series of different length. Each component $\mathcal{F}^{n}_0 = \boldsymbol{u}_{n}$ plays the role of input channel for the NN.

The testing of the NN is done on instances $\lbrace \mathcal{F}^{n}_{*} \rbrace_{n=1}^{N_0} = \mathbb{U}^*_i$, obtained through the numerical model as structural response
\begin{equation*}
\mathbb{U}^*_i = \left[ \boldsymbol{u}_1 \left(\boldsymbol{d}_i,\boldsymbol{l}_i^* \right) \ | \ \ldots \  | \  \boldsymbol{u}_{N_0} \left(\boldsymbol{d}_i,\boldsymbol{l}_i^* \right) \right] , \qquad i=1, \ldots, V^{test}
\end{equation*}
to loading conditions $\boldsymbol{l}^*_i \in \mathcal{C}_l$, $i=1,\ldots, V^{test}$, unseen (that is, associated to testing values $\boldsymbol{\eta}_l$ from $\mathcal{P}_l$ not sampled) when building the datasets $\mathbb{D}^d_{train}$, $\mathbb{D}^l_{train}$, $\mathbb{D}^d_{val}$ and $\mathbb{D}^l_{val}$. All these instances are collected into two datasets $\mathbb{D}^d_{test}$ and $\mathbb{D}^l_{test}$.

The testing is done by verifying the correct identification of the class ($\{0,1\}$ for $\mathcal{G}_{d}$, $\{0,1,\ldots, G\}$ for $\mathcal{G}_{l}$) associated with the simulated signals. In concrete terms, a probability is estimated for each possible class, thus yielding the confidence level that the given class is assigned to the data, and the class with highest confidence is compared with the one associated to the simulated signal. No k-fold cross validation is used.

Once tested, $\mathcal{G}_{d}$ and $\mathcal{G}_{l}$ can make a prediction once a new signal $\lbrace \mathcal{F}^{n}_{*} \rbrace_{n=1}^{N_0} = \mathbb{U}^{*}$ has been experimentally acquired from the real sensor network used to monitor the structure. \\

\begin{figure}[t!]
\centerline{
\includegraphics[width=0.725\textwidth]{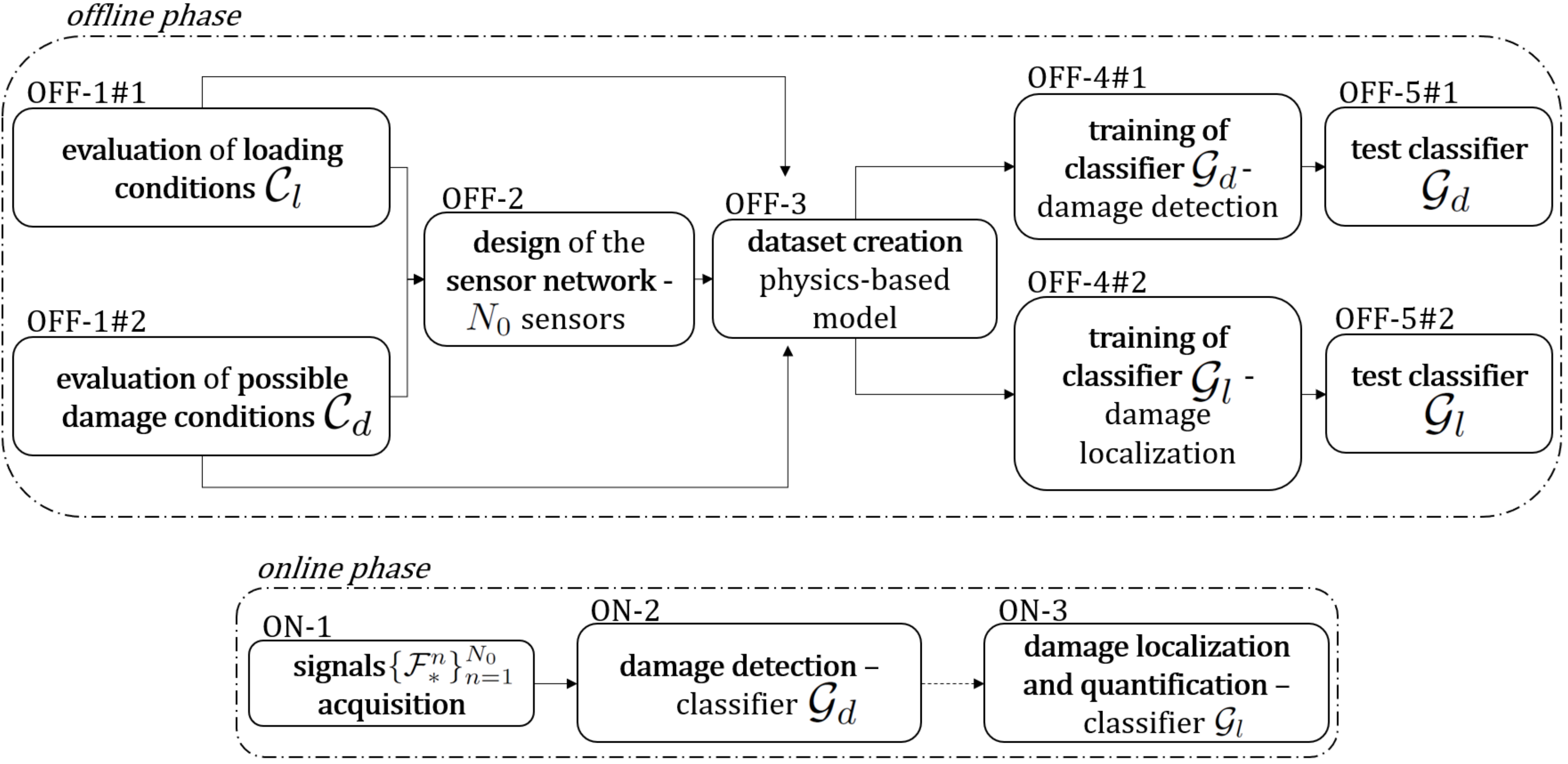}
}
\caption{SBC + FCN classifier. 
The offline phase is performed before the start of operations of the structure, while the online stage during its normal operations.\label{fig:methodology_graph}}
\end{figure}

Let us now recap the procedure steps exploiting the schematic representation reported in Fig.~\ref{fig:methodology_graph}.

\newpage

For the sake of convenience, we can split our procedure into:
\begin{itemize}
\item an offline phase, where, as first step, the loading conditions $\mathcal{C}_l $ (OFF-1\#1) and the most probable damage scenarios $\mathcal{C}_d $ are evaluated (OFF-1\#2).
Accordingly, a sensor network with $N_0$ sensors is designed (OFF-2). The datasets $\mathbb{D}^d_{train}$, $\mathbb{D}^l_{train}$, $\mathbb{D}^d_{val}$, $\mathbb{D}^l_{val}$, $\mathbb{D}^d_{test}$ and $\mathbb{D}^l_{test}$ are then constructed (OFF-3) by exploiting the physics-based {\em digital twin} of the structure. The classifiers $\mathcal{G}_{d}$ (OFF-4\#1) and $\mathcal{G}_{l}$ (OFF-4\#2) are therefore trained by using $\mathbb{D}^d_{train}$ and $\mathbb{D}^l_{train}$ and performing the validation using $\mathbb{D}^d_{val}$ and $\mathbb{D}^l_{val}$. Finally, the classification capacity of $\mathcal{G}_{d}$ and $\mathcal{G}_{l}$ is assessed by using numerically simulated signals $\lbrace \mathcal{F}^{n}_{*} \rbrace_{n=1}^{N_0} = \mathbb{U}^*$ belonging to $\mathbb{D}^d_{test}$ and $\mathbb{D}^l_{test}$, respectively (OFF-5\#1 and OFF-5\#2);

\item an online phase, in which for any new signal $\lbrace \mathcal{F}^{n}_{*} \rbrace_{n=1}^{N_0} = \mathbb{U}^*$ acquired by the real monitoring system and provided to the classifiers (ON-1), damage detection (ON-2) is performed through $\mathcal{G}_{d}$, and damage localization is performed through $\mathcal{G}_{l}$ (ON-3).
\end{itemize}

In lack of recordings coming from a real monitoring system, and having assumed the experimental signals $\mathbb{U}^*$ equal to the noise-corrupted output of the numerical model, steps OFF-5\#1 and OFF-5\#2 of the offline phase indeed coincide with steps ON-2 and ON-3 of the online procedure\footnote{\footnotesize For this reason, the acquired signals are denoted with the same notation $\mathbb{U}^*$ employed for the recordings previously used to test the FCN to highlight that, for the time being, the experimental signals are taken as realizations of the noise-corrupted outputs of the numerical model.}. Moreover, we highlight that only those damage scenarios $\boldsymbol{d} \in \mathcal{C}_d$ that have been numerically simulated in the offline phase can be classified during the online phase. Moreover, damage is considered temporary frozen within a fixed observation interval, enabling to treat the structure as linear \cite{art:Azam_Mariani}. To model the effect of damage, we consider the  stiffness degradation of each structural member; this assumption is acceptable if the rate of the evolving damage is sufficiently small with respect to the observation interval. 

It is not possible to identify from the beginning the most suitable number of instances $V^{train}$ to be used to train the network. The easiest procedure (even if time-consuming) would be to assess the performances of $\mathcal{G}_{d}$ and $\mathcal{G}_{l}$ for different sizes $V^{train}$, aiming at finding a trade-off between the computational burden required to construct the dataset and train the NN, and the classification capabilities. Beyond a certain critical size, massive dataset enlargements might lead to small improvements in the NN performance, as shown in our numerical results. 

Finally, concerning the setting of the loading conditions $\mathcal{C}_l$, in this work we have (i) identified a set of possible loading scenarios that can significantly affect the response of the structure; (ii) subdivided this set into a certain number of subsets, representative of different possible dynamic effects of the applied load; (iii) sampled each subset almost the same number of times.

\section{Fully Convolutional Networks} 
\label{sec:FCN}

\subsection{Neural Network architecture}
\label{sec:NN_architecture}

We now describe the FCN architecture employed for the sake of classification. As discussed in the previous section, $\lbrace \mathcal{F}^{n}_0 \rbrace_{n=1}^{N_0}$ are the inputs adopted during the training phase (for which we know the instance label associated), while $\lbrace \mathcal{F}^{n}_{*} \rbrace_{n=1}^{N_0}$ are the inputs that we require the FCN to classify. 

\begin{figure}[h!]
 \begin{center}
\centerline{
\includegraphics[width=0.6\textwidth]{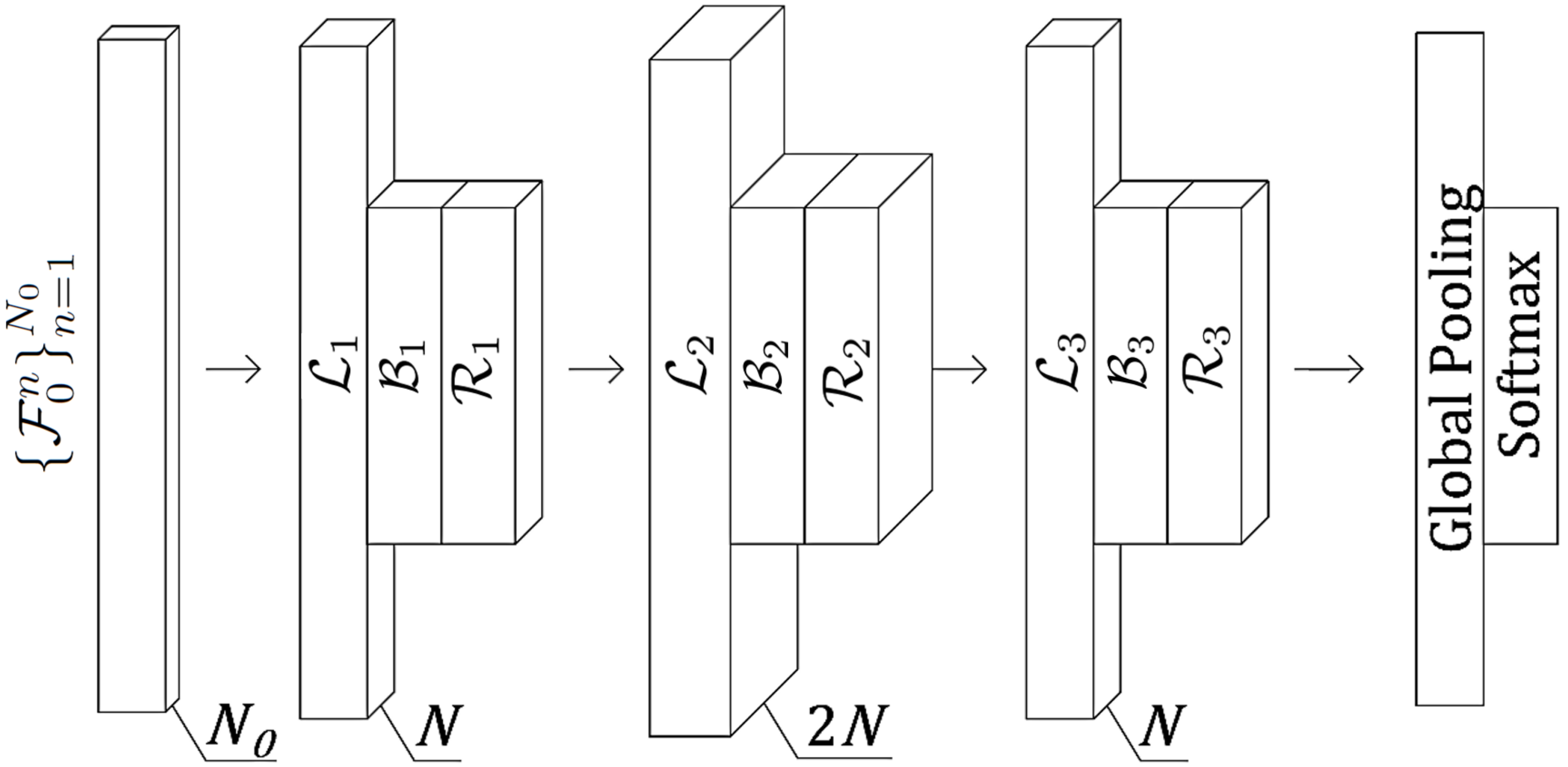}
}
\caption{FCN architecture in the case of a single data type. Here $N_0$ represents the number of input channels and $N$ represents the adopted number of filters. For sake of clarity, the dimensionality of the building blocks has been enhanced: a three-dimensional parallelepipeds is used to depict the two-dimensional output of each convolutional layer; a two-dimensional rectangle is used to depict the one-dimensional output of the global pooling layer and of the softmax layer.\label{fig:FCN_architecture}}
 \end{center}
\end{figure}

We have adopted a FCN stacking three convolutional layers $\mathcal{L}_i$, $i=\lbrace 1,2,3 \rbrace$, with different filter sizes $h_i$, followed by a global pooling layer and a softmax classifier (the choice of the NN hyperparameters will be discussed in the following). Each convolutional layer $\mathcal{L}_i$ has been used together with a Batch-Normalization (BN) layer $\mathcal{B}_i$ and a Rectified Linear Unit (ReLU) activation layer $\mathcal{R}_i$ \cite{proc:time_series_base, art:multivariate_LSTM_FCN}, see Fig.~\ref{fig:FCN_architecture}. 

\begin{figure}[h!]
 \begin{center}
\centerline{
\includegraphics[width=0.6\textwidth]{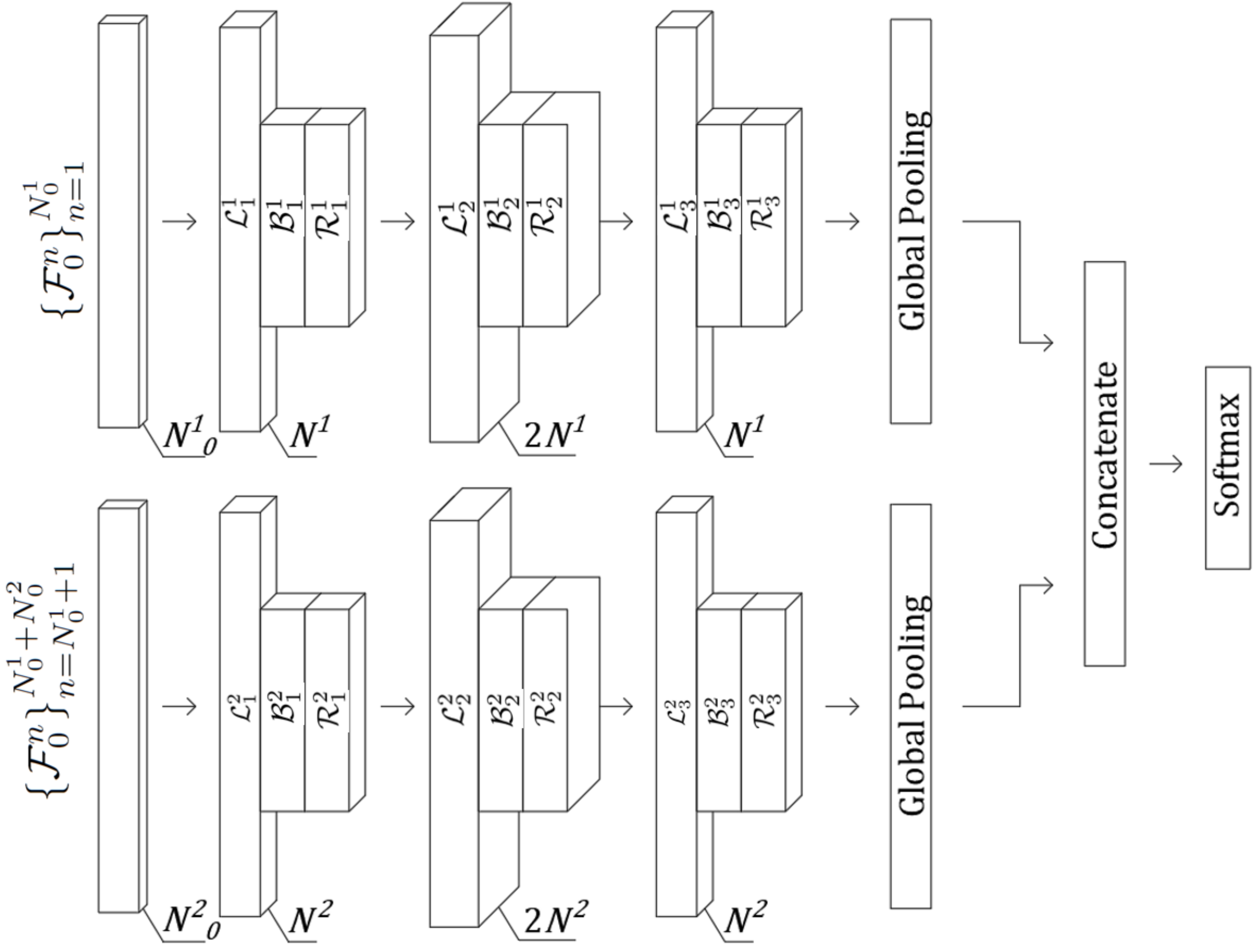}
}
\caption{FCN architecture in the case of two data types. Here $N^1_0$ and $N^2_0$ represent the number of input channels (possibly different) of the two NN branches; $N^1$ and $N^2$ represent the number of filters adopted. For sake of clarity, the dimensionality of the building blocks has been enhanced: a three-dimensional parallelepipeds is used to depict the two-dimensional output of each convolutional layer; a two-dimensional rectangle is used to depict the one-dimensional output of the global pooling layer and of the softmax layer.\label{fig:FCN_architecture_bis}}
\vspace{-0.785cm}
 \end{center}
\end{figure}

When the input signals are made up by $W$ multivariate time series with different length:
\begin{equation*}
  \lbrace \mathcal{F}^{n}_0 \rbrace_{n=1}^{N^1_0 + ...+  N^i_0+... + N^W_0} =
    \begin{cases}
      \lbrace \mathcal{F}^{n}_0 \rbrace_{n=1}^{N^1_0} = \mathbb{U}_{i1} \in \mathbb{R}^{N^1_0 \times L_0^1} \\
      \lbrace \mathcal{F}^{n}_0 \rbrace_{n=N^1_0 +1 }^{N^1_0 + N^2_0} = \mathbb{U}_{i2} \in \mathbb{R}^{N^2_0 \times L_0^2} \\
      \vdots \\
      \lbrace \mathcal{F}^{n}_0 \rbrace_{n=N^1_0 + ... + N^{W-1}_0 +1 }^{N^1_0 + ... + N^W_0} = \mathbb{U}_{iW} \in \mathbb{R}^{N^W_0 \times L_0^W}
    \end{cases} ~,
\end{equation*}
for each one we first adopt the described convolutional architecture separately and then, through a concatenation layer, we perform data fusion on the extracted features.  
Classification is finally pursued through a softmax layer. The corresponding NN architecture is sketched in Fig.~\ref{fig:FCN_architecture_bis} in the case of time series with two different lengths $L_0^1$ and $L_0^2$, but can be easily generalised. Tensorflow \cite{code:tensorflow} has been used for the sake of NN construction.

\subsection{Use of convolutional layers}
\label{sec:convolutional_layers}

Let us now show how convolutional layers can be adopted to extract features from multivariate time series. $\lbrace \mathcal{F}^n_0 \rbrace_{n=1}^{N_0}$ are provided to the $1$-st convolutional layer $\mathcal{L}_{1}$. The output of $\mathcal{L}_1$, $\lbrace \mathcal{F}^{n}_{1} \rbrace_{n=1}^{N_1}$, still shaped as time series (of length $L_1$), do not represent displacement and/or acceleration any more. Indeed, they are features extracted from the input channels $\lbrace \mathcal{F}^{n}_{0} \rbrace_{n=1}^{N_0}$. The following layers operate in the same manner: the outputs $\lbrace \mathcal{F}^{n}_{i} \rbrace_{n=1}^{N_i}$ of the $(i-1)$-th convolutional layer $\mathcal{L}_{i-1}$ are the inputs of the $i$-th convolutional layer $\mathcal{L}_{i}$ and become features of higher and higher level.

\begin{figure}
\centerline{
\includegraphics[width=0.615\textwidth]{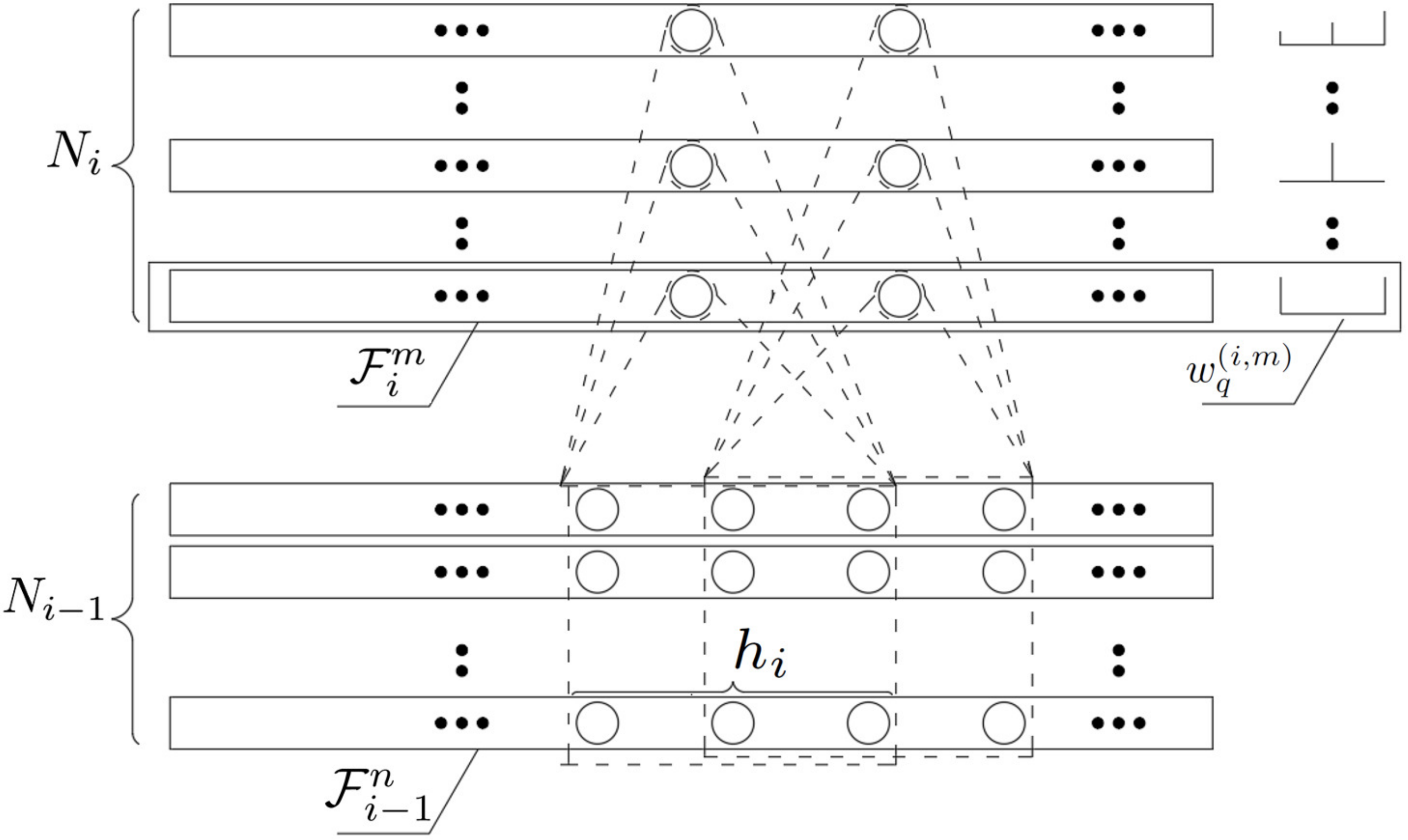}
}
\caption{Sketch of a 1D convolutional layer. Here $h_i$ specifies the kernel dimension. As a filter is associated to each feature map $n$, to represent it bars of different heights are used in relation to the amplitude of the filter weights.}\label{fig:single_conv_layer_1d}
\end{figure}

In concrete terms, the tasks performed by the $i$-th convolutional layer $\mathcal{L}_{i}$ are: the subdivision of the inputs $\lbrace \mathcal{F}^{n}_{i-1} \rbrace_{n= 1}^{N_{ i -1 }}$ into data sequences, whose length $h_i$ determines the receptive field of $\mathcal{L}_{i}$; and the multiplication of each data sequence
by a set of weights $\boldsymbol{w}^{\left(i,m \right)}$ called filter, where the output $\mathcal{F}^{n}_{i}$ of each filter is called feature map. Mono-dimensional (1D) receptive field must be used in time series analysis, being each channel monodimensional. In Fig.~\ref{fig:single_conv_layer_1d} the fundamental architecture of $\mathcal{L}_i$ is depicted, linking the inputs $\lbrace \mathcal{F}^{n}_{i-1} \rbrace_{n=1}^{N_{ i -1 }}$ and the outputs $\lbrace \mathcal{F}^{m}_i \rbrace_{m=1}^{N_i}$ through:

\begin{equation}
\label{eq:single_conv_layer_1d}
z^{\left(i,m \right)}_{h} = \sum_{q=0}^{h_i-1}\sum_{n=1}^{N_{i-1}} w^{\left(i,m \right)}_{q} x^{\left(i-1,n \right)}_{p} + b^{\left(i,m \right)} \qquad \text{with} \quad p=h + q ~,
\end{equation}
where:
\begin{itemize}

\item $z^{\left(i,m \right)}_{h}$ is the $h$-th entry of $\mathcal{F}^{m}_{i}$;

\item $ b^{\left(i,m \right)}$ is the bias of $\mathcal{F}^{m}_{i}$;

\item $x^{\left(i-1, n\right)}_{p}$ is the $p$-th entry of $\mathcal{F}^{n}_{i-1}$;

\item $w^{\left(i-1,n \right)}_{q}$ is the $q$-th connection weight of the $m$-th filter applied to the $p$-th entry of $\mathcal{F}^{n}_{i-1}$.

\end{itemize}

As the goal of stacking several convolutional layers is to provide nonlinear transformations of $\lbrace \mathcal{F}_0^n \rbrace_{n=1}^{N_0}$, their overall effect is to make the classes to be recognised linearly separable  \cite{book:Haykin}. In this way, a linear classifier is suitable to carry out the final task. Every nonlinear transformation can be interpreted, as discussed, as an automatic extraction of features.

\subsection{Batch Normalization, ReLU activation, Global Pooling and Softmax classifier}
\label{sec:technicalities}

The Batch Normalization (BN) layer $\mathcal{B}_i$ is introduced after each convolutional layer $\mathcal{L}_i$ to address the issue related to the vanishing/exploding gradients possibly experienced during the training of deep architectures \cite{proc:BN}. It relies on normalization and zero-centering of the outputs $\lbrace \mathcal{F}^{n}_{i} \rbrace_{n=1}^{N_i}$ of each layer $\mathcal{L}_{i}$. We express the output of $\mathcal{B}_i$ as $\lbrace \mathcal{F}^{n}_{\mathcal{B} i} \rbrace_{n=1}^{N_i}$. For the same reason, the ReLU activation function is preferred instead of saturating ones \cite{proc:Glorot_Bengio}. The ReLU layer $\mathcal{R}_i$ transforms $\lbrace \mathcal{F}^{n}_{\mathcal{B} i} \rbrace_{n=1}^{N_i}$, through
\begin{equation*}
\mathcal{F}^{n}_{\mathcal{R} i} \left( u \right) = \text{max} \left( 0,\mathcal{F}^{n}_{\mathcal{B} i} \left( u \right) \right) \qquad \text{with} \quad u=1,\ldots,L_i ~.
\end{equation*}
where:

\begin{itemize}
\item $\mathcal{F}^{n}_{\mathcal{B} i} \left(u \right)$ is the $u$-th entry of the $n$-th feature map of $\mathcal{B}_i$;
\item $\mathcal{F}^{n}_{\mathcal{R} i} \left(u \right)$ is the $u$-th entry of the $n$-th feature map of $\mathcal{R}_i$.
\end{itemize}

In the adopted FCN architecture, the features to be used in the classification task are extracted from $\lbrace \mathcal{F}_0^n \rbrace_{n=1}^{N_0}$ by the blocks $\lbrace \mathcal{L}_i + \mathcal{B}_i + \mathcal{R}_i \rbrace_{i=1}^3$. The final number of features equals the number $N_3$ of filters of the last convolutional layer. By applying next a global average pooling \cite{proc:glob_pooling}, the extracted features $\lbrace \mathcal{F}^{n}_{\mathcal{R} 3} \rbrace_{n=1}^{N_3}$ are condensed in a single channel $\boldsymbol{b}\in \mathbb{R}^G$, being $G$ the total number of classes.

The softmax activation layer finally performs the classification task. First, the channel $\boldsymbol{b}$ is mapped onto the target classes, by computing a score $s_g$ 
\begin{equation}
\label{eq:softmax_score}
s_g(\boldsymbol{b})=\boldsymbol{\theta}^T_g \cdot \boldsymbol{b}, \qquad g=1,\ldots,G~,
\end{equation}
for each class $g$, where the vector $\boldsymbol{\theta}_g\in \mathbb{R}^G$ collects the weights related to the $g$-th class.
The softmax function is then used to estimate the probability $p_g\in \left[0,1\right]$ that the input channels belongs to the $g$-th class, according to:
\begin{equation}
\label{eq:softmax_function}
p_g=\frac{e^{s_g(\boldsymbol{b})}}{\sum_{j=1}^G e^{s_j(\boldsymbol{b})}} \quad g=1,\ldots, G~.
\end{equation}
The input channels $\lbrace \mathcal{F}^{n}_0 \rbrace_{n=1}^{N_0}$ are then assigned to the class with associated label $g$ featuring the highest estimated probability $p_g$, which then represents the estimated confidence level that class $g$ is assigned to the data.

\subsection{Neural Network training}
\label{sec:network_training}

The NN training consists of tuning the weights $\boldsymbol{w}^{\left(i,n \right)}$ and $\boldsymbol{\theta}_g$, respectively appearing in Eqs.~\eqref{eq:single_conv_layer_1d} and \eqref{eq:softmax_score} by minimizing a loss function depending on the data. In this respect, the Adam optimization method \cite{art:Adam}, a widespread stochastic gradient-based optimization method, has been used. For classification purposes, the most commonly adopted loss function is the cross entropy, defined for the classifier $\mathcal{G}_d$ as: 	\vspace{-0.15cm}
\begin{equation}
\label{eq:cross_entropy}
J_d\left(\boldsymbol{Y},\boldsymbol{p}\right)=-\frac{1}{V^{train}}\sum^{V^{train}}_{i=1}\sum^G_{g=1}y_{i}^g \text{log}\left(p^g\right)~, \vspace{-0.15cm}
\end{equation}
where:
\begin{itemize}
\item $g$ is the label of the instance provided to the NN during the traning;
\item  $y_{i}^g\in \lbrace 0,1 \rbrace$ is the confidence that the $i$-th instance should be labelled as the $g$-th class, with \vspace{-0.15cm}
\[
 y_{i}^g= \left\{
 \begin{array}{ll}
        1  & \text{if for the }i\text{-th instance the }g\text{-th class is the target class}\\
        0 & \text{otherwise}
      \end{array} \right. \vspace{-0.15cm}
\]
\item  $\boldsymbol{Y} \in \lbrace 0, 1 \rbrace^{V^{train}}$ collects all the $y_{i}^g$ confidence values;
\item $\boldsymbol{p}\in \mathbb{R}^{G}$ collects the estimated probabilities $p^g$, see Eq.~\eqref{eq:softmax_function}.
\end{itemize}
The loss function $J_l\left(\boldsymbol{Y},\boldsymbol{p}\right)$ for the classifier $\mathcal{G}_l$ is defined analogously.

Regarding the employed datasets:
\begin{itemize}
\item $\mathbb{D}^d_{train}$ is used to train the NN by back-propagating the classification error;
\item $\mathbb{D}^d_{val}$ is used to possibly interrupt the training in case of overfitting, but not to modify the NN weights;
\item $\mathbb{D}^d_{test}$ is used to verify the prediction capabilities of the NN, after the training phase has been performed.
\end{itemize}

The same splitting applies to the data used for training $\mathcal{G}_{l}$. In order to assess the offline phase of the proposed procedure, we have tested $\mathcal{G}_{d}$ and $\mathcal{G}_{l}$ on their respective test sets $\mathbb{D}_{test}^{d}$ and $\mathbb{D}_{test}^l$ (steps OFF5\#1 and OFF5\#2 of Fig. \ref{fig:methodology_graph}). The number of times $\mathbb{D}^d_{train}$ and $\mathbb{D}^l_{train}$ are evaluated during the training of $\mathcal{G}_d$ and $\mathcal{G}_l$ corresponds to the number of epochs: in this work, we have bounded to $1500$ the maximum number of epochs allowed. We have also provided the possibility of an early-stop of the training when, after having performed at least $750$ epochs, the validation loss has not decreased three times in a row.

To control the training process, a learning rate $\xi$ is usually introduced to scale the correction of the NN weights provided by back-propagating the classification error. In out case, the learning rate has been forced to decrease linearly with the number of epochs, moving from $10^{-3}$ at the beginning of the training till $10^{-4}$ at its end \cite{art:insight_LSTM}. After having performed at least $750$ epochs, an additional factor $\zeta=1/\sqrt[3]{2}$ is used to scale down the learning rate if the loss function $J\left(\boldsymbol{Y},\boldsymbol{p}\right)$ is not reduced within the successive $100$ epochs, as suggested in \cite{art:insight_LSTM}. Random subsamples (also called minibatches) of the data points belonging to the training set are employed for the sake of gradient evaluation when running the Adam optimization method  \cite{art:Adam, book:Haykin}.

\subsection{Hyperparameters setting}
\label{sec:hyperparameters}
The setting of the NN hyperparameters, namely the dimensions of the kernels $h_i$ and the number of feature maps $N_i$, is done according to \cite{proc:time_series_base, art:insight_LSTM}. In this work, we choose $h_1=8$, $h_2=5$, $h_3=3$ as kernel dimensions for the three convolutional layers. Since no zero-padding has been employed, the dimension of the time series is progressively reduced passing through the convolutional layer $\mathcal{L}_i$ from $L_{i-1}$ to $L_{i} = L_{i-1}-h_i+1$. Accordingly, considering the parameters and the length of the time series used in this work, the dimension reduction related to a single convolutional layer is on the order of $1\%$. We have verified that the classification accuracy is barely affected by this reduction and, more in general, by the use or not use of the zero-padding. It is possible to further improve the NN performances by operating a (necessarily problem-dependent) finer tuning of the NN hyperparameters, but only at the cost of a time-consuming repeated evaluation of the NN outcomes.

The number of filters $N_i$ to be adopted depends on the complexity of the classification task: the more complex the classification, the higher the number of filters needed. However, increasing the number of filters beyond a certain threshold, which depends on the problem complexity and the task to be performed, has no effect on the prediction capabilities of the NN; indeed, the risk would be to increase computational costs, and to overfit the training dataset. Therefore, it looks convenient to initially employ a small number of filters, and then increase it if the NN performs poorly during the training phase. A possible choice suggested in \cite{proc:time_series_base} is to consider $N_1 = 128$, $N_2 = 256$, and $N_3 = 128$ as a suitable choice independently of the dataset to be analysed. Here we have kept the proportion $N_1 = N$, $N_2 = 2 N$, and $N_3 = N$ as filter sequence, and verified that increasing $N$ beyond $N=16$ does not affect the NN performances. To carry out the comparison of FCN architectures with one or two convolutional branches, we have kept $N=16$ independently of the classification task.


\section{Numerical results}
\label{sec:numerical_results}

\subsection{Dataset construction}

\begin{figure}[h!]
 \centerline{
 \includegraphics[scale=0.3]{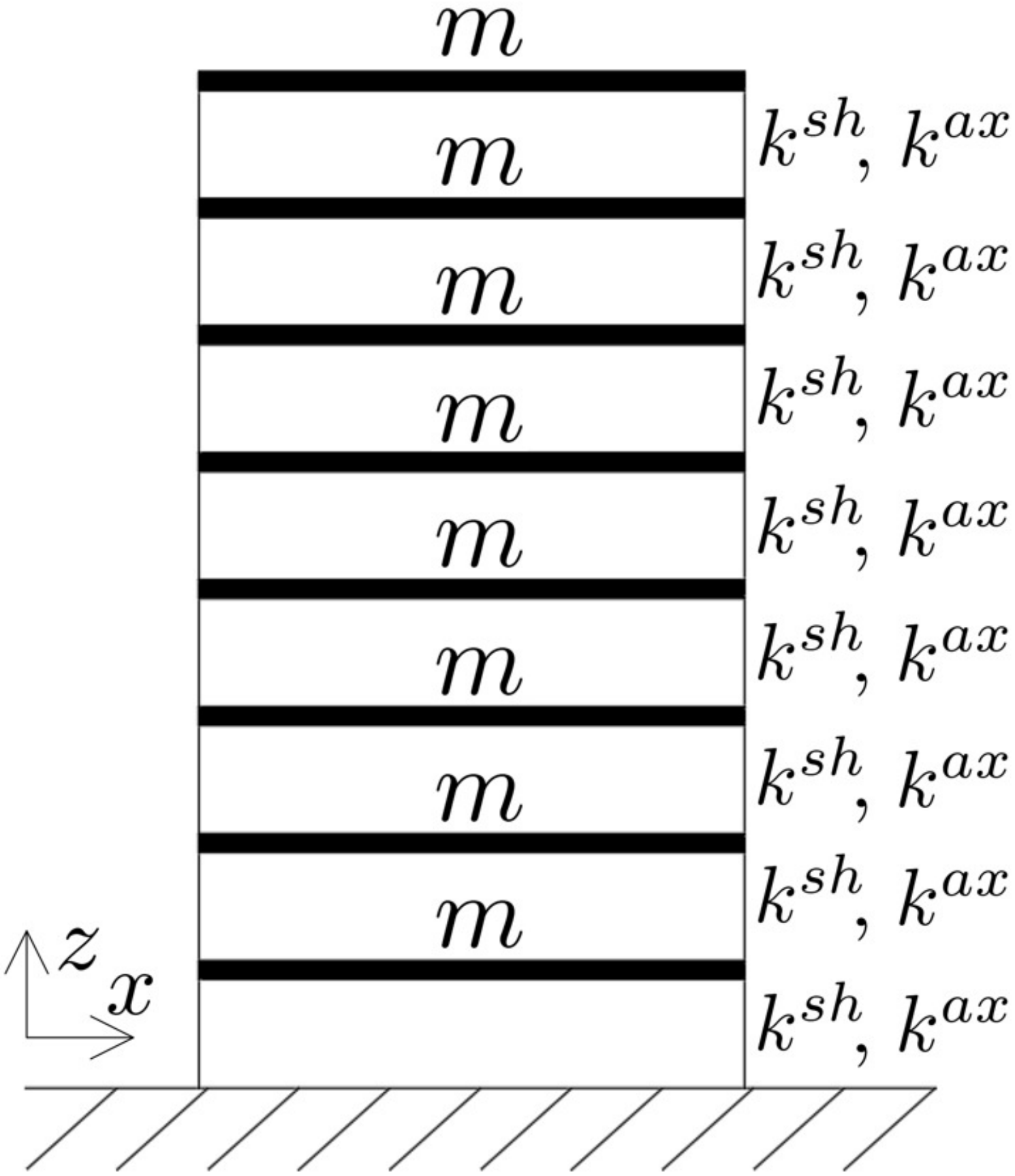}
 }
 \caption{Linear elastic shear model of an eight-story building with constant story mass and constant story stiffness\label{fig:shear_model}.}
\end{figure}

The proposed methodology is now assessed through the numerical benchmark shown in Fig.~\ref{fig:shear_model}, and originally proposed in \cite{art:Moaveni_benchmark}. 
The considered structure is an idealised eight-story shear building model, featuring a constant floor mass of $m = 625~\text{t}$ and a constant inter-story stiffness of $k^{sh} = 10^6 \text{kN/m}$. The proposed SHM strategy has been designed to handle signals related to different types of damage-sensitive structural responses characterized by different magnitude and sampling rate. Hence, in the following both the horizontal and the vertical motions of each story are allowed for and recorded. The longitudinal stiffness of the columns has been set to  $k^{ax} = 10^8 \text{kN/m}$, and a slenderness (given by the ratio their length and thickness) of $10$ has been assumed for the same columns. The numerical model employs $M=16$ dofs ($8$ in the $x$ direction and $8$ in the $z$ direction), and $N_0 = 16$ virtual sensors are used to measure the noise-free displacements $\boldsymbol{r}_n$ (collecting both horizontal displacements $\boldsymbol{r}_n$, $n=1,\ldots,8$, and vertical displacements $\boldsymbol{r}_n$, $n=9,\ldots,16$ the vertical displacements). 
 Although a non-classical damping was originally proposed in \cite{art:Moaveni_benchmark}, the relevant effect on system identification or model update has shown to be marginal if the structure is continuously excited during the monitoring stage, see e.g. \cite{art:Azam_Mariani, art:Corigliano_Mariani}. Therefore in this feasibility study, no damping has been taken into account. The dofs are numbered from $1$ for the ground floor up to $8$ for the eighth floor in both directions.

Due to the building geometry, eight different damage scenarios $\boldsymbol{d}(1), \ldots, \boldsymbol{d}(8)$ can be considered, each one characterized by a reduction of $25\%$ of one inter-story stiffness only, that is,
\begin{equation*}
\boldsymbol{d}\left( g \right) =
\left\{
\begin{array}{ll}
0.75 k_j & \quad \mbox{if } j= g \ \mbox{or } j = g +8 \\
k_j & \quad \mbox{otherwise}
\end{array}
\right.
\end{equation*}
where 
\begin{equation*}
k_j =
\left\{
\begin{array}{ll}
k^{sh}  & \quad \mbox{if } j= 1, \ldots, 8  \\
k^{ax}  & \quad \mbox{if } j= 9, \ldots, 16.  \\
\end{array}
\right.
\end{equation*}

The label $g$ is used to denote each damage scenario, ranging from $1$ for the first floor up to $8$ for the eighth floor; by convention, $\boldsymbol{d}(0)$ refers to the undamaged case. 
Before assessing the classification capability of the NN, a parametric analysis has been carried out to check the sensitivity to damage of the vibration frequencies. Tab.~\ref{tab:eigen} collects the results regarding the horizontal motion; for the analysed system, the axial frequencies can be obtained by scaling the reported frequencies by a factor $10$. Any considered damage state reduces all the frequencies, despite the variation is rather limited even with a stiffness reduction by $25\%$, see Tab.~\ref{tab:eigen}. Moreover, the capability to perform damage localization just by exploiting these data can be largely ineffective, since some trends in the table, such as the monotonic dependence of the frequencies of a vibration mode on the damage inter-story, can be hardly recognized.

As proposed in \cite{art:Azam_Mariani, art:Azam_Mariani_2}, the shape of the vibration modes -- in particular that of the fundamental one in the case of a building featuring constant mass and stiffness at each story as for the case at hand -- should be taken into account in the analysis, in order to localise and quantify damage. As previously remarked, employing FCN allows us not only to analyse separately each recorded signal, but also to exploit their interplay. 
Moreover, even if the sensitivity to damage of displacements in horizontal and vertical directions is the same, their joint use enabled by the FCN can lead to an improvement the NN performances.

\begin{table}
\footnotesize
\[
\begin{array}{cccccccccc}
\hline
 & \multicolumn{9}{c}{\text{shear frequencies (Hz)}} \\
\hline
& \multicolumn{9}{c}{\text{damage scenario}} \\
\text{mode} & 0 & 1 & 2 & 3 & 4 & 5 & 6 & 7 & 8 \\
\hline
1 & 1.175 &	1.131 &	1.134 &	1.140 &	1.146 &	1.154 & 1.162 &	1.169 & 1.173 \\
2 & 3.484 & 3.368 &	3.427 &	3.480 &	3.467 & 3.400 & 3.355 &	3.375 & 3.445 \\
3 & 5.678 &	5.522 &	5.668 &	5.570 &	5.467 & 5.614 & 5.647 &	5.488 &	5.522 \\
4 & 7.673 &	7.521 &	7.630 &	7.401 &	7.663 & 7.448 & 7.532 &	7.600 &	7.394 \\
5 & 9.409 &	9.287 &	9.143 &	9.322 &	9.126 & 9.362 &	9.114 &	9.396 &	9.105 \\
6 & 10.825 & 10.745 & 10.471 & 10.764 &	10.665 & 10.490 & 10.812 & 10.528 & 10.598 \\
7 & 11.873 & 11.833 & 11.659 & 11.510 &	11.750 & 11.857 & 11.596 & 11.560 & 11.755 \\
8 & 12.516 & 12.505 & 12.456 & 12.378 & 12.273 & 12.227 & 12.328 & 12.421 & 12.484 \\
\hline
\end{array}
\]
\caption{Shear vibration frequencies of the considered eight-story building, for the undamaged case ($0$) and under different damage scenarios, each one featuring a reduction by $25\%$ of the stiffness at the inter-story corresponding to the scenario label.\label{tab:eigen}}
\end{table}

Due to the different range of values of vibration frequencies in the case of horizontal or vertical excitation of the structure, the axial response turns out to be richer in high-frequency vibrations. To correctly record the signals, the sampling rates have been set to $66.7$ Hz to monitor the horizontal vibrations, and $667$ Hz to monitor the vertical vibrations. For the same reason, each instance is made up by two multivariate time series, one for each excitation type, referring to different time intervals: $I=[0,10]s$ for the shear case and $I=[0,1]s$ for the axial case, respectively. Accordingly, the time series lengths are equal to $L^1_0 = 667$ and to $L^2_0 = 667$ for both the displacements in $x$ and $z$ direction. This benchmark has been exploited to test FCN architecture with either one convolutional branch or two convolutional branches (see Fig.~\ref{fig:FCN_architecture} and \ref{fig:FCN_architecture_bis})  Indeed, what we want to assess is the NN ability to perform the data fusion of the information extracted through the concatenation layer, rather than the capacity to deal with time series of different lengths.

Two load types have been considered: first, we have excited the structure with lateral and vertical loads characterised by narrow frequency ranges, randomly sampled from an interval including, but not limited to, the structural frequencies; then we have applied, once again at each story, a white noise, assessing both the case in which all the shear frequencies have been excited, and the one in which just some of them have been covered by the noise frequency spectrum. With these two load types, we have been able to assess the NN performances in two different cases:
\begin{itemize}
\item case 1 (sinusoidal load case), in which the applied load is characterized by only few (a priori, random) frequencies;
\item case 2 (white noise load case), in which the applied load is characterized by a higher number of (a priori, random) frequencies, lying in a given range; 
\end{itemize}
This latter case corresponds to the one of  random vibrations, for instance due to low-energy seismicity of natural or anthropic (urban) source \cite{art:seismic_noise}, and is frequently adopted in literature, see e.g. \cite{art:AVT}; the characteristic frequency range of seismic vibrations is site-dependent, being determined by the geographical and geological properties of the site. For example, in deep soft basins, the seismic vibrations are richer in low frequency components with respect to the ones in rock sites. For this reason, without any site characterization, it makes sense to assume more than a single frequency spectrum for the random vibrations.

\subsubsection{Case 1 (sinusoidal load case)}
\label{sec:load_case_1}
In this first analysis, two different load combinations in the horizontal ($x$) and vertical ($z$) directions have been considered, to affect both the shear and axial vibration modes of the building. For each direction, the loads applied to the stories of the structure are given by the sum of two sinusoidal functions, whose amplitudes and time variations have been randomly generated. This expression for the load has been adopted to keep its description simple and, in comparison with single sinusoidal component case, to increase the set of frequencies that excite the structure. The applied load $\boldsymbol{l} =  [\boldsymbol{l}^{sh}, \boldsymbol{l}^{ax}  ]$ reads: \vspace{-0.1cm}
\begin{align*}
{l}^{sh}_{i} \left(t, \boldsymbol{\eta}^{sh}_l \right)= \sum_{j=1}^{2} F_i^{sh}\gamma^{sh}_{i,j} \text{sin}(2 \pi f^{sh}_{j} t), \qquad   i=1,\ldots,8 ~, \vspace{-0.1cm} \\
{l}^{ax}_{i} \left(t, \boldsymbol{\eta}^{ax}_l \right)= \sum_{j=1}^{2} F_i^{ax}\gamma^{ax}_{j} \text{sin}(2 \pi f^{ax}_{j} t), \qquad  i=1,\ldots,8 ~, \vspace{-0.1cm}
\end{align*}
where: $l^{sh}_{i}\left(t, \boldsymbol{\eta}^{sh}_l \right)$ and $l^{ax}_{i}\left(t, \boldsymbol{\eta}^{ax}_l\right)$ are the amplitudes of the horizontal and vertical loads acting on the $i$-th floor; $F_i^{sh}=10^4$ kN and $F_i^{ax}=10^3$ kN are scaling parameters used to set the magnitude of the applied loads; $\boldsymbol{\eta}^{sh}_l = [\gamma^{sh}, f^{sh}  ]$ and $\boldsymbol{\eta}^{ax}_l =  [\gamma^{ax}, f^{ax}  ]$; $\gamma^{sh} \in \mathbb{R}$ and $\gamma^{ax} \in \mathbb{R}$ are random scaling factors; $f^{sh}, f^{ax} >0$ set the frequencies of the sinusoidal components (see Tab.~\ref{tab:load_composition} for the adopted random generation rules).
\begin{table}[h!]
\footnotesize
\begin{minipage}{1\linewidth}
\centering
\subfloat[$x$ direction\label{tab:load_x_param}]{%
\begin{tabular}{*3c}
\hline
parameter &   measurement  unit & adopted random  generation rule \\
\hline
$f^{sh}$ & $Hz$ &  $\lbrace \texttt{takerand}\left(\left[1,2.75,4.5,6.25,8,9.75,11.5,13.25,15\right]\right) \ \cdot \left(\texttt{randn}(0,\sqrt{2})\right)\rbrace$ \\
\\[-0.75em]
$\gamma^{sh}$ & $-$ & \parbox{6.5cm}{\centering $\lbrace \gamma^{dof}_i \cdot \texttt{randn(0,1)} \rbrace$ \\ with $\gamma^{dof}_i=\gamma^{dof}(i)$, where $i$ is the dof label and $\gamma^{dof}=\left[0.13,0.25,0.38,0.50,0.63,0.75,0.88,1.00 \right]$} \smallskip \\
\hline
\end{tabular}
}
\end{minipage}
\centering
\subfloat[$z$ direction\label{tab:load_z_param}]{%
\begin{tabular}{*3c}
\hline
parameter &   measurement  unit & adopted random  generation rule \\
\hline
$f^{ax}$ & $Hz$ &  $\lbrace \texttt{takerand}\left(\left[10,27.5,45,62.5,80,97.5,115,132.5,150\right]\right) \cdot \left( 2\texttt{randn(0,1)}\right)\rbrace$ \\
\\[-0.75em]
$\gamma^{ax}$ & $-$ & $\lbrace \texttt{randn(0,1)} \rbrace$ 	\smallskip \\
\hline
\end{tabular}
}
\caption{Adopted random generation rule for the parameters $\boldsymbol{\eta}^{1sh}_l$ and $\boldsymbol{\eta}^{1ax}_l$ tuning the frequency and the magnitude of the applied sinusoidal load components in the $x$ \protect\subref{tab:load_x_param} and $z$ \protect\subref{tab:load_z_param} directions respectively. Here, we indicate with \texttt{randn}($0,\sigma$) the sampling from a Gaussian probability distribution $\mathcal{N}\left(0,\sigma^2\right)$, where $\sigma^2$ is its variance, and with \texttt{takerand}$\left(\left[\boldsymbol{v}\right]\right)$ the uniform sampling from the discrete set of values $\left[ \boldsymbol{v} \right]$. \label{tab:load_composition}}
\end{table}

\begin{figure}[h!]
\captionsetup[subfigure]{justification=centering}
\centering
\subfloat[$f_{1,2}^{sh}=\left(21.1, 69.2 \right) \gamma_{1,2}^{sh}=\left(-0.058,-0.199\right)$][\label{fig:load_flex_43}$f_{1,2}^{sh}=\left(21.1, 69.2 \right)$ \\ $\gamma_{1,2}^{sh}=\left(-0.058,-0.199\right)$]{\includegraphics[scale=0.3]{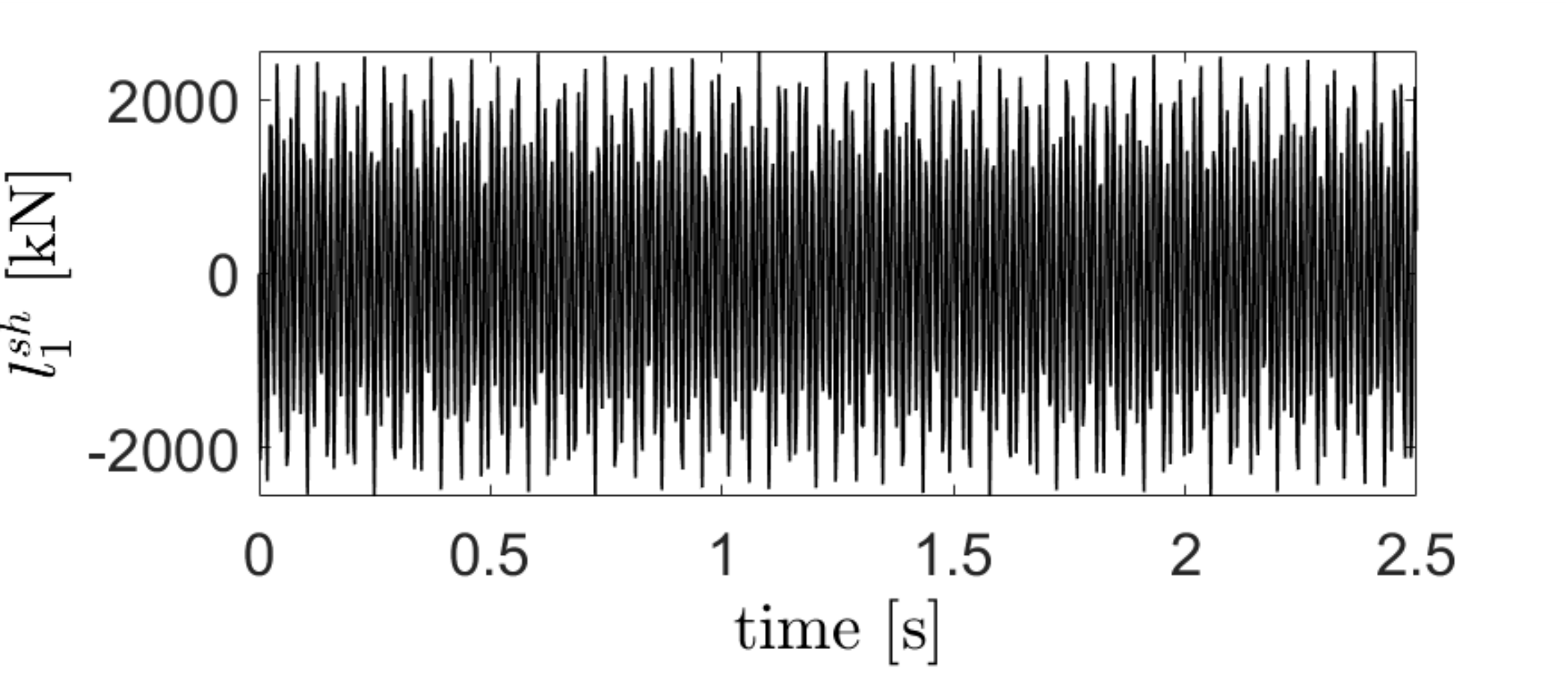}} $~$
\subfloat[$f_{1,2}^{ax}=\left(32.8, 28.2 \right) \gamma_{1,2}^{ax}=\left(1.38,1.38\right)$][\label{fig:load_axial_43}$f_{1,2}^{ax}=\left(32.8, 28.2 \right)$ \\ $\gamma_{1,2}^{ax}=\left(1.38,1.38\right)$]{\includegraphics[scale=0.3]{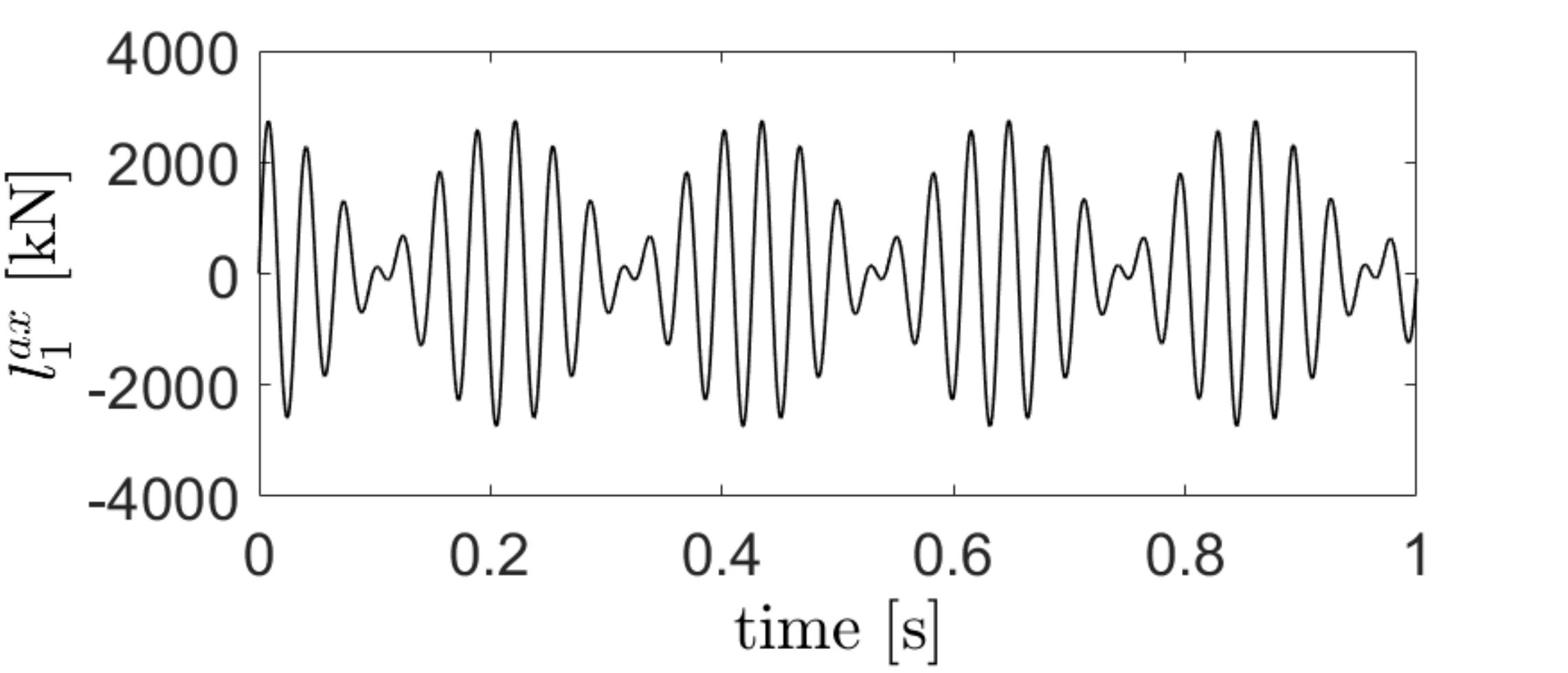}}\\ 
~\subfloat[$f_{1,2}^{sh}=\left(14.5, 2.36 \right) \gamma_{1,2}^{sh}=\left(0.025,-0.159\right)$][\label{fig:load_flex_44}$f_{1,2}^{sh}=\left(14.5, 2.36 \right)$ \\ $\gamma_{1,2}^{sh}=\left(0.025,-0.159\right)$]{\includegraphics[scale=0.3]{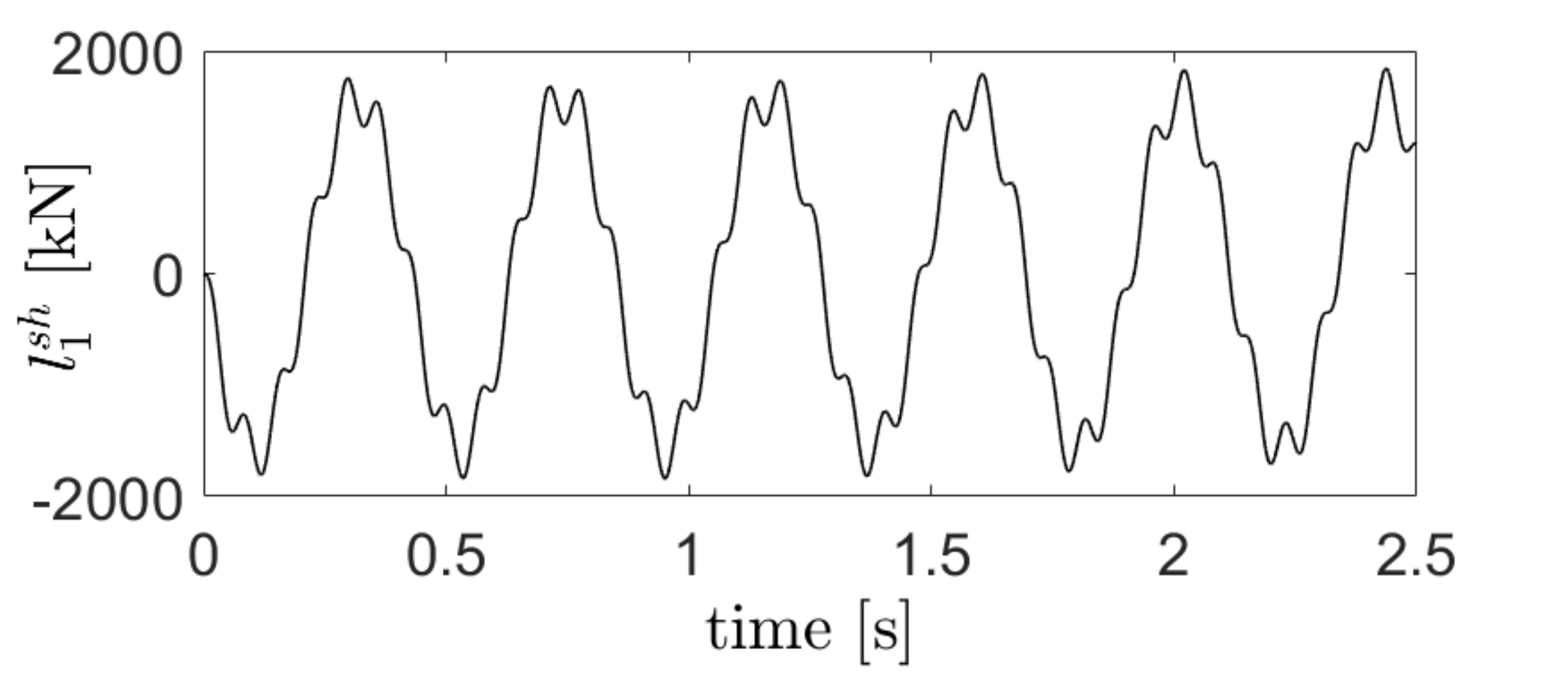}} $~$
\subfloat[$f_{1,2}^{ax}=\left(15.5, 22.0 \right) \gamma_{1,2}^{ax}=\left(1.133,-1.140\right)$][\label{fig:load_axial_44}$f_{1,2}^{ax}=\left(15.5, 22.0 \right)$ \\ $\gamma_{1,2}^{ax}=\left(1.133,-1.140\right)$]{\includegraphics[scale=0.3]{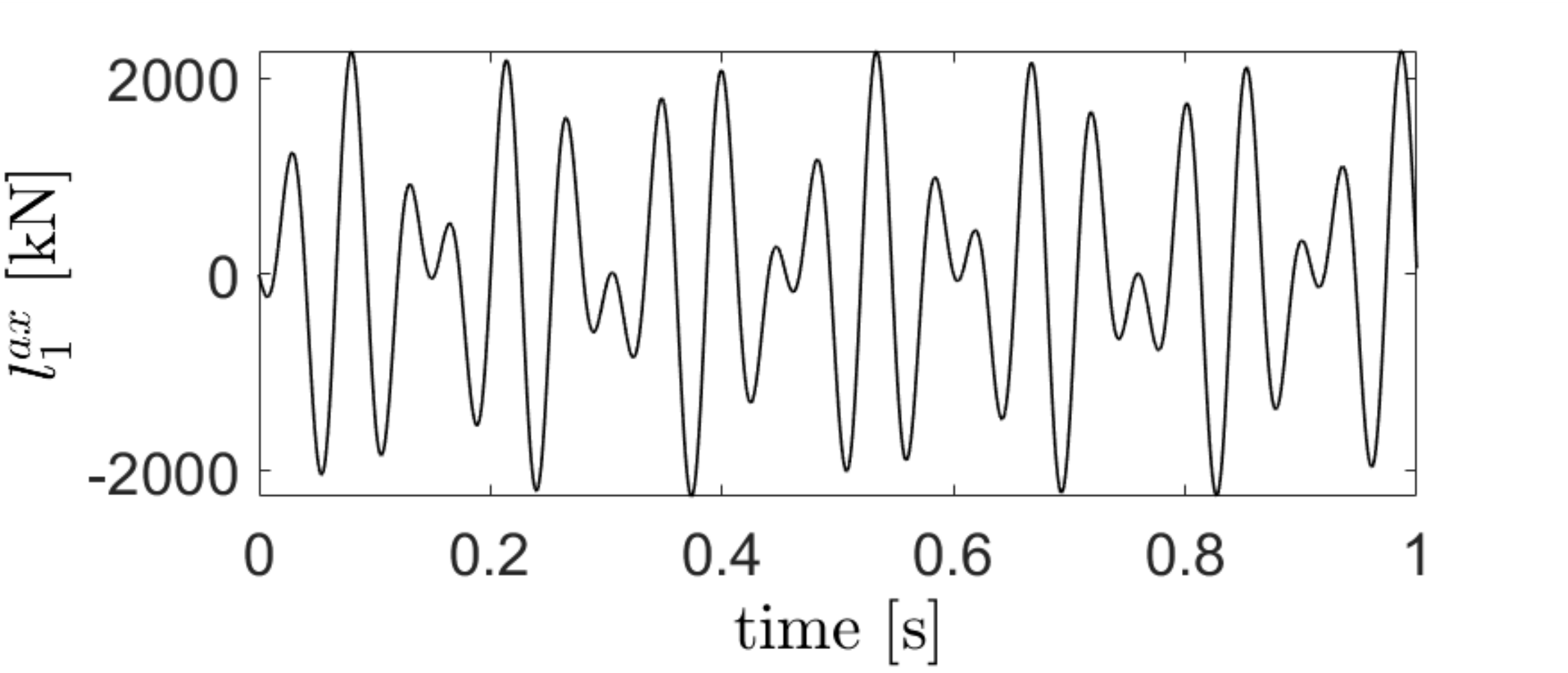}}
\caption{Examples of time evolutions of the loads (case 1) applied to the first floor of the building in the $x$ (left column) and $z$ (right column) directions. For the sake of visualisation, the sketched time interval for the loads applied in the $x$ direction has been restricted to $I=\left[0,2.5\right]s$.\label{fig:load_cases}}
\end{figure}

The two sets of values adopted for the generation rule of $f^{sh}$ and $f^{ax}$ are chosen on the basis of the structural frequencies that could be excited both in the horizontal and vertical directions. At the same time, thanks to the adopted sampling rule, $f^{sh}$ and $f^{ax}$ may exceed these frequency ranges, producing instances in which the shear frequencies and/or the axial frequencies of the structure are not excited. Regarding the generation rule of the scaling parameter $\gamma^{sh}$, its dependency on the dofs of the structure through the factor $\gamma^{dof}$ has been introduced in Table \ref{tab:load_composition} in order to mimic the load distribution usually considered in a preliminary design process, when the shear behaviour of a regular building is evaluated. Keeping in mind that our principal interest here is to assess the prediction capacities of the NN architecture, this choice has enabled us to obtain displacement time series similar to the ones expected during the monitoring of the structure, although adopting a very simple generation rule for the applied lateral loads. Some examples of the time evolutions of the generated loads, applied to the first floor of the structure (hence of $l_1^{sh}$ and $l_1^{ax}$), are shown in Fig.~\ref{fig:load_cases}.

Through Eq.~\eqref{eq:add_noise}, we have added a measurement noise to mimic the output of a real monitoring system. For the sake of simplicity, the covariance matrix $\boldsymbol{\Sigma}_{\epsilon} \in \mathbb{R}^{16 \times 16}$ of such noise has been assumed to be diagonal, i.e. $\boldsymbol{\Sigma}_{\epsilon} = \sigma^2 \mathbb{I}$ where $\sigma^2$ is the variance of the measurement error $\epsilon$ in horizontal and vertical directions for each floor, and $\mathbb{I} \in \mathbb{R}^{16 \times 16}$ is the identity matrix.

Two sources of randomness have been assumed for the noise, due to environmental effects and to the transmission of the electrical signal. Their effects are superimposed in the covariance matrix with diagonal entries respectively amounting to ${\sigma}_{env}^2$ and ${\sigma}_{el}^2$. 

The environmental noise has been assumed to induce vibrations of the same amplitude and/or to affect in the same way the converted electrical signals, independently of the building floor. Given that horizontal motions at the top of the buildings are in general greater than displacements at the lower levels, this assumption leads small amplitude signals to be more affected, in relative terms, by the environmental noise. This is reasonable if we assume that the localised disturbances that arise because of the surrounding environment have the same magnitude indipendently of the building levels.

\begin{figure}[h!]
\captionsetup[subfigure]{justification=centering}
\centering
\vspace{-0.5 cm}
\subfloat[{1-st floor\label{fig:signal_43_0_1_flex}}]{\includegraphics[scale=0.325]{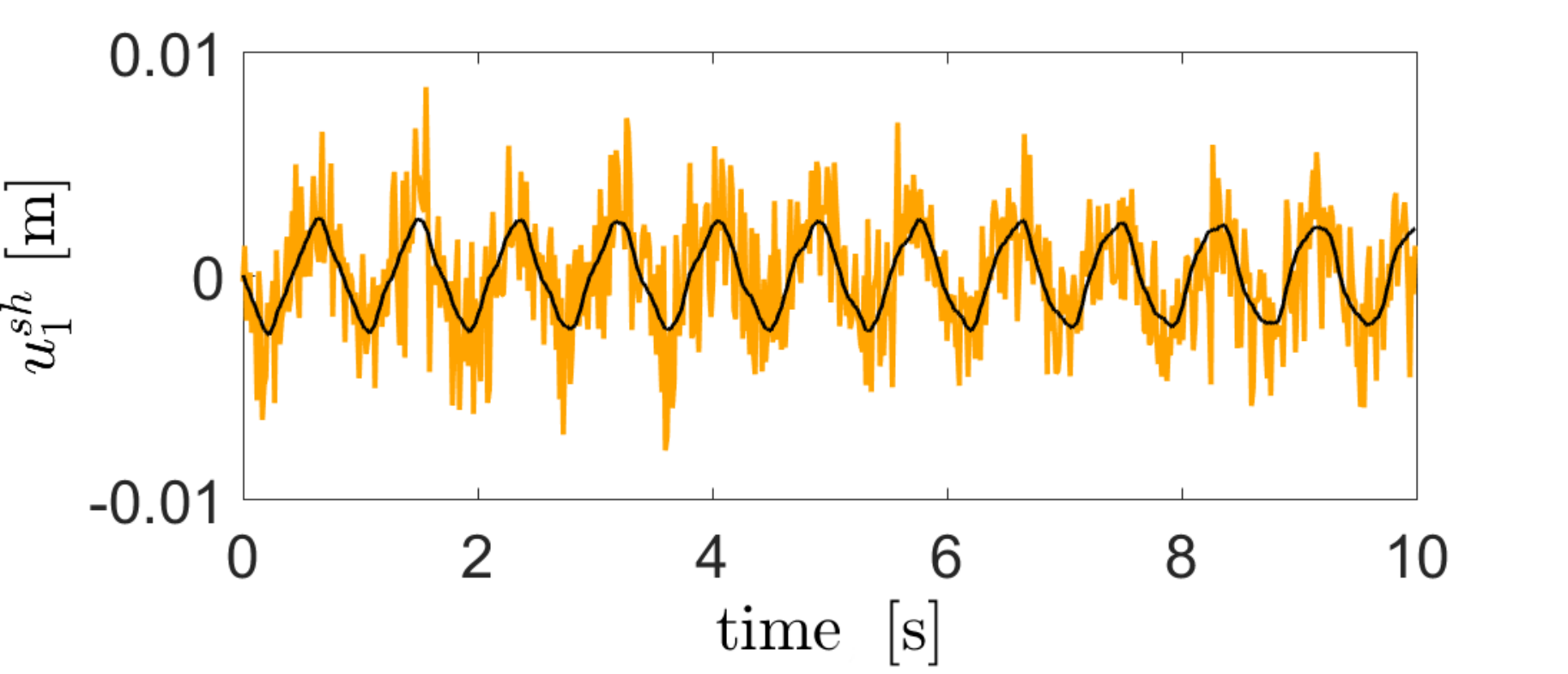}} $~$ \subfloat[{1-st floor\label{fig:signal_43_0_1_flex_enlarg}}]{\includegraphics[scale=0.325]{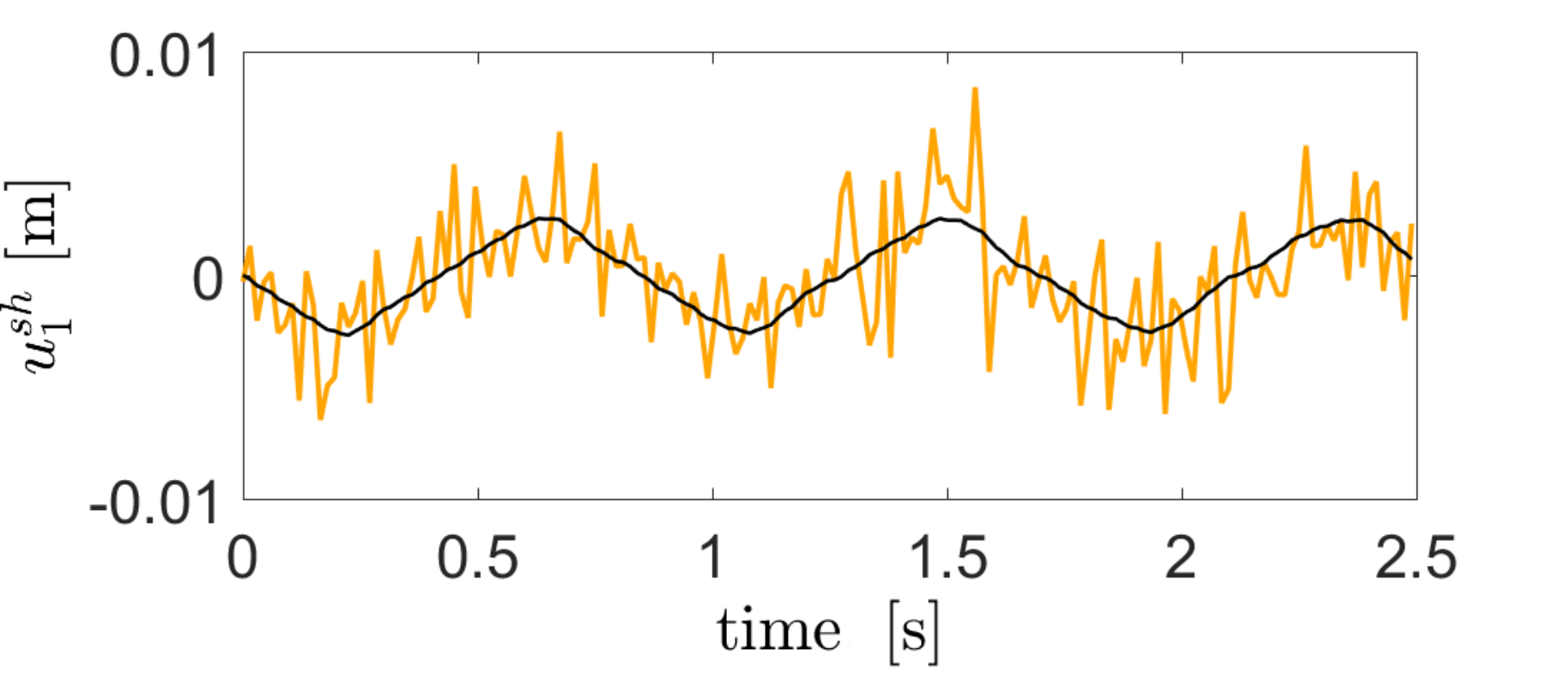}} \\
\vspace{-0.25 cm}
\subfloat[{4-th floor\label{fig:signal_43_0_4_flex}}]{\includegraphics[scale=0.325]{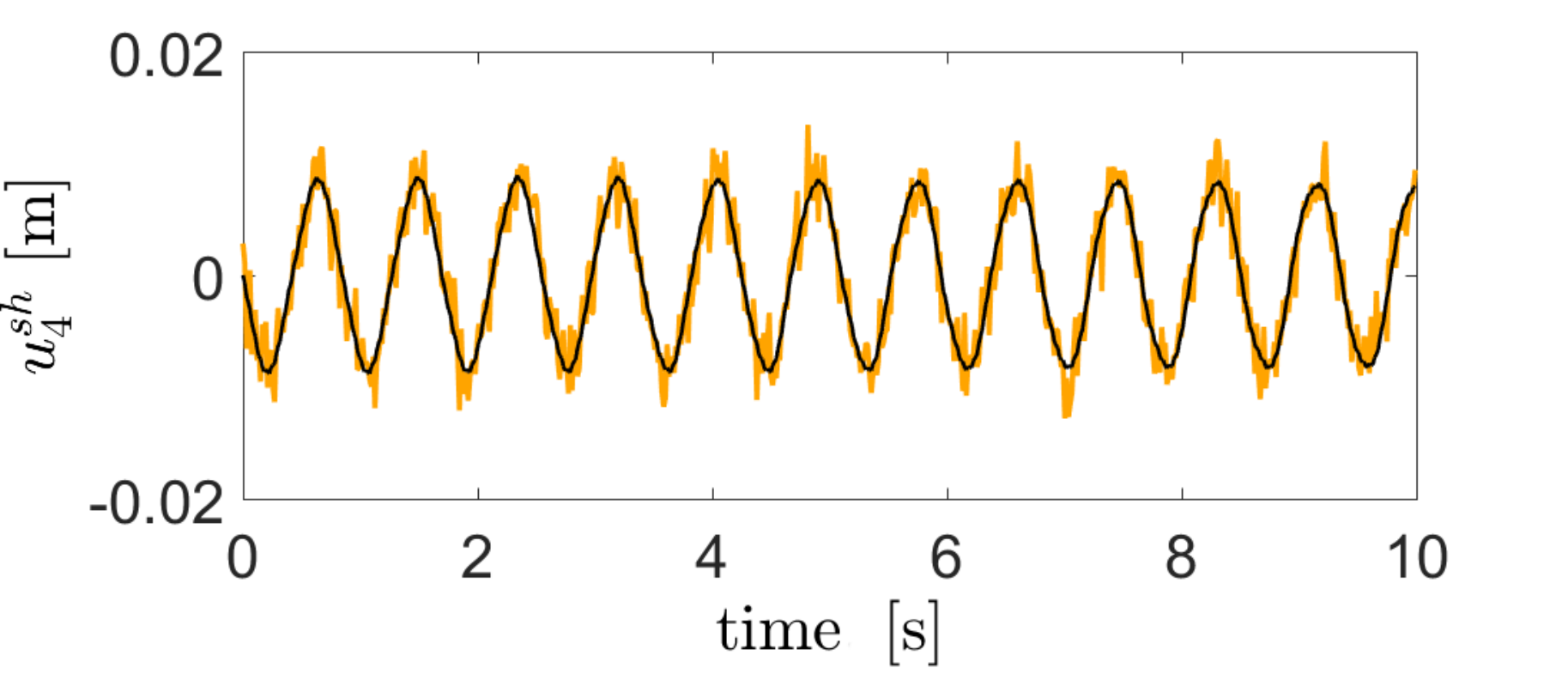}} $~$ \subfloat[{4-th floor\label{fig:signal_43_0_4_flex_enlarg}}]{\includegraphics[scale=0.325]{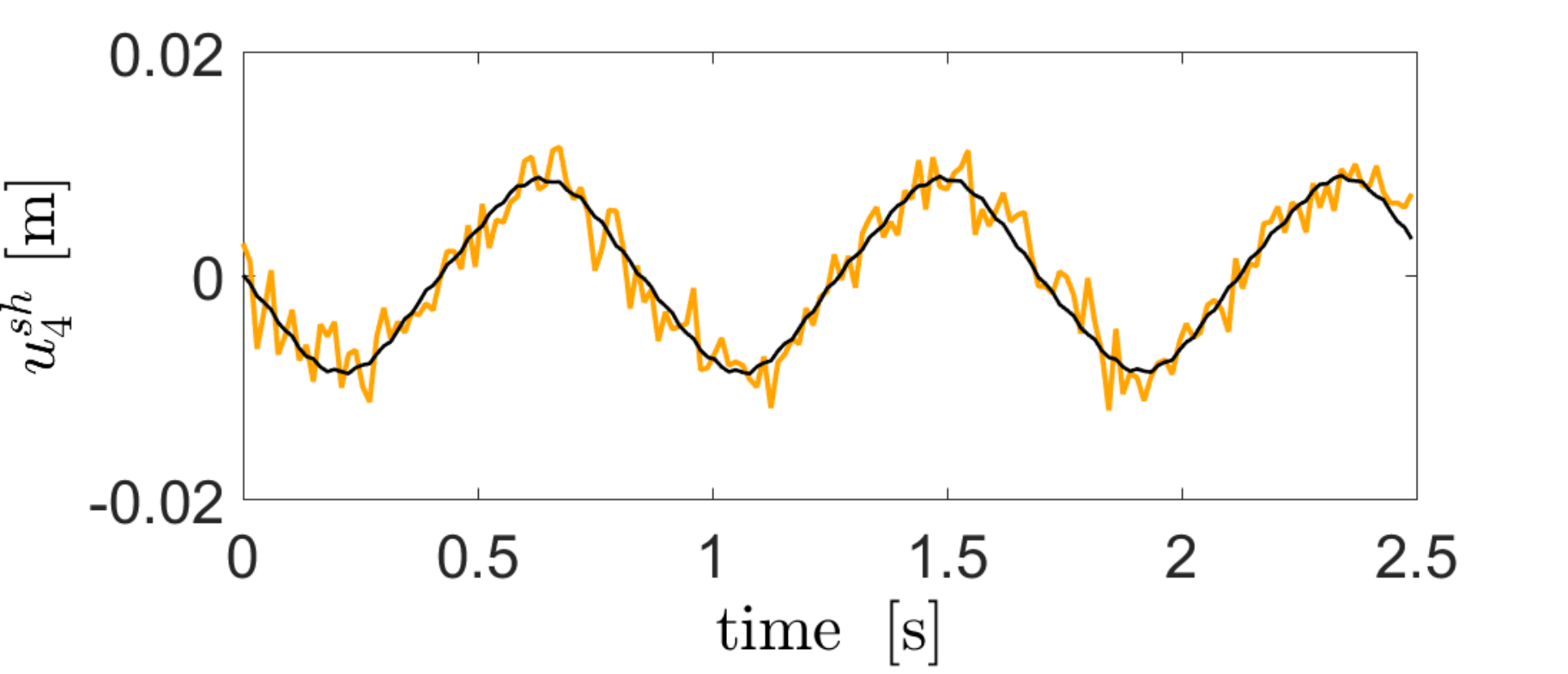}} \\
\vspace{-0.25 cm}
\subfloat[{8-th floor\label{fig:signal_43_0_8_flex}}]{\includegraphics[scale=0.325]{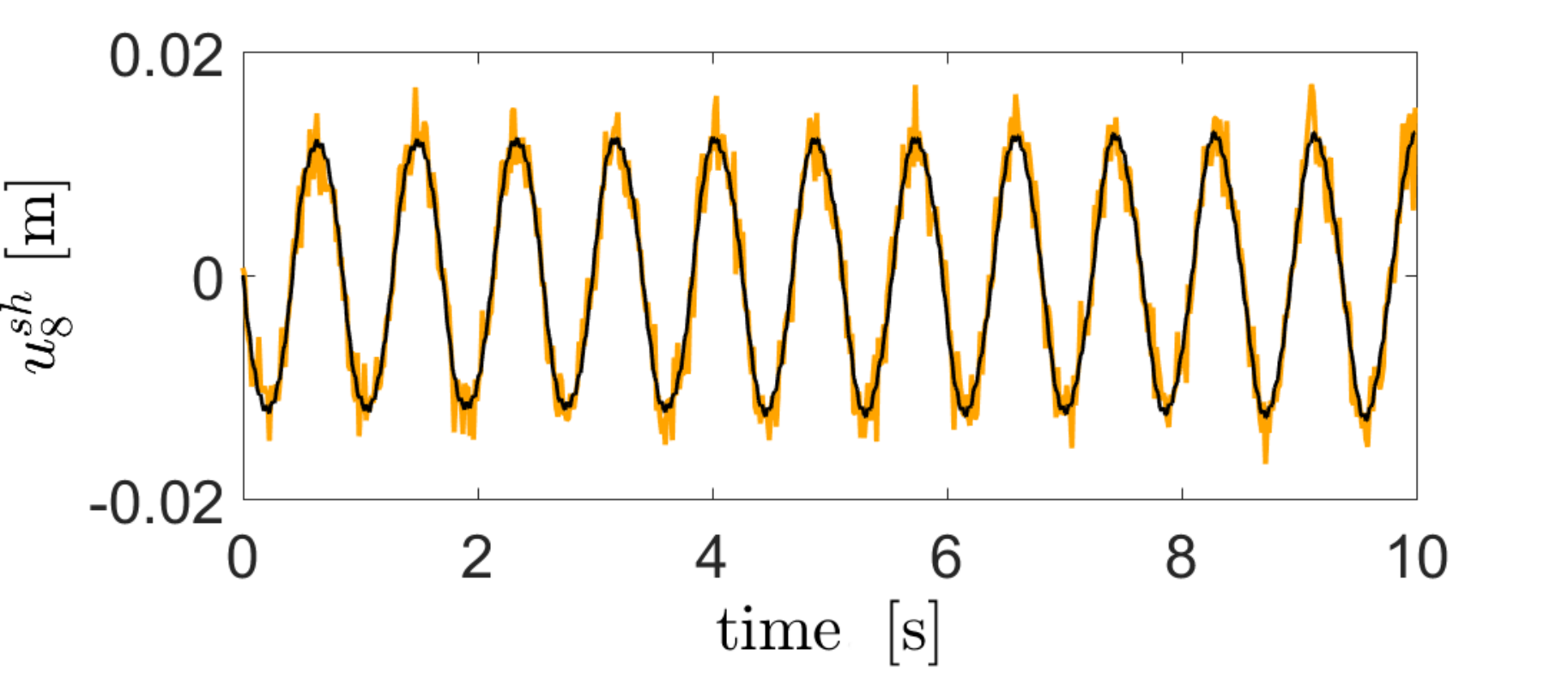}} $~$ \subfloat[{8-th floor\label{fig:signal_43_0_8_flex_enlarg}}]{\includegraphics[scale=0.325]{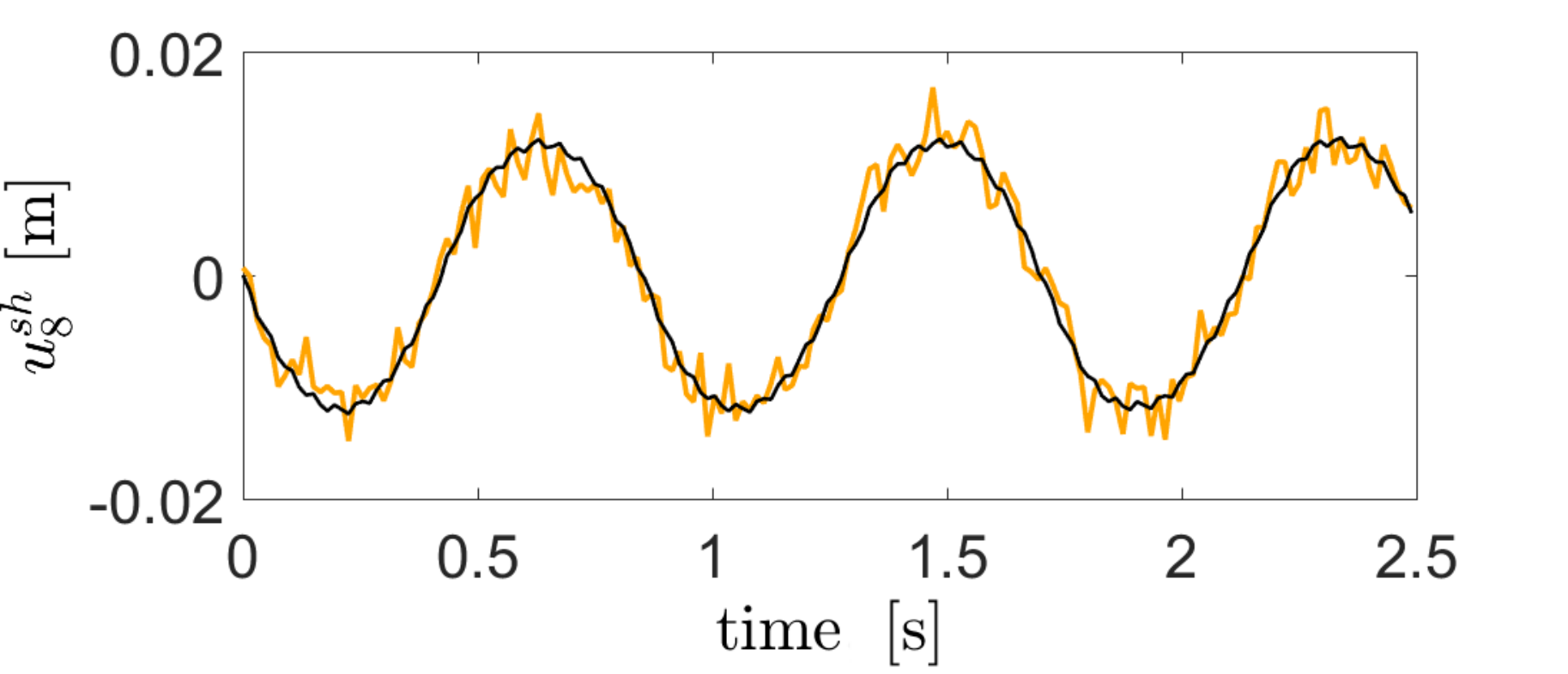}} \\
\vspace{-0.1 cm}
\subfloat[{1-st floor\label{fig:signal_44_0_1_flex}}]{\includegraphics[scale=0.325]{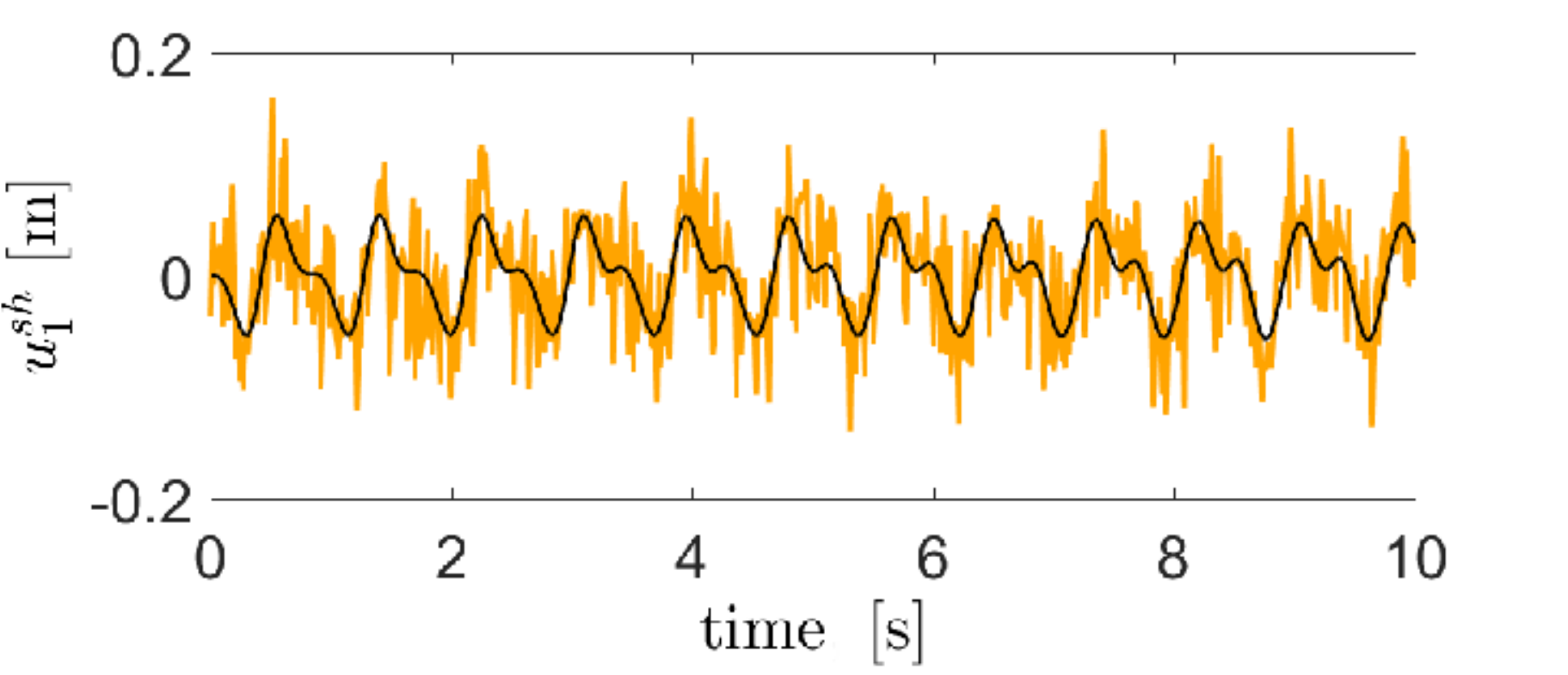}} $~$ \subfloat[{1-st floor\label{fig:44_0_1_flex_enlarg}}]{\includegraphics[scale=0.325]{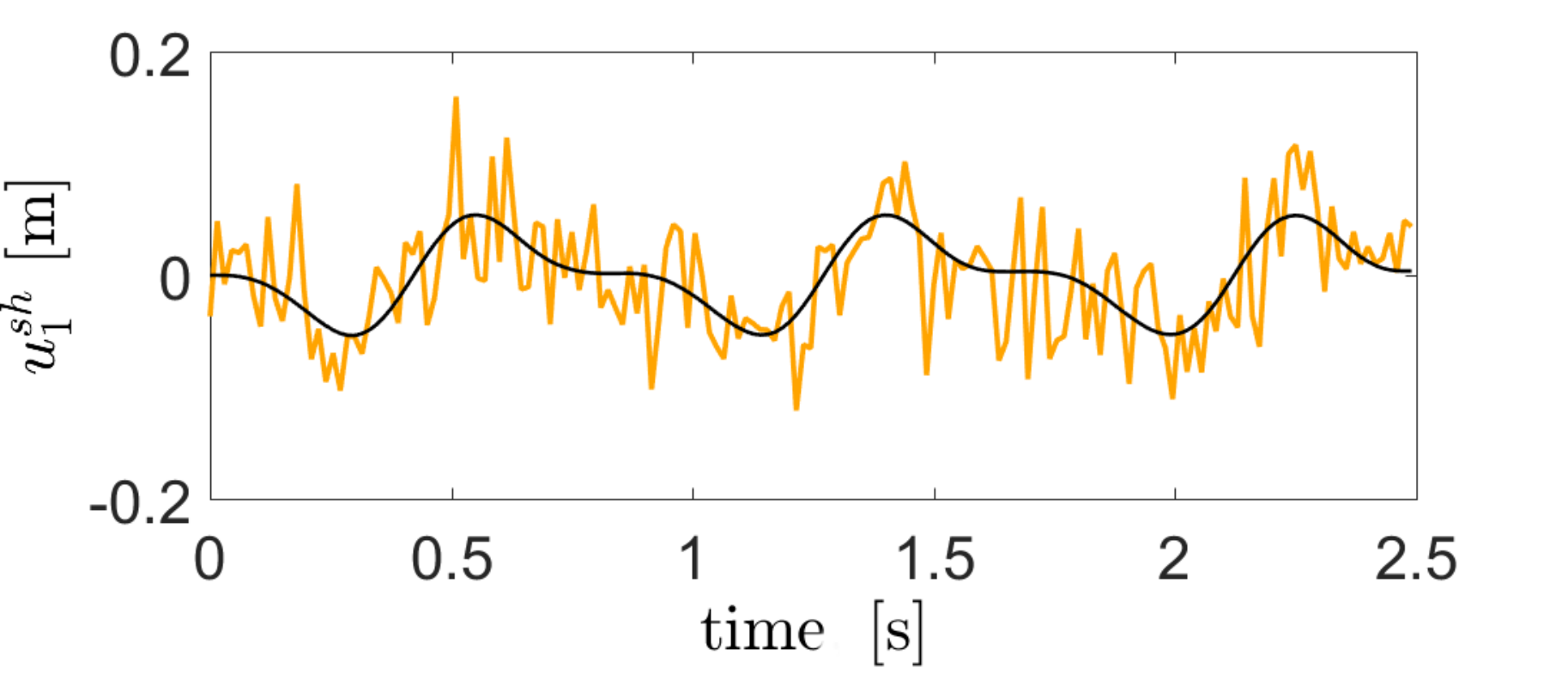}} \\
\vspace{-0.25 cm}
\subfloat[{4-th floor\label{fig:signal_44_0_4_flex}}]{\includegraphics[scale=0.325]{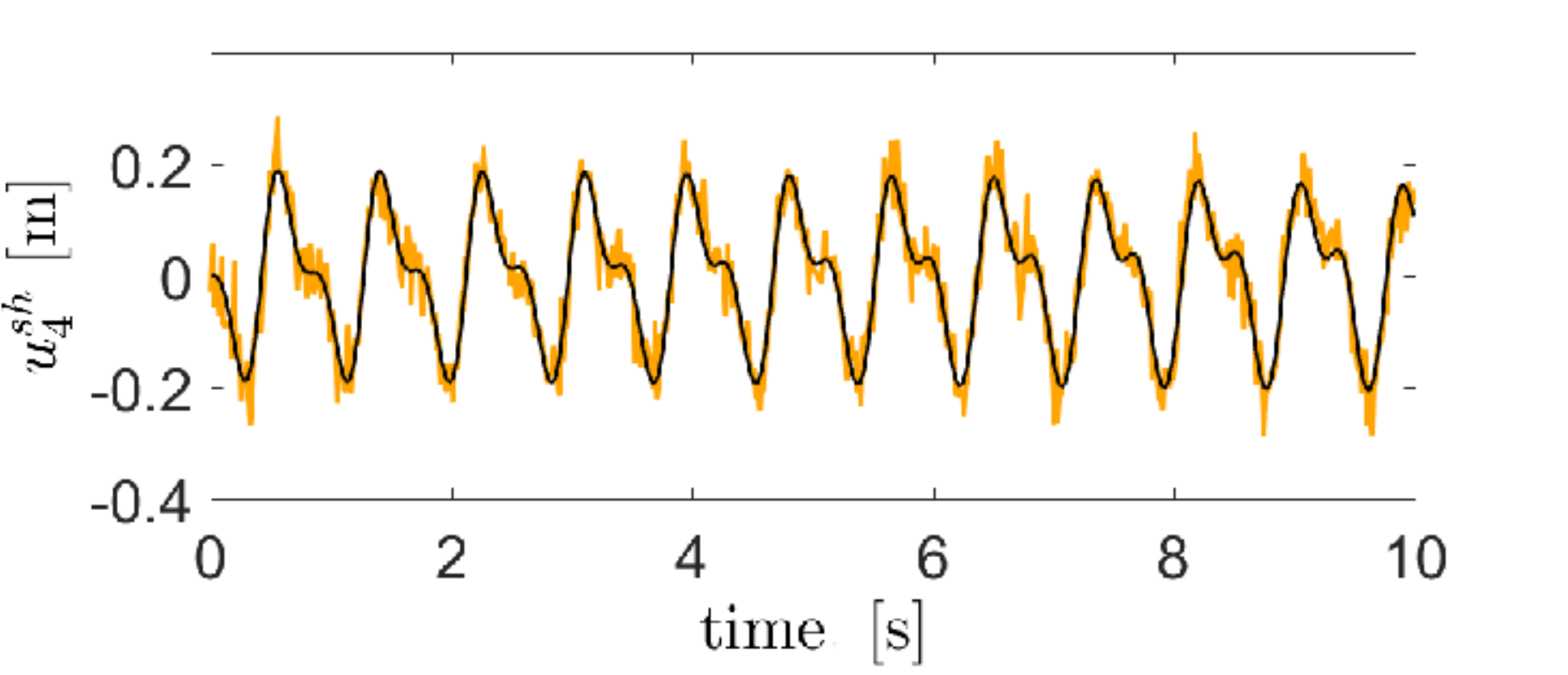}} $~$ \subfloat[{4-th floor\label{fig:signal_44_0_4_flex_enlarg}}]{\includegraphics[scale=0.325]{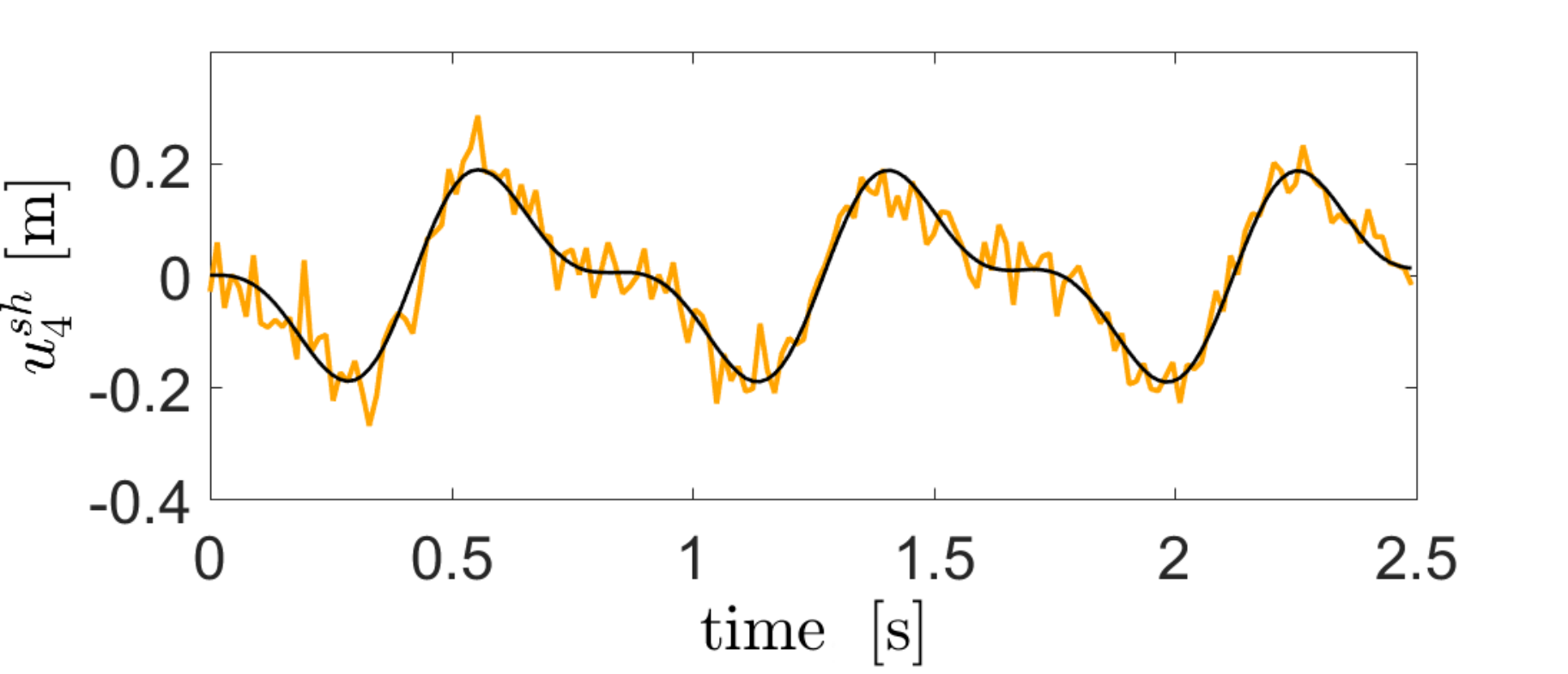}} \\
\vspace{-0.25 cm}
\subfloat[{8-th floor\label{fig:signal_44_0_8_flex}}]{\includegraphics[scale=0.35]{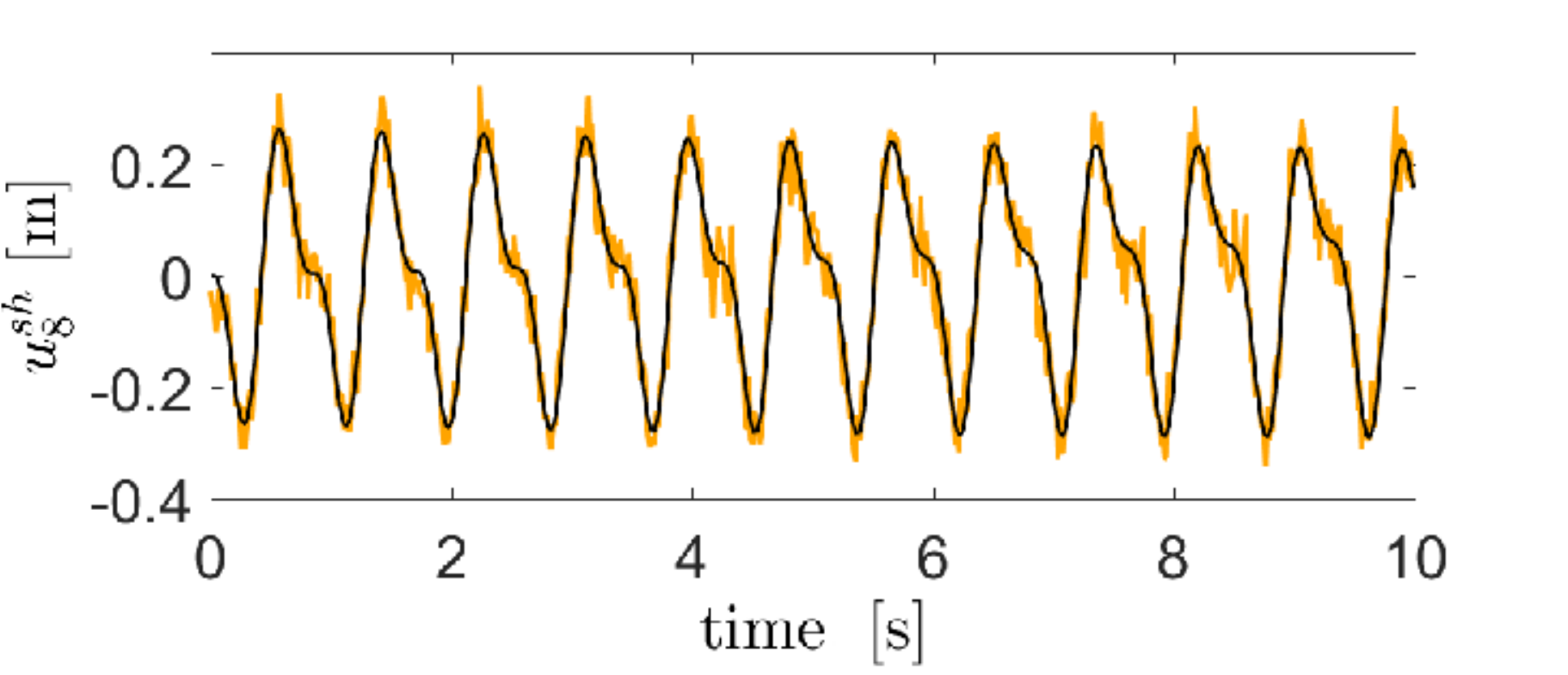}} $~$ \subfloat[{8-th floor\label{fig:signal_44_0_8_flex_enlarg}}]{\includegraphics[scale=0.35]{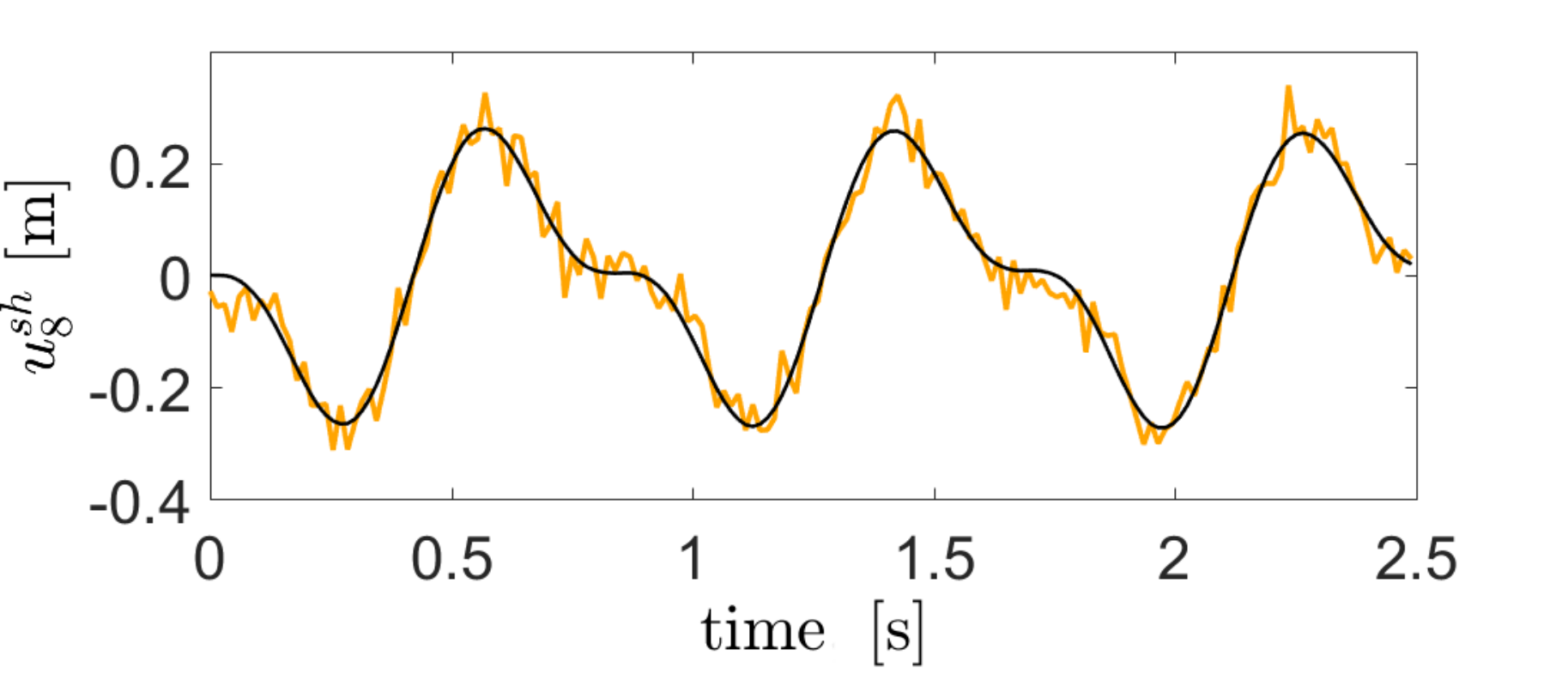}} 
\caption{Example of time evolutions of $x$ displacements for stories $1, 4, 8$ with SNR$=15$ dB (from \ref{fig:signal_43_0_1_flex} to \ref{fig:signal_43_0_8_flex_enlarg}) and SNR$=10$ dB (from \ref{fig:signal_44_0_1_flex} to \ref{fig:signal_44_0_8_flex_enlarg}), undamaged state. Low-noise case: $f_{1,2}^{sh}=\left(21.1, 69.2 \right)$, $\gamma_{1,2}^{sh}=\left(-0.058,-0.199\right)$. High-noise case: $f_{1,2}^{sh}=\left(14.5, 2.36 \right)$, $\gamma_{1,2}^{sh}=\left(0.025,-0.159\right)$. Orange lines represent $\boldsymbol{u}$, whereas black lines stand for $\boldsymbol{r}$, according to Eq.~\eqref{eq:add_noise}. On the right side, a closer view for each left side plot is reported.\label{fig:signal_flex}}
\end{figure}

\begin{figure}[h!]
\captionsetup[subfigure]{justification=centering}
\centering
\vspace{-0.25 cm}
\subfloat[{1-st floor\label{fig:signal_43_0_1_axial}}]{\includegraphics[scale=0.325]{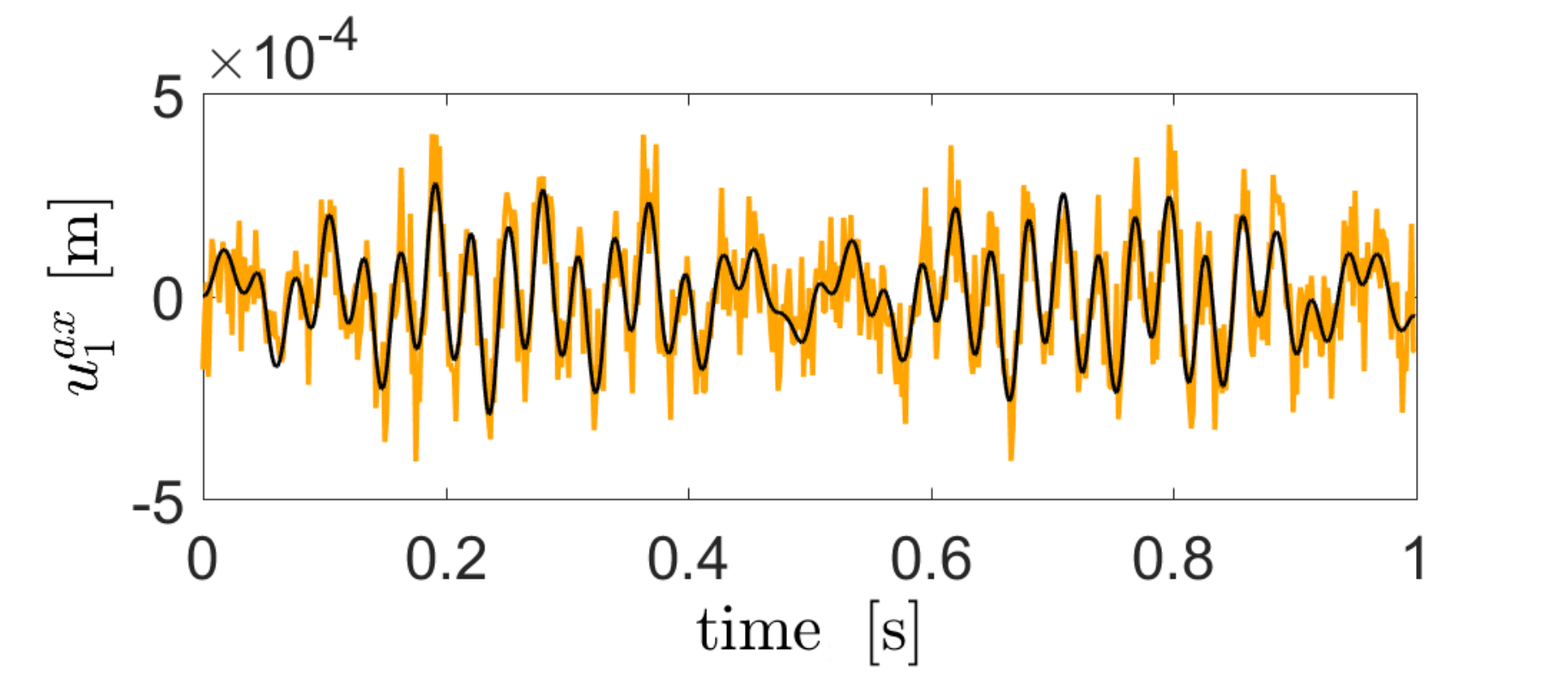}} $~$ \subfloat[{1-st floor\label{fig:signal_43_0_1_axial_enlarg}}]{\includegraphics[scale=0.325]{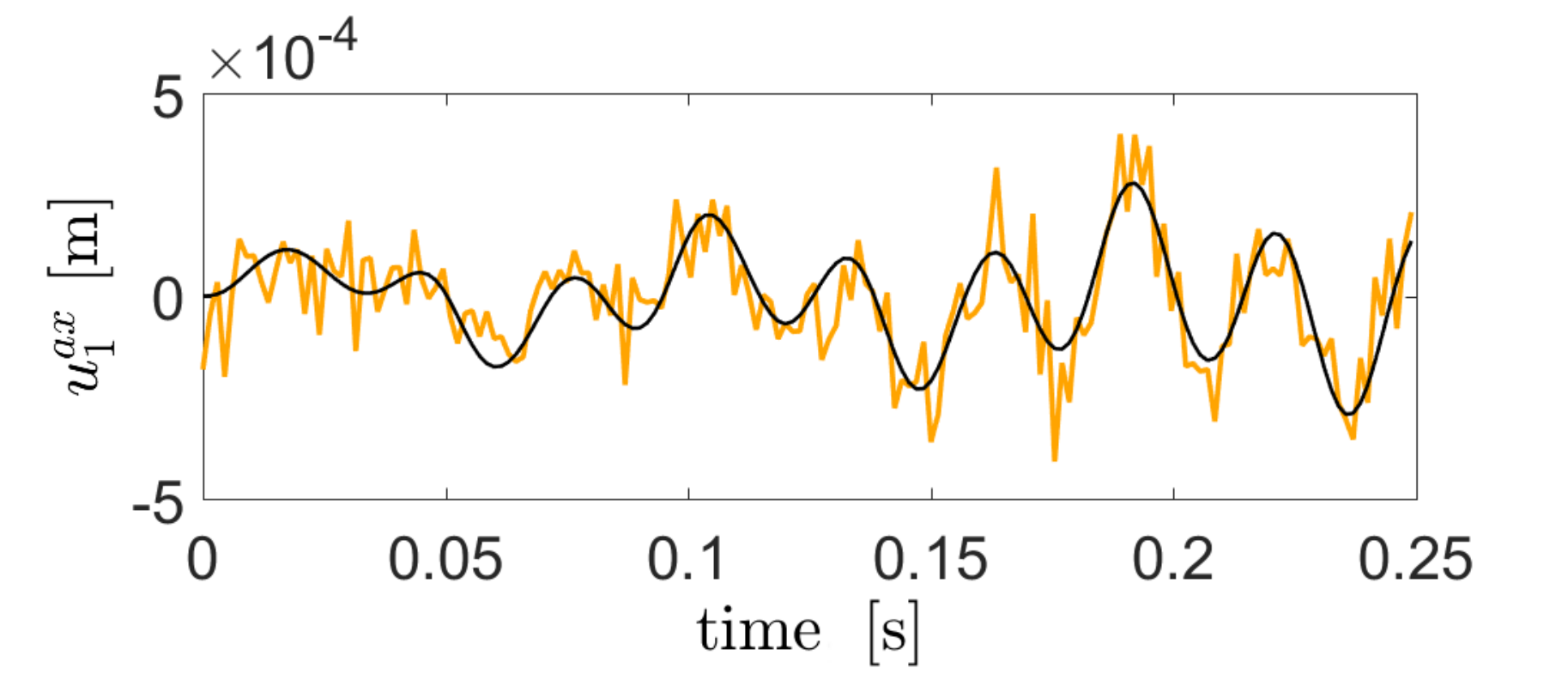}} \\
\vspace{-0.25 cm}
\subfloat[{4-th floor\label{fig:signal_43_0_4_axial}}]{\includegraphics[scale=0.325]{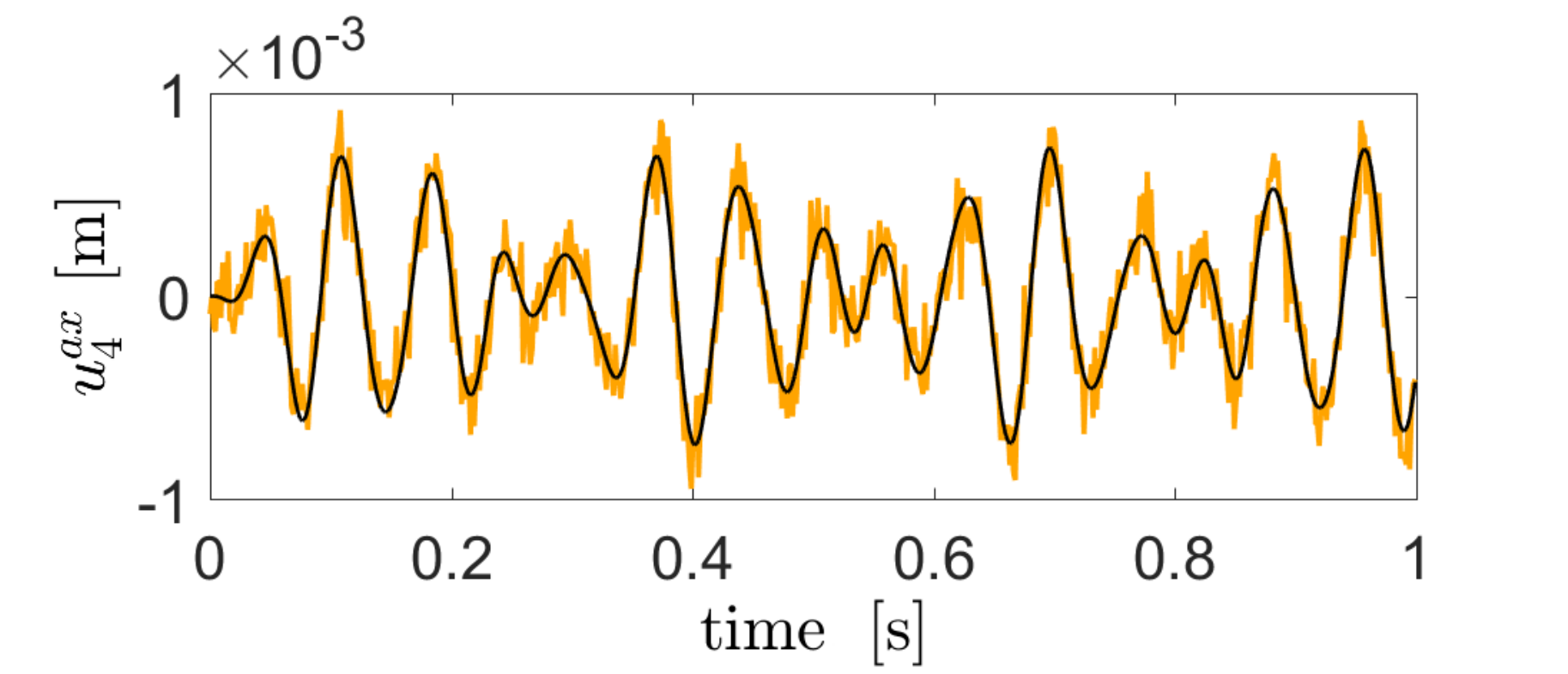}} $~$ \subfloat[{4-th floor\label{fig:signal_43_0_4_axial_enlarg}}]{\includegraphics[scale=0.325]{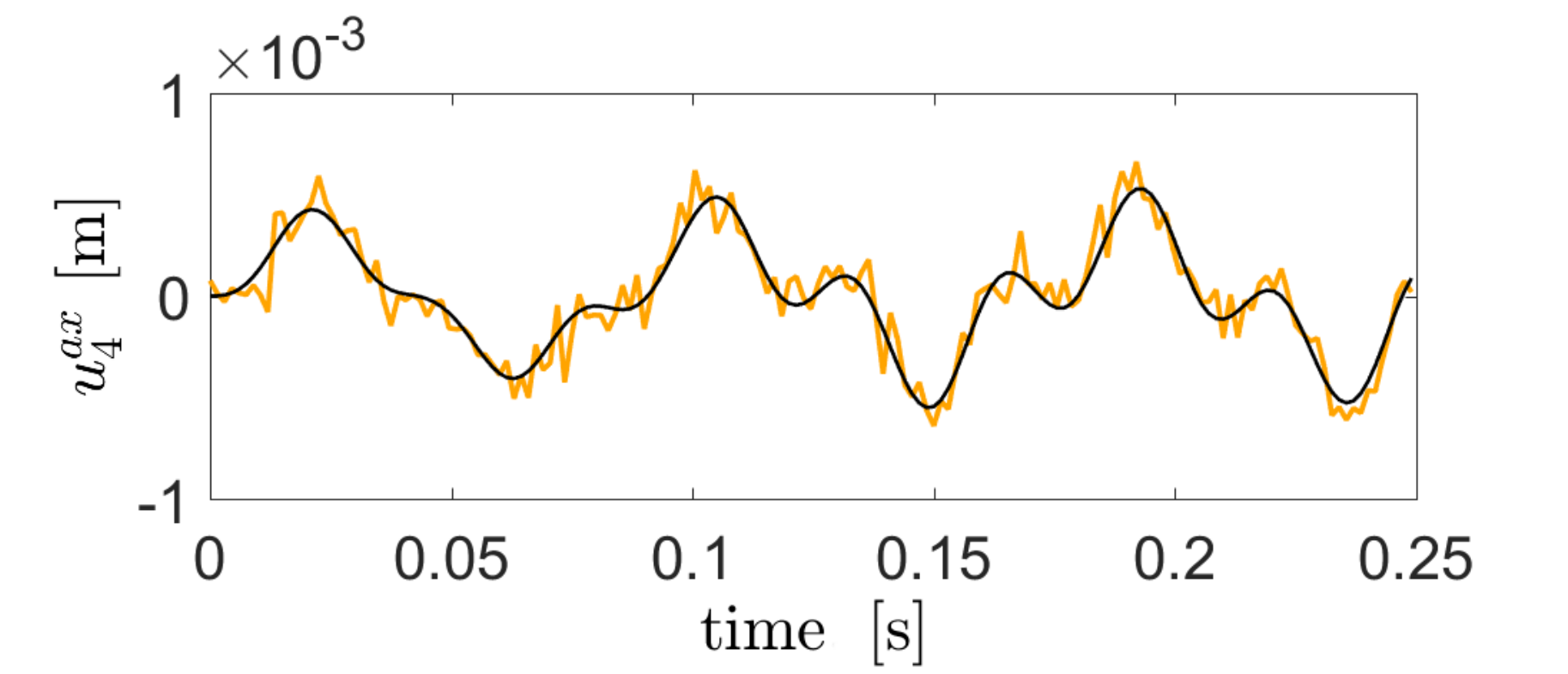}} \\
\vspace{-0.25 cm}
\subfloat[{8-th floor\label{fig:signal_43_0_8_axial}}]{\includegraphics[scale=0.325]{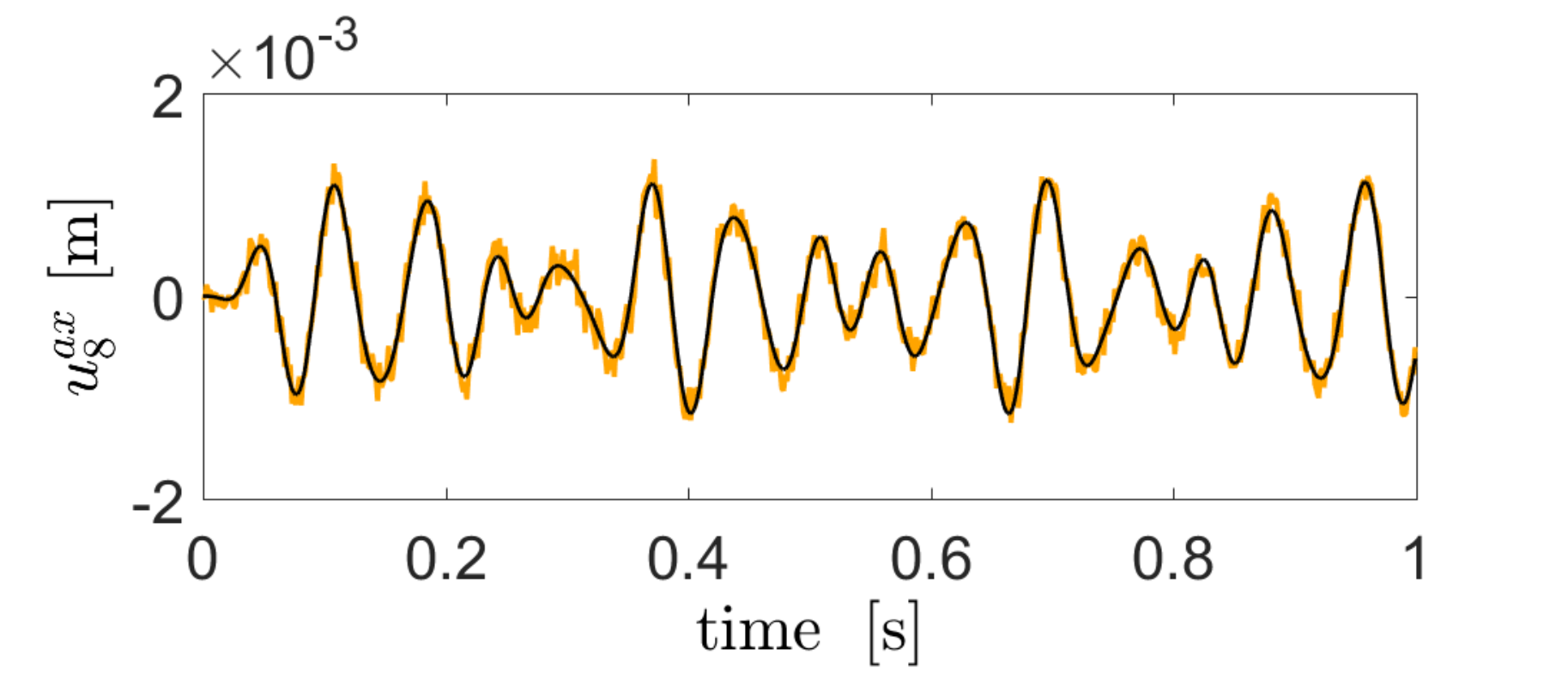}} $~$ \subfloat[{8-th floor\label{fig:signal_43_0_8_axial_enlarg}}]{\includegraphics[scale=0.325]{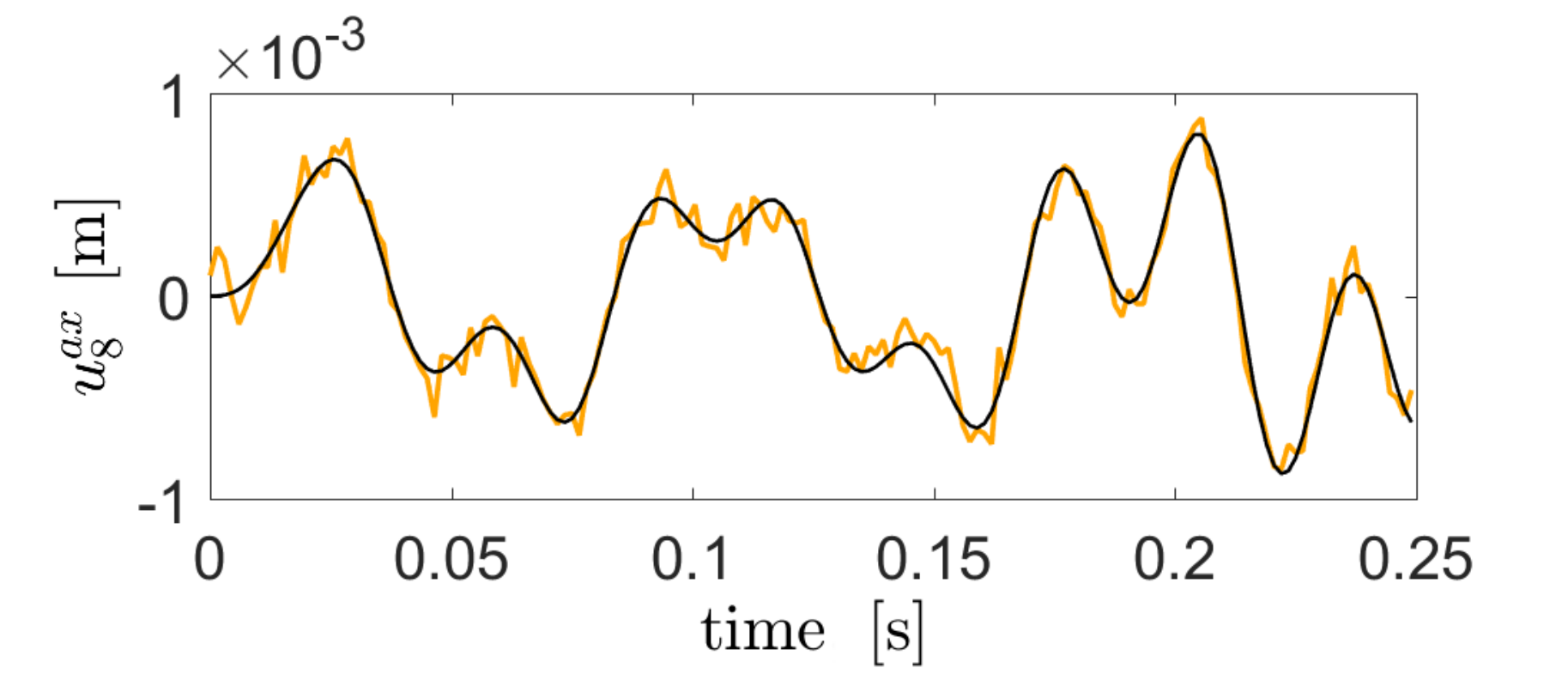}} \\
\vspace{-0.1 cm}
\subfloat[{1-st floor\label{fig:signal_44_0_1_axial}}]{\includegraphics[scale=0.325]{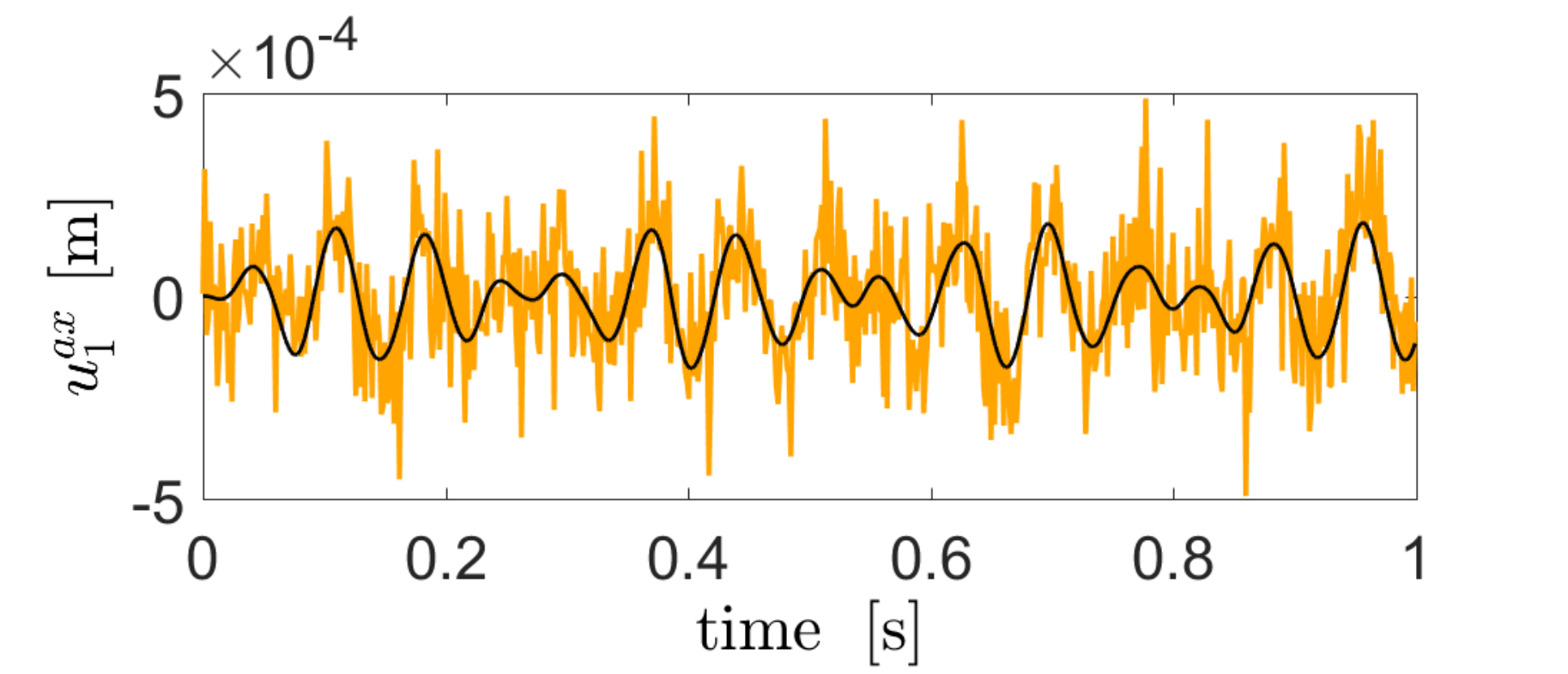}} $~$ \subfloat[{1-st floor\label{fig:44_0_1_axial_enlarg}}]{\includegraphics[scale=0.325]{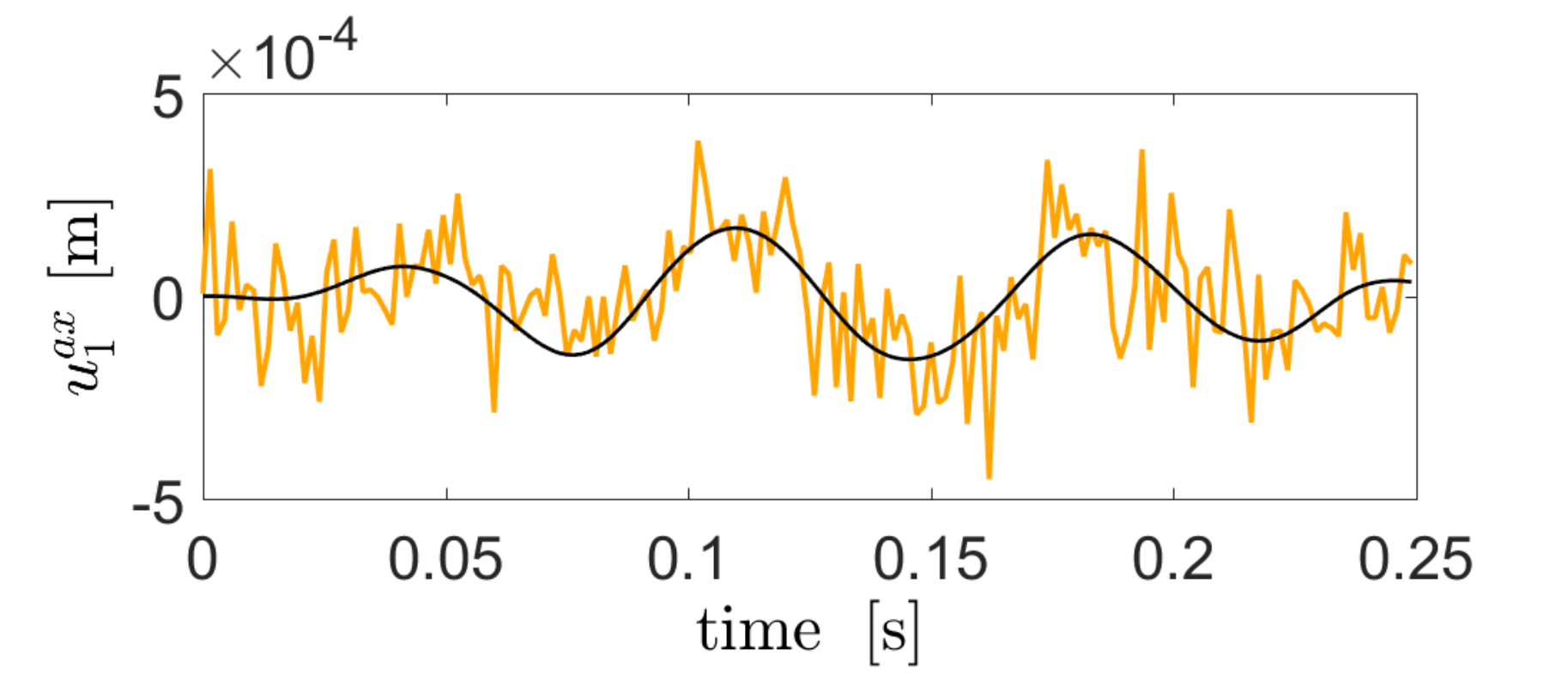}} \\
\vspace{-0.25 cm}
\subfloat[{4-th floor\label{fig:signal_44_0_4_axial}}]{\includegraphics[scale=0.325]{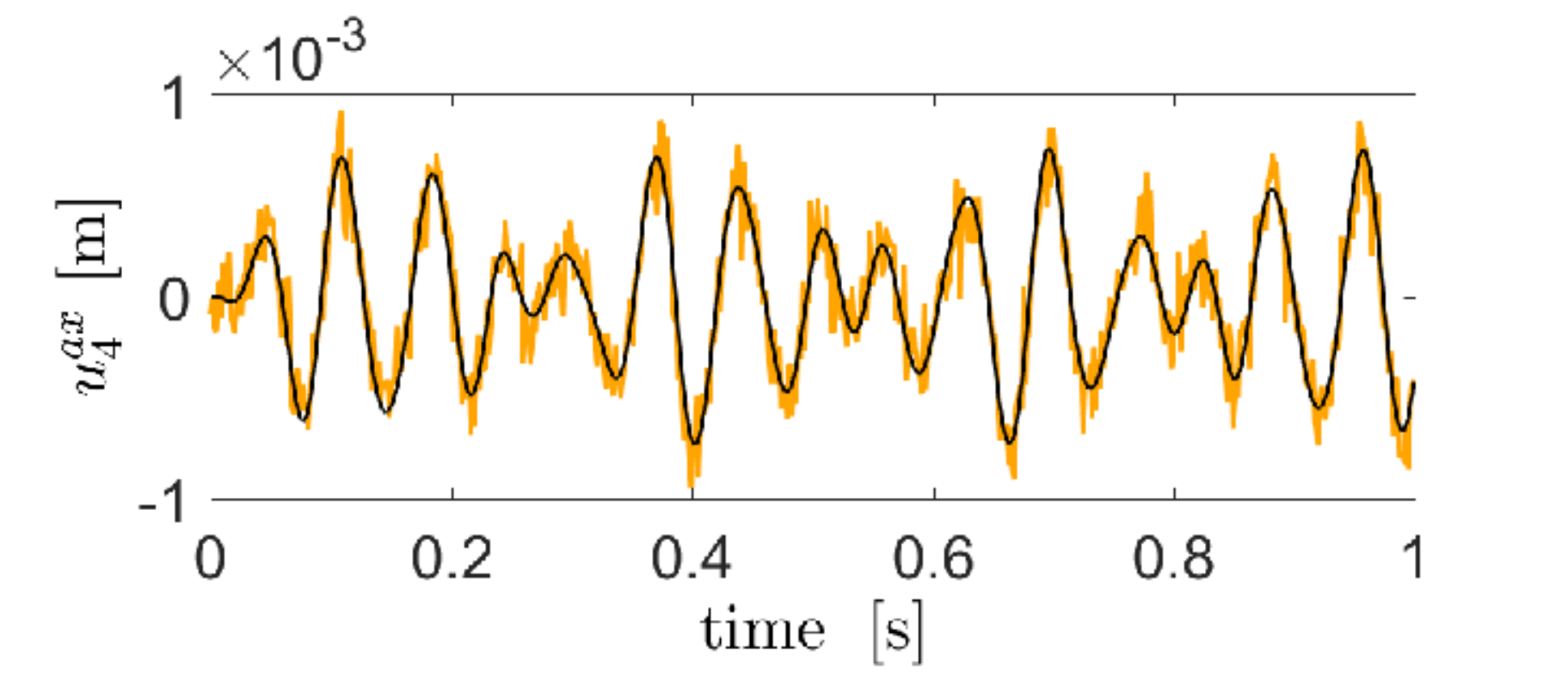}} $~$ \subfloat[{4-th floor\label{fig:signal_44_0_4_axial_enlarg}}]{\includegraphics[scale=0.325]{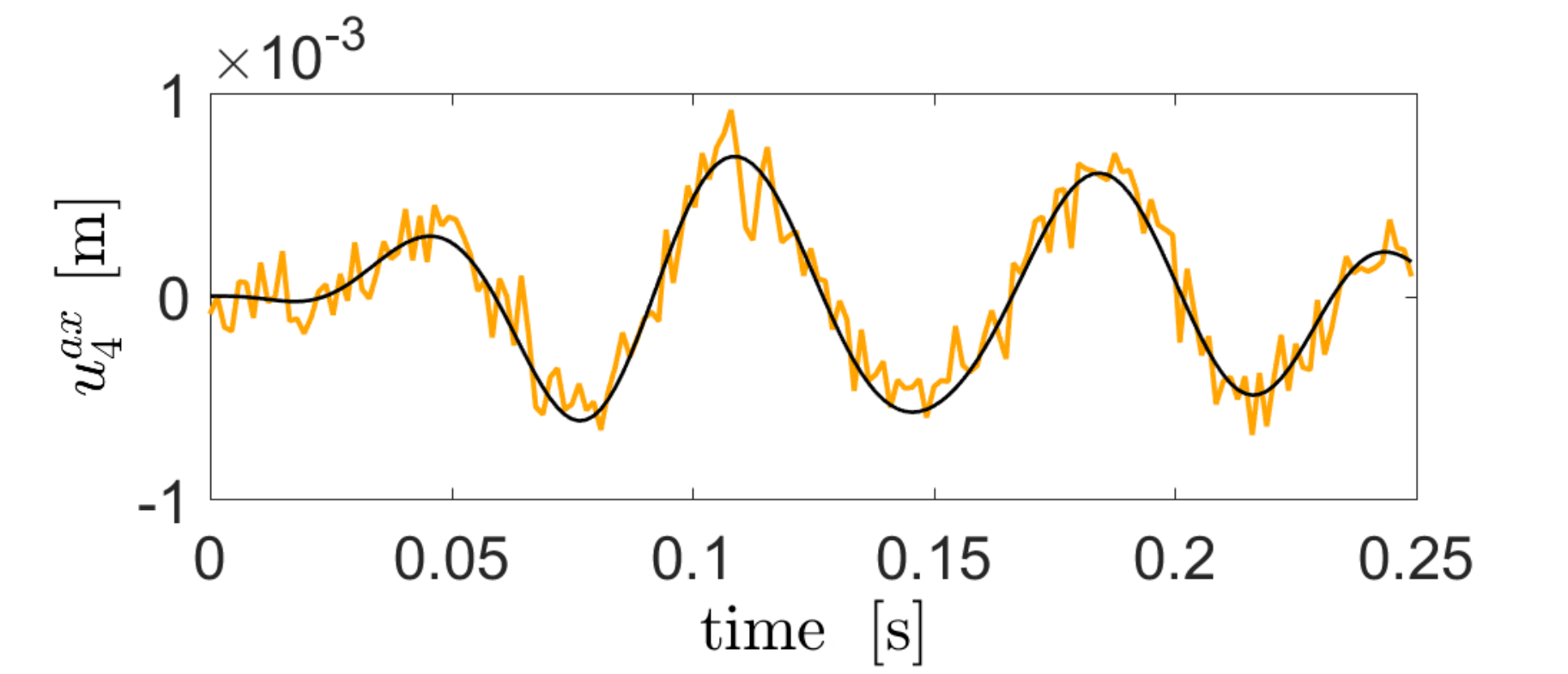}} \\
\vspace{-0.25 cm}
\subfloat[{8-th floor\label{fig:signal_44_0_8_axial}}]{\includegraphics[scale=0.325]{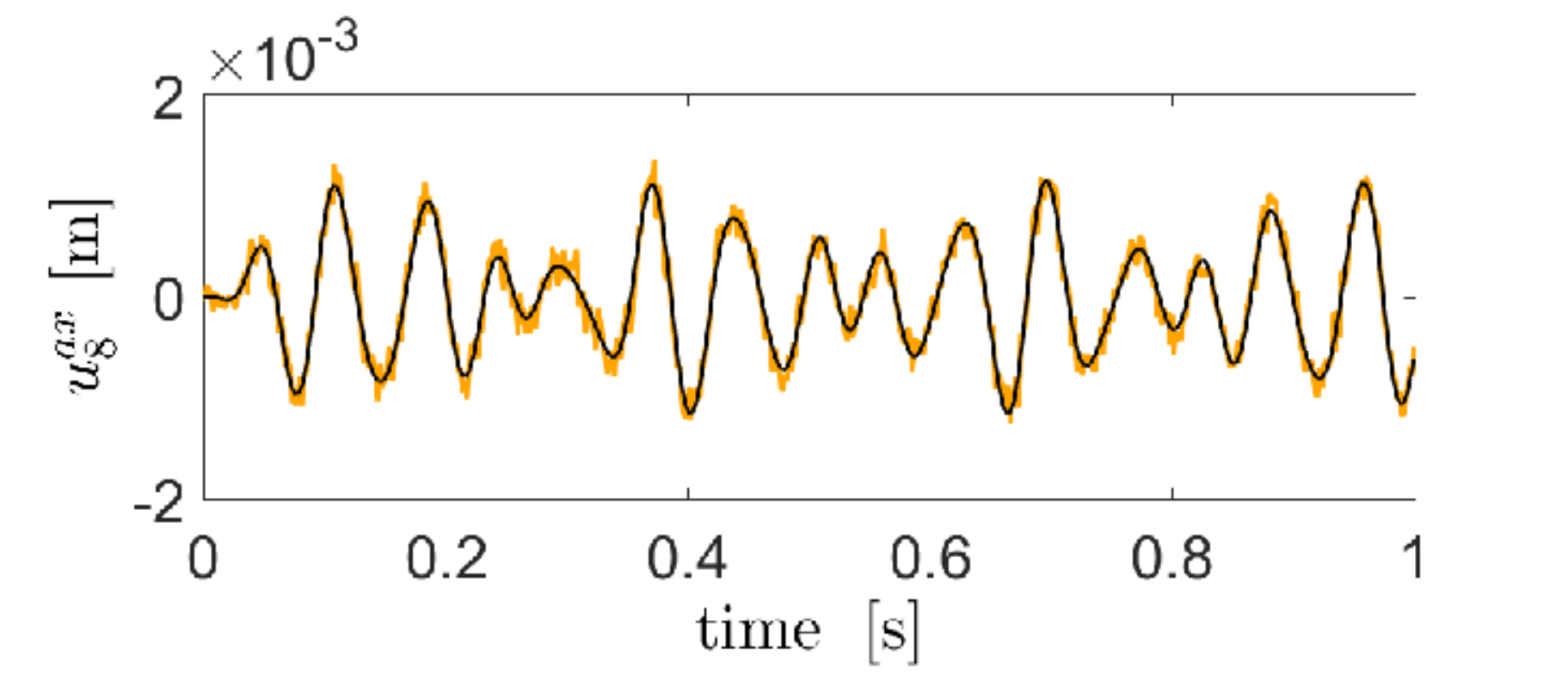}} $~$ \subfloat[{8-th floor\label{fig:signal_44_0_8_axial_enlarg}}]{\includegraphics[scale=0.325]{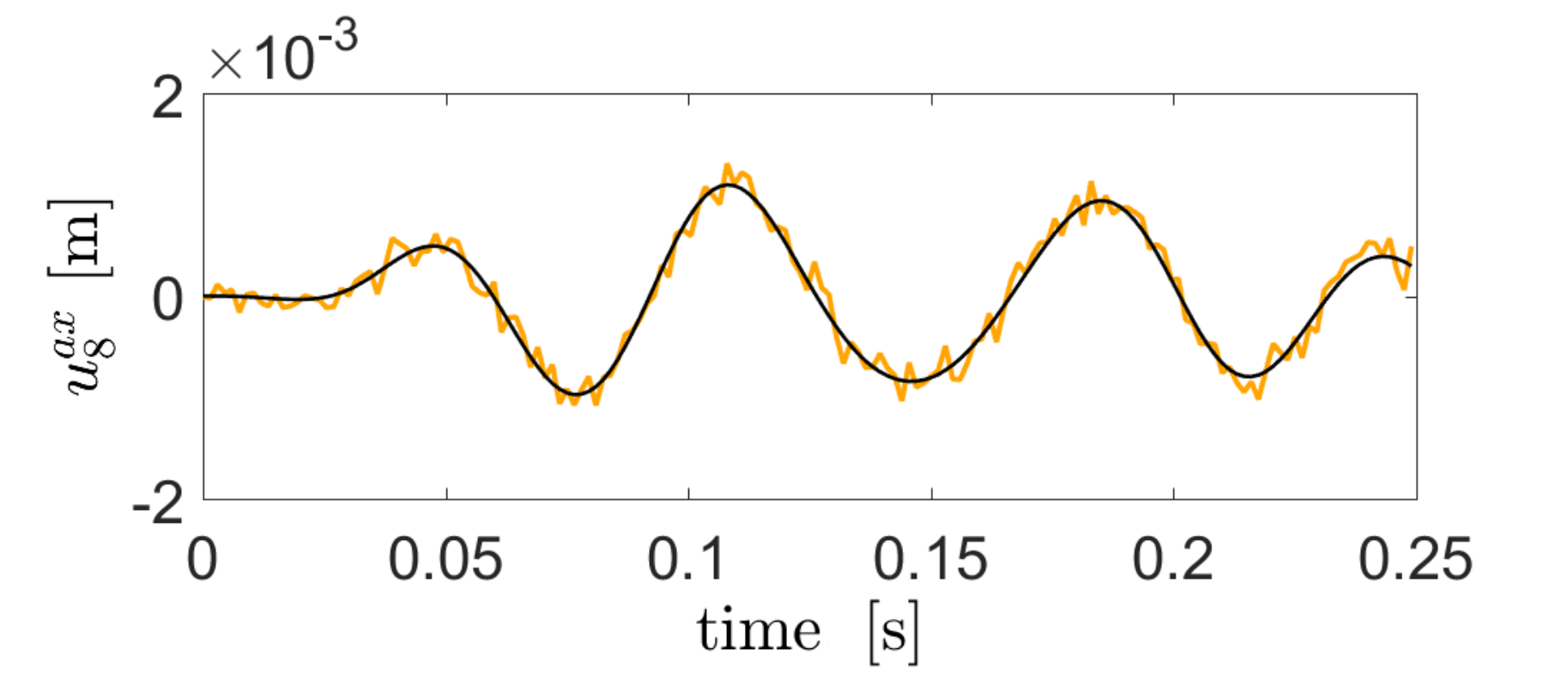}} 
\caption{Example of time evolutions of $z$ displacements for stories $1, 4, 8$ with SNR$=15$ dB (from \ref{fig:signal_43_0_1_axial} to \ref{fig:signal_43_0_8_axial_enlarg}) and SNR$=10$ dB (from \ref{fig:signal_44_0_1_axial} to \ref{fig:signal_44_0_8_axial_enlarg}), undamaged state. Low-noise case: $f_{1,2}^{ax}=\left(32.8, 28.2 \right)$, $\gamma_{1,2}^{ax}=\left(1.38,1.38\right)$. High-noise case: $f_{1,2}^{ax}=\left(15.5, 22.0 \right)$, $\gamma_{1,2}^{ax}=\left(1.133,-1.140\right)$. Orange lines represent $\boldsymbol{u}$, whereas black lines stand for $\boldsymbol{r}$, according to Eq.~\eqref{eq:add_noise}. On the right side, a closer view for each left side plot is reported.\label{fig:signal_axial}}
\end{figure}

Regarding the electrical disturbance, the same noise level has been assumed both in directions $x$ and $z$, despite of the usually different technical specifications for sensors measuring displacements with different magnitude. This means that the electrical disturbances have the same effect, in statistical terms, on the measurement outcomes in horizontal direction ${u}^{sh}_i$ and in vertical direction ${u}^{ax}_i$. Fig.~\ref{fig:signal_flex} and Fig.~\ref{fig:signal_axial} respectively show examples of time evolutions of horizontal and vertical displacements, to highlight the effects of the above assumptions on the structural signals. These displacement components always refer to the undamaged case, and to the load conditions specified in the captions. According to what highlighted, it is noted that the displacements of the $8$-th story are less affected by noise than the ones of the $1$-st story.

Due to the random generation of the applied load, different structural frequencies are excited in each simulation. To provide different scenarios also in terms of sensor accuracy (see also \cite{art:Giovanni_Capellari_2}) two levels of Signal to Noise Ratio (SNR) of $15$ dB and $10$ dB have been adopted. The SNR is a summary indicator, referring to the overall level of noise corruption for the displacements in one direction. Still referring to Fig.~\ref{fig:signal_flex} and Fig.~\ref{fig:signal_axial}, differences in terms of corruption levels between the two sensor accuracy scenarios can be appreciated.

\newpage

\newpage

\begin{figure}[h!]
\captionsetup[subfigure]{justification=centering}
\centering
\vspace{-0.1cm}
\subfloat[{undamaged scenario\label{fig:signal_44_0_8_flex_enlarg_2}}]{\includegraphics[scale=0.325]{44_0_8_flex_enlarg.pdf}} \vspace{-0.15cm}\\
\subfloat[{damaged scenario 1\label{fig:comp_signal_44_1_8_flex_enlarg}}]{\includegraphics[scale=0.325]{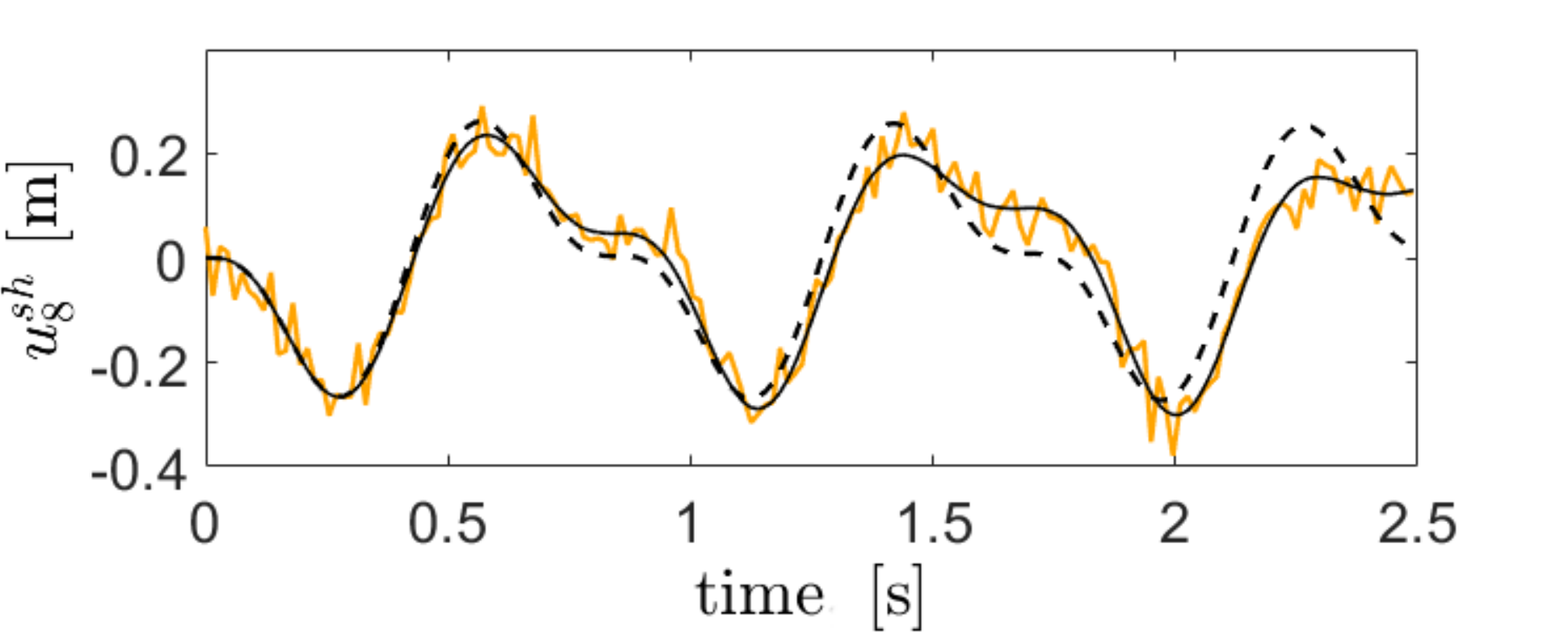}} $~$ \subfloat[{damaged scenario 2\label{fig:comp_signal_44_2_8_flex_enlarg}}]{\includegraphics[scale=0.325]{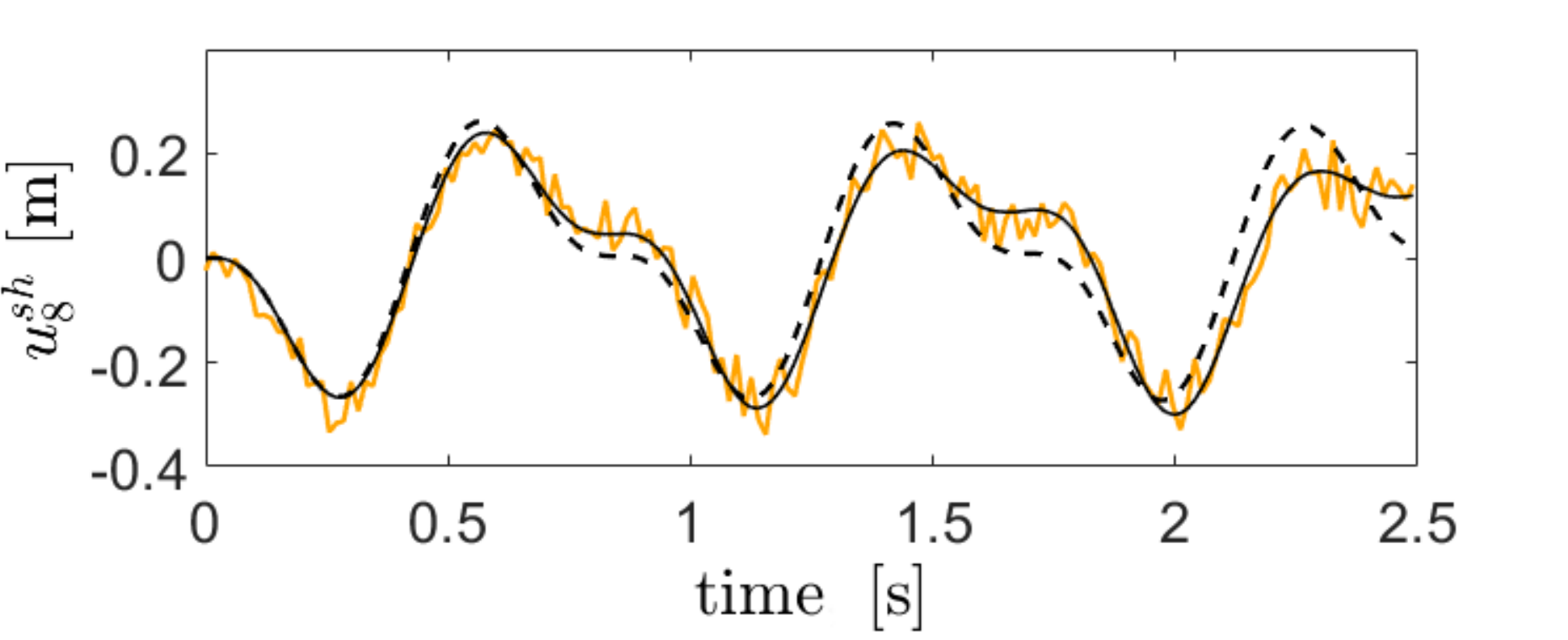}} \vspace{-0.15cm}\\
\subfloat[{damaged scenario 3\label{fig:comp_signal_44_3_8_flex_enlarg}}]{\includegraphics[scale=0.325]{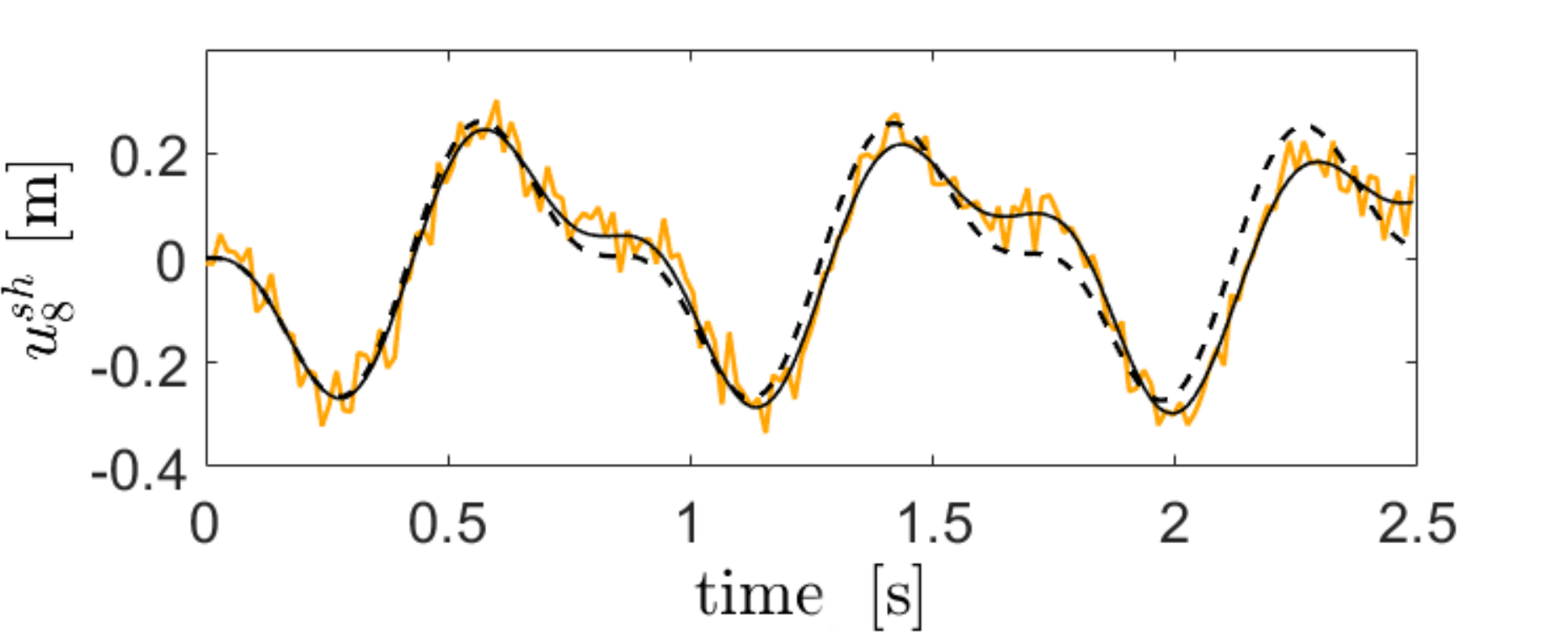}} $~$ \subfloat[{damaged scenario 4\label{fig:comp_44_4_8_flex_enlarg}}]{\includegraphics[scale=0.325]{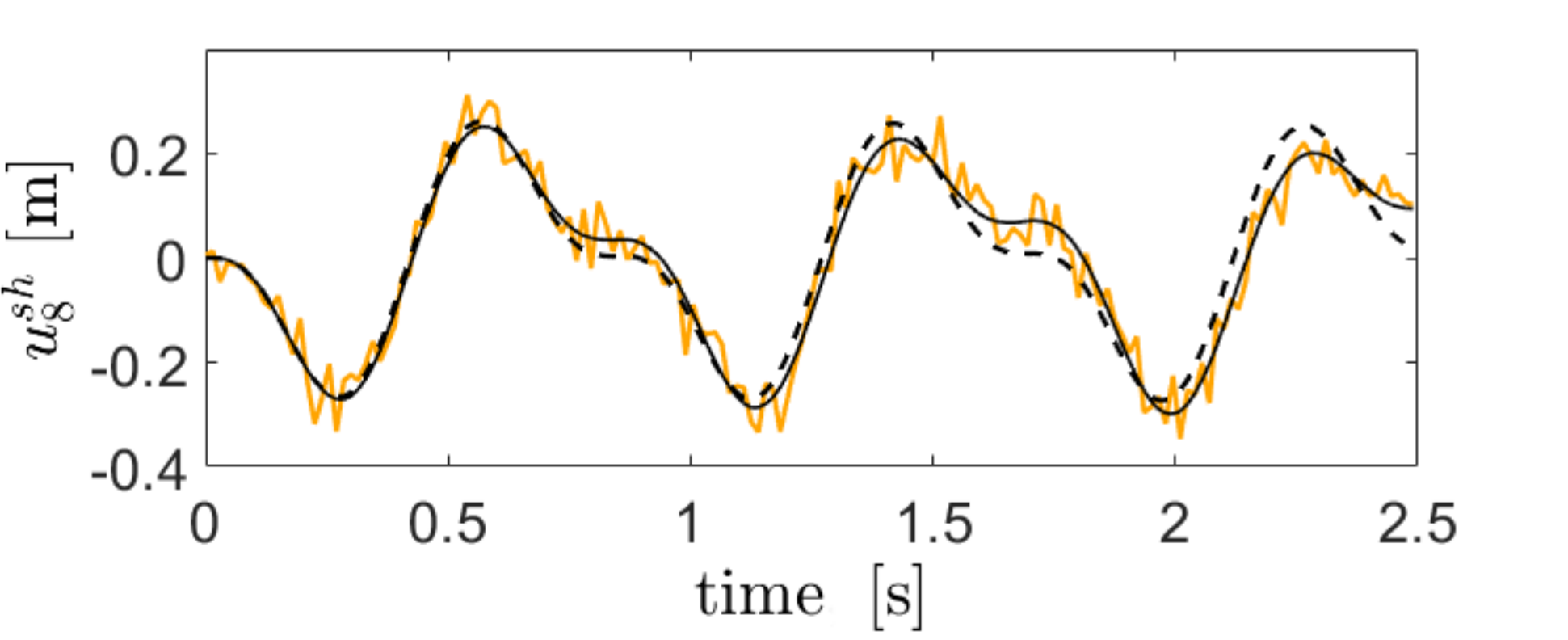}}\vspace{-0.15cm} \\
\subfloat[{damaged scenario 5\label{fig:comp_signal_44_5_8_flex_enlarg}}]{\includegraphics[scale=0.325]{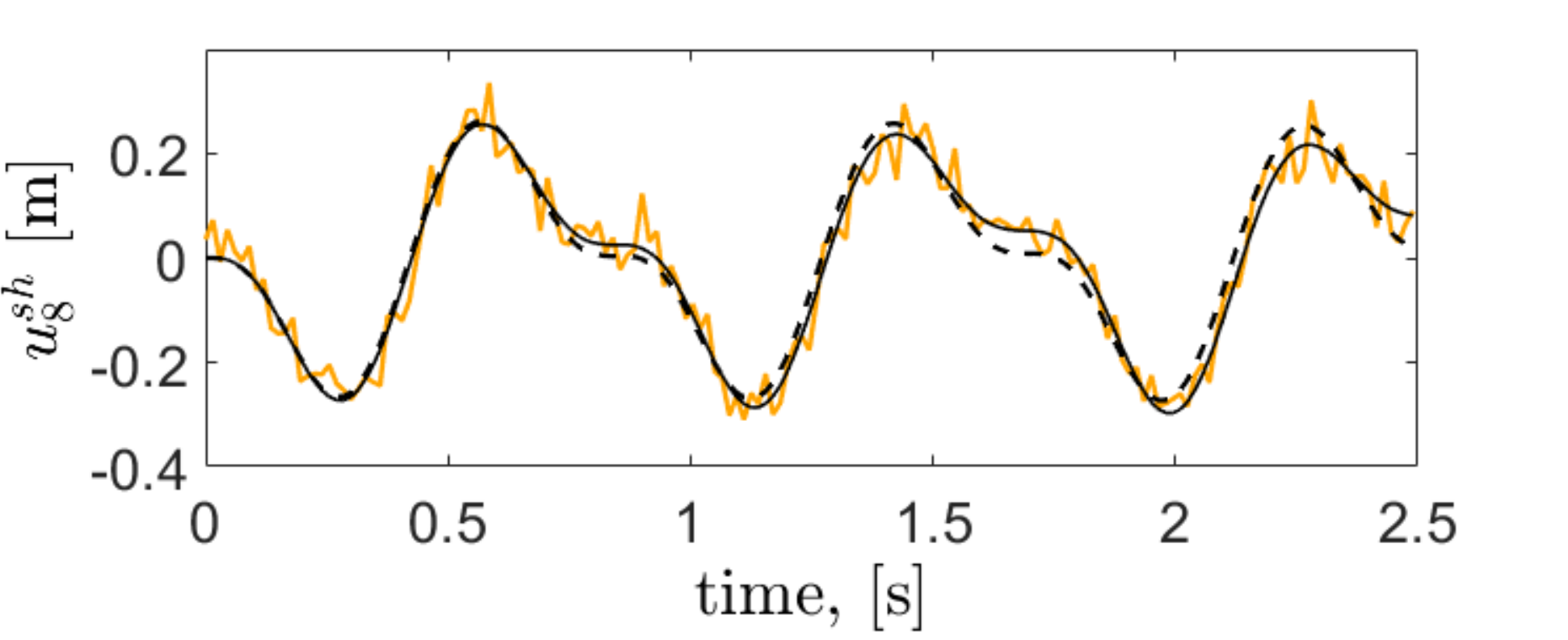}} $~$ \subfloat[{damaged scenario 6\label{fig:comp_signal_44_6_8_flex_enlarg}}]{\includegraphics[scale=0.325]{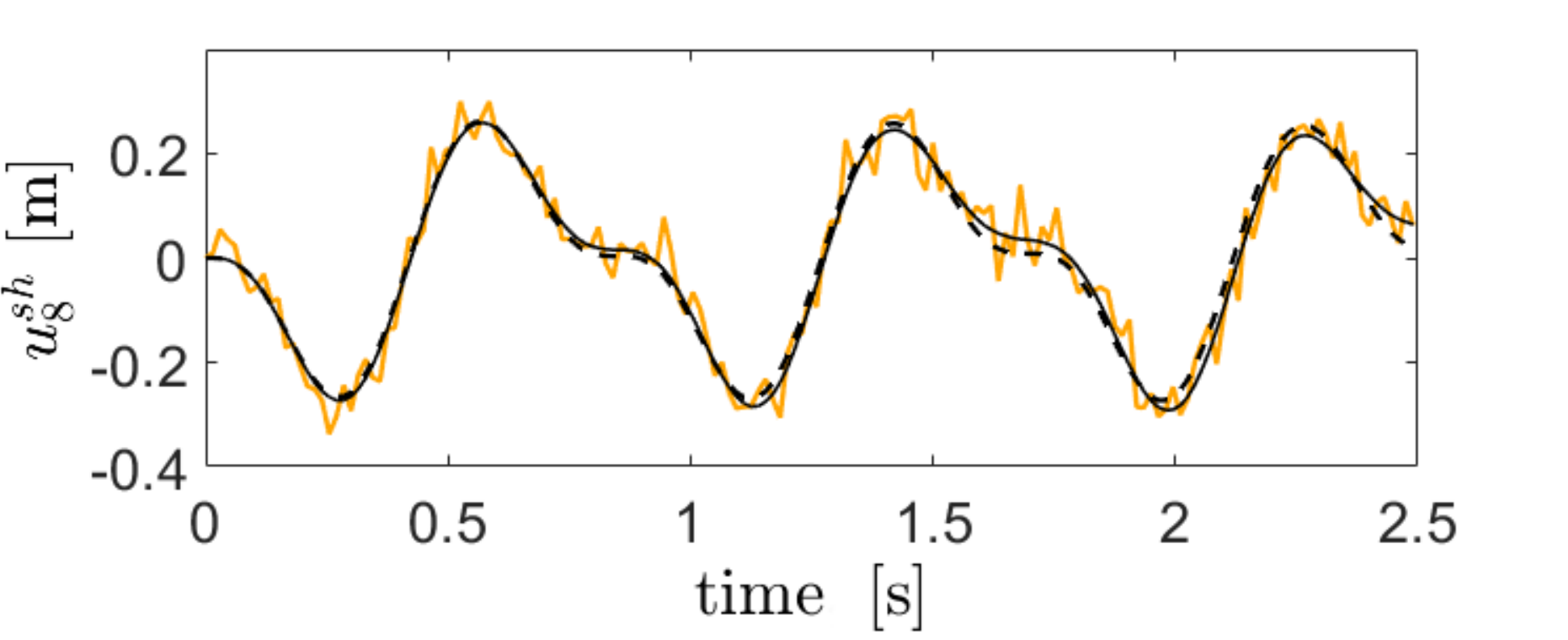}}\vspace{-0.15cm} \\
\subfloat[{damaged scenario 7\label{fig:comp_signal_44_7_8_flex_enlarg}}]{\includegraphics[scale=0.325]{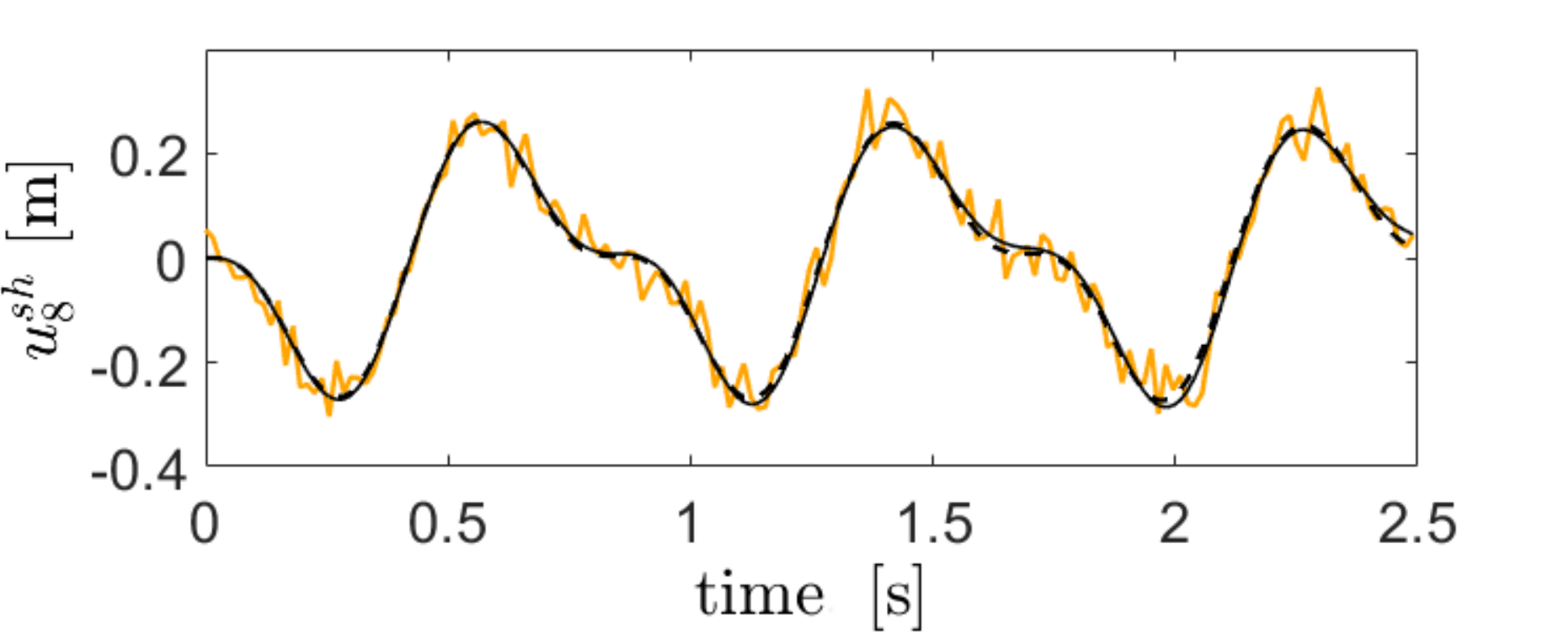}} $~$ \subfloat[{damaged scenario 8\label{fig:comp_signal_44_8_8_flex_enlarg}}]{\includegraphics[scale=0.325]{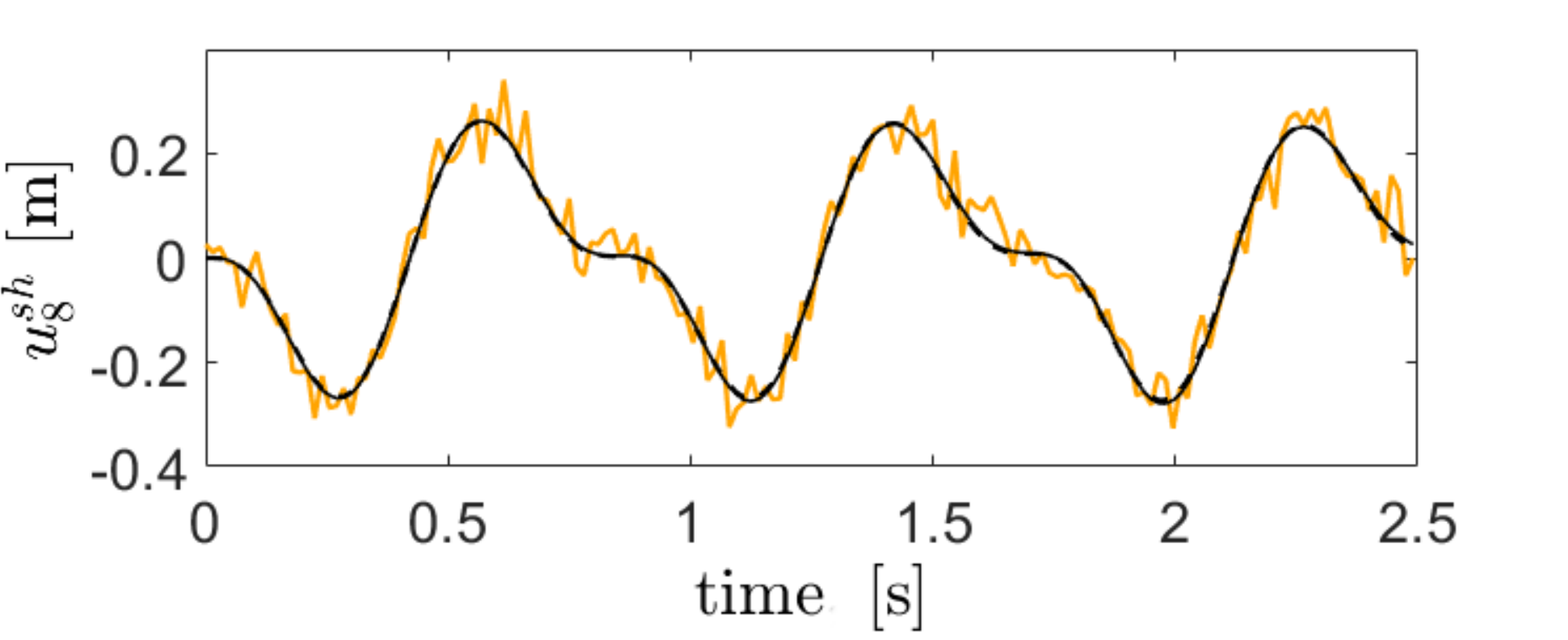}} 
\caption{Example of time evolutions of displacements in the $x$ direction of the $8$-th story for SNR$=10$ dB, with $f_{1,2}^{sh}=\left(14.5, 2.36 \right)$, $\gamma_{1,2}^{sh}=\left(0.025,-0.159\right)$, in the undamaged scenario (\ref{fig:signal_44_0_8_flex_enlarg_2}) and all possible damage scenarios (\ref{fig:comp_signal_44_1_8_flex_enlarg}-\ref{fig:comp_signal_44_8_8_flex_enlarg}). Orange lines represent $\boldsymbol{u}$, whereas black lines stand for $\boldsymbol{r}$, according to Eq.~\eqref{eq:add_noise}. To show the effects of damage on the structural dynamics, the black dotted lines in \ref{fig:comp_signal_44_1_8_flex_enlarg}-\ref{fig:comp_signal_44_8_8_flex_enlarg} report the noise-free structural dynamics related to the undamage scenario.\label{fig:damaged_signal_flex}}
\end{figure}

\begin{figure}[h!]
\captionsetup[subfigure]{justification=centering}
\centering \vspace{-0.185cm}
\subfloat[{undamaged scenario\label{fig:signal_44_0_8_axial_enlarg_2}}]{\includegraphics[scale=0.32]{44_0_8_axial_enlarg.pdf}} \vspace{-0.185cm} \\
\subfloat[{damaged scenario 1\label{fig:comp_signal_44_1_8_axial_enlarg}}]{\includegraphics[scale=0.32]{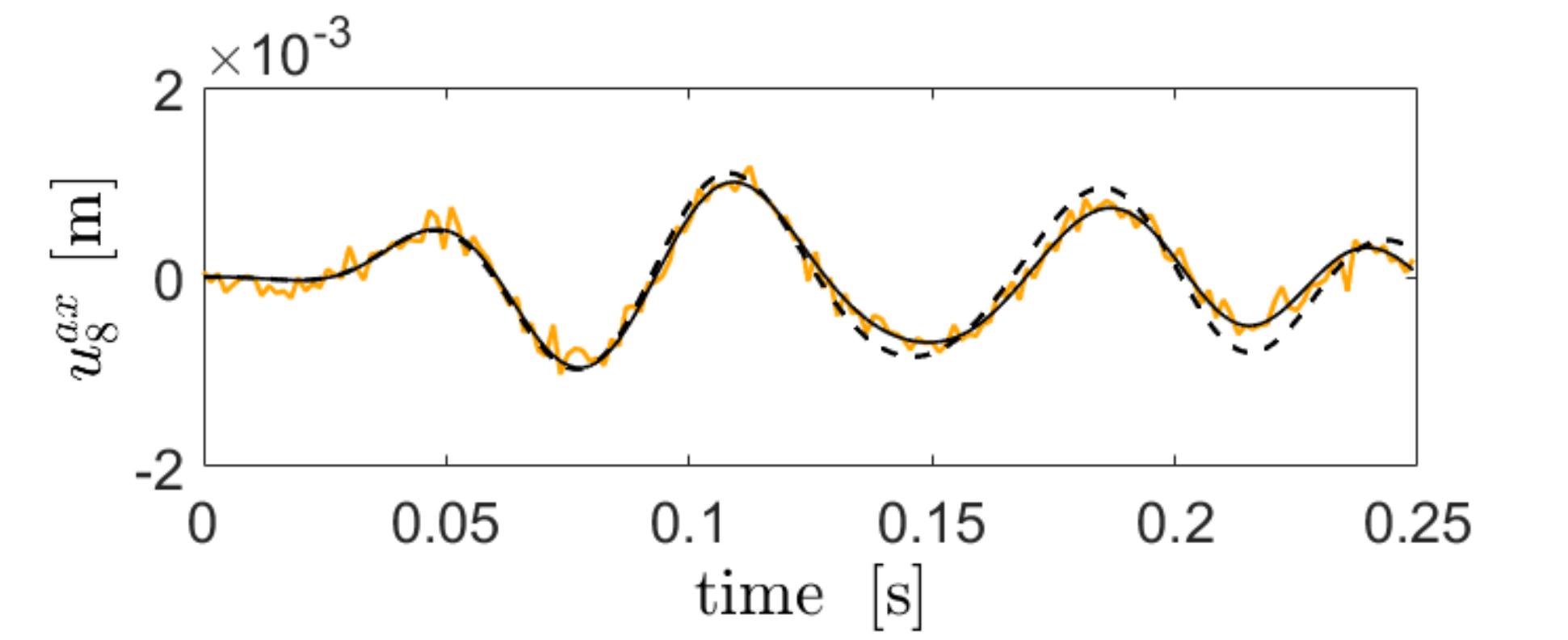}} $~$ \subfloat[{damaged scenario 2\label{fig:comp_signal_44_1_8_axial_enlarg}}]{\includegraphics[scale=0.32]{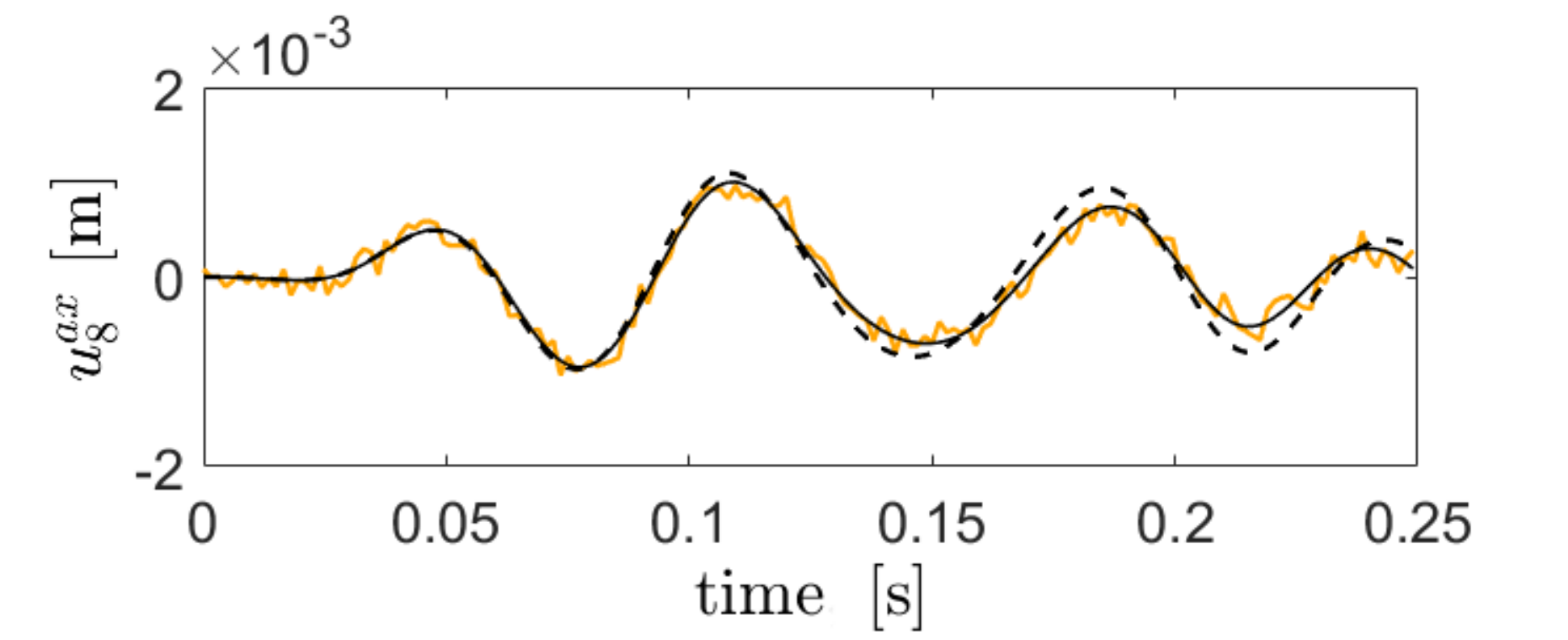}} \vspace{-0.185cm} \\
\subfloat[{damaged scenario 3\label{fig:comp_signal_44_2_8_axial_enlarg}}]{\includegraphics[scale=0.32]{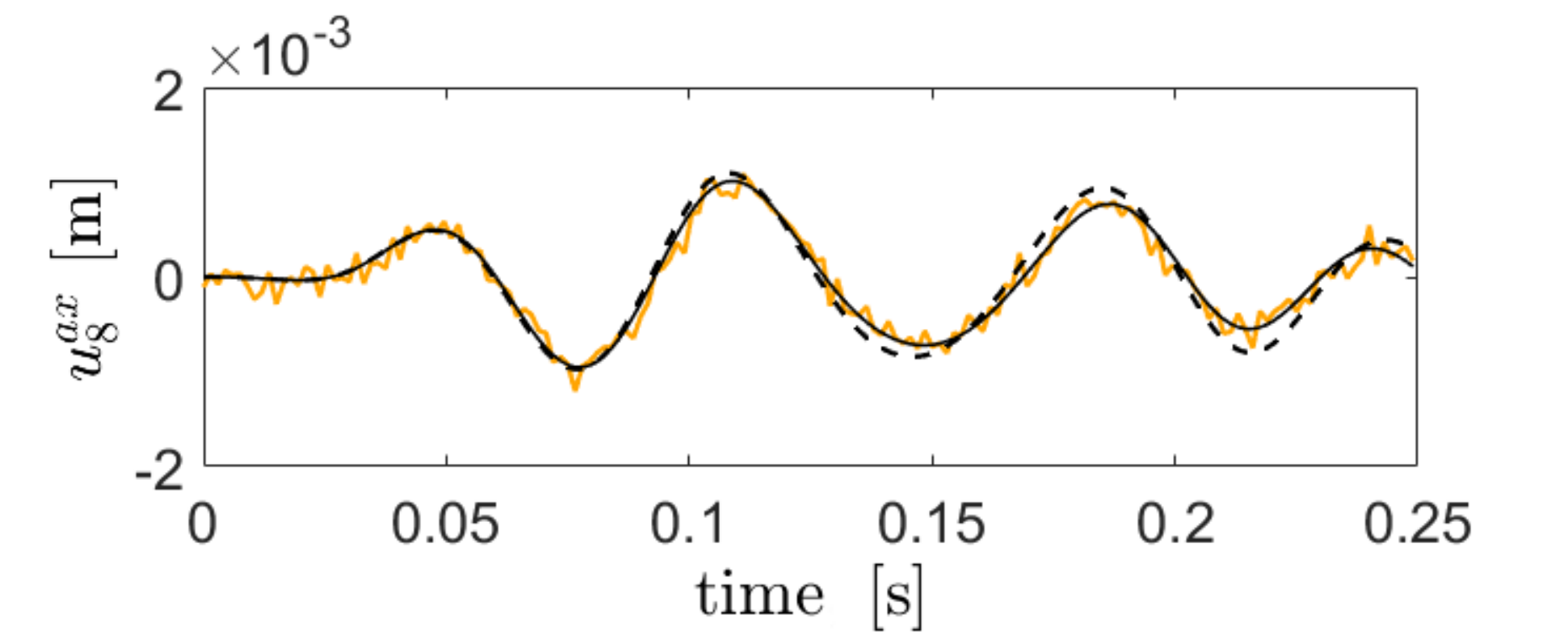}} $~$ \subfloat[{damaged scenario 4\label{fig:comp_44_0_8_axial_enlarg}}]{\includegraphics[scale=0.32]{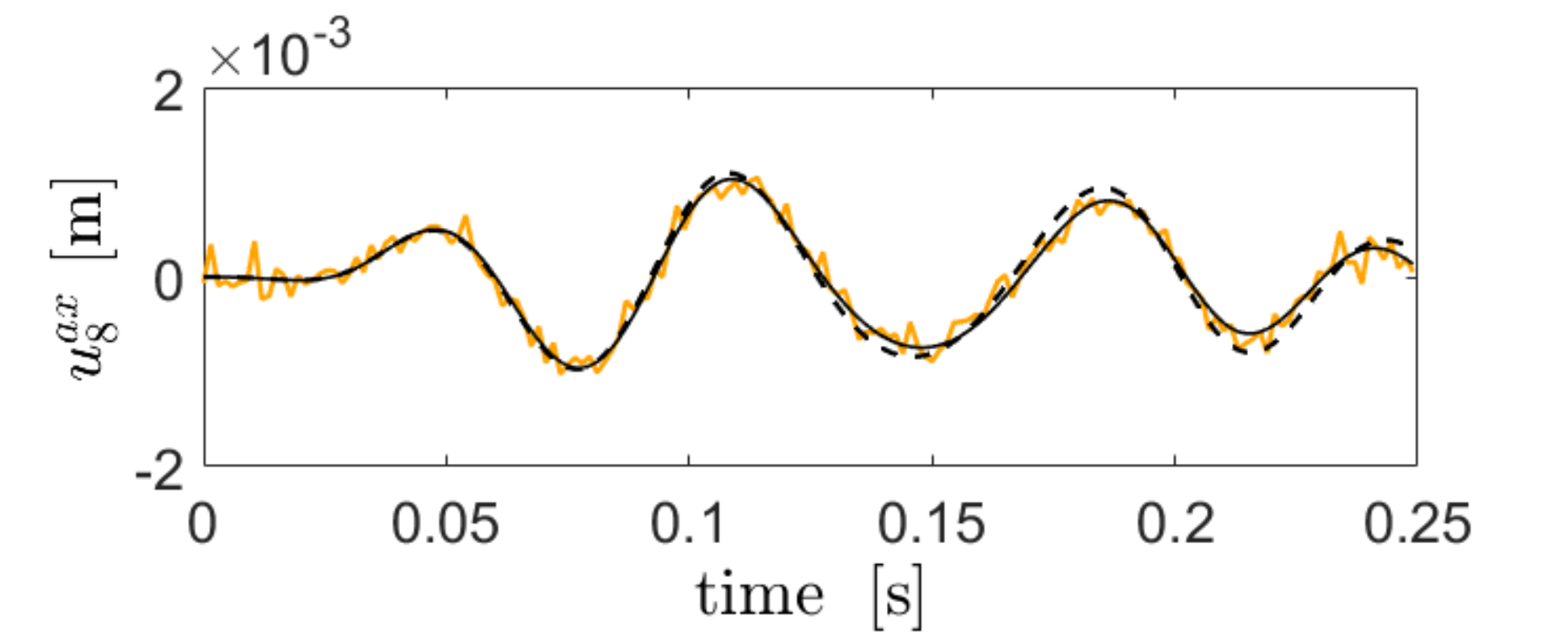}}\vspace{-0.185cm} \\
\subfloat[{damaged scenario 5\label{fig:comp_signal_44_1_8_axial_enlarg}}]{\includegraphics[scale=0.32]{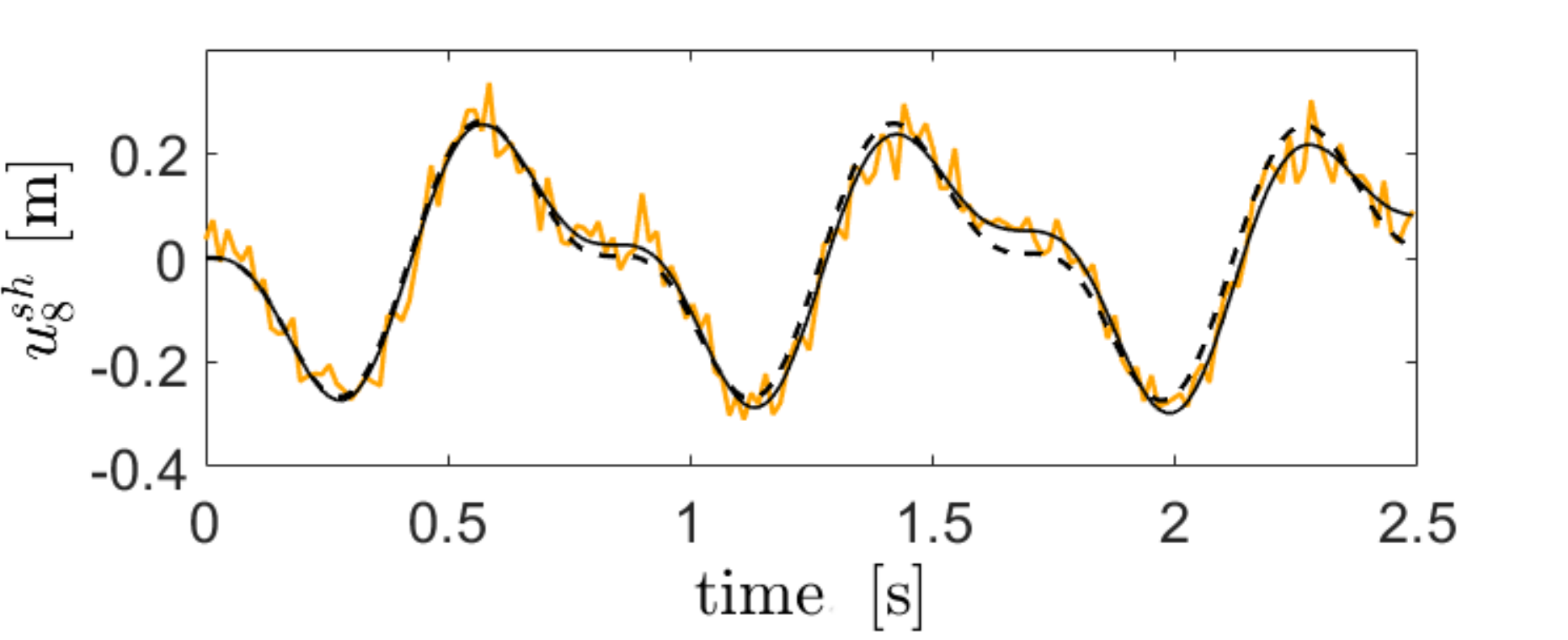}} $~$ \subfloat[{damaged scenario 6\label{fig:comp_signal_44_1_8_axial_enlarg}}]{\includegraphics[scale=0.32]{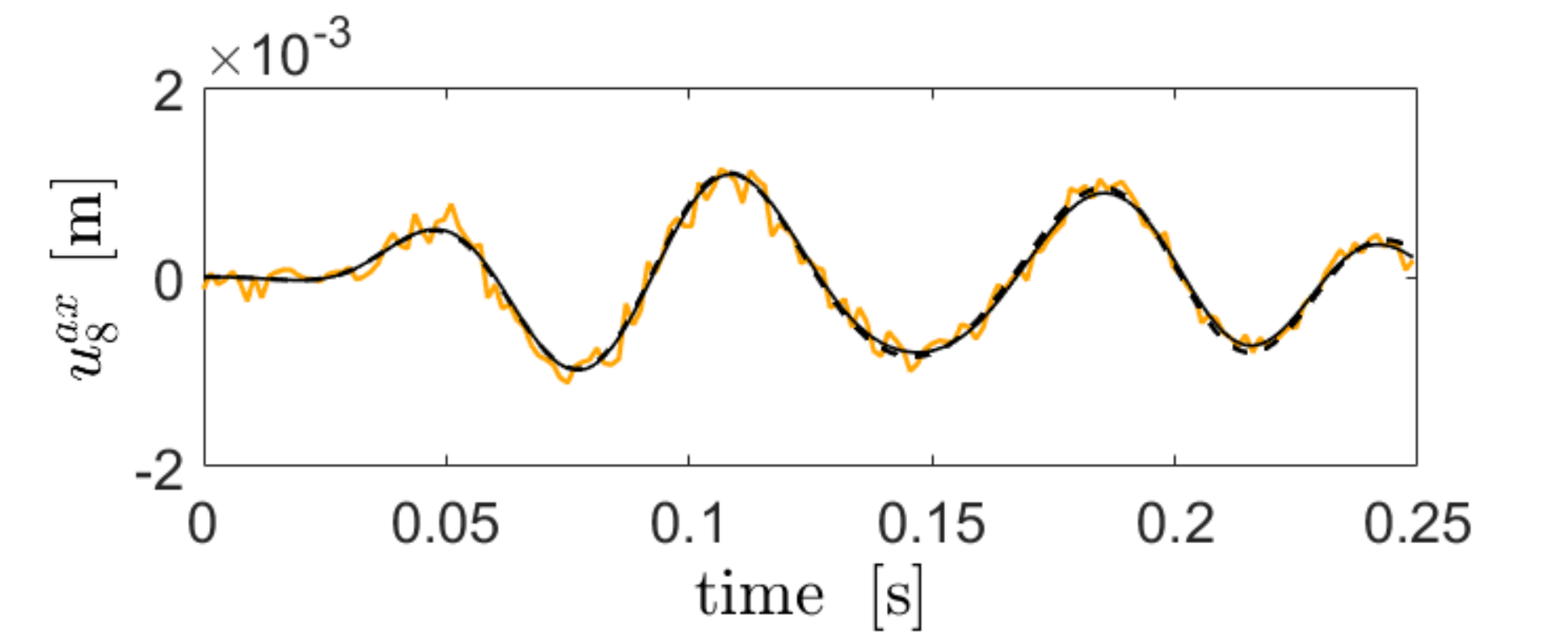}} \vspace{-0.185cm}  \\
\subfloat[{damaged scenario 7\label{fig:comp_signal_44_1_8_axial_enlarg}}]{\includegraphics[scale=0.32]{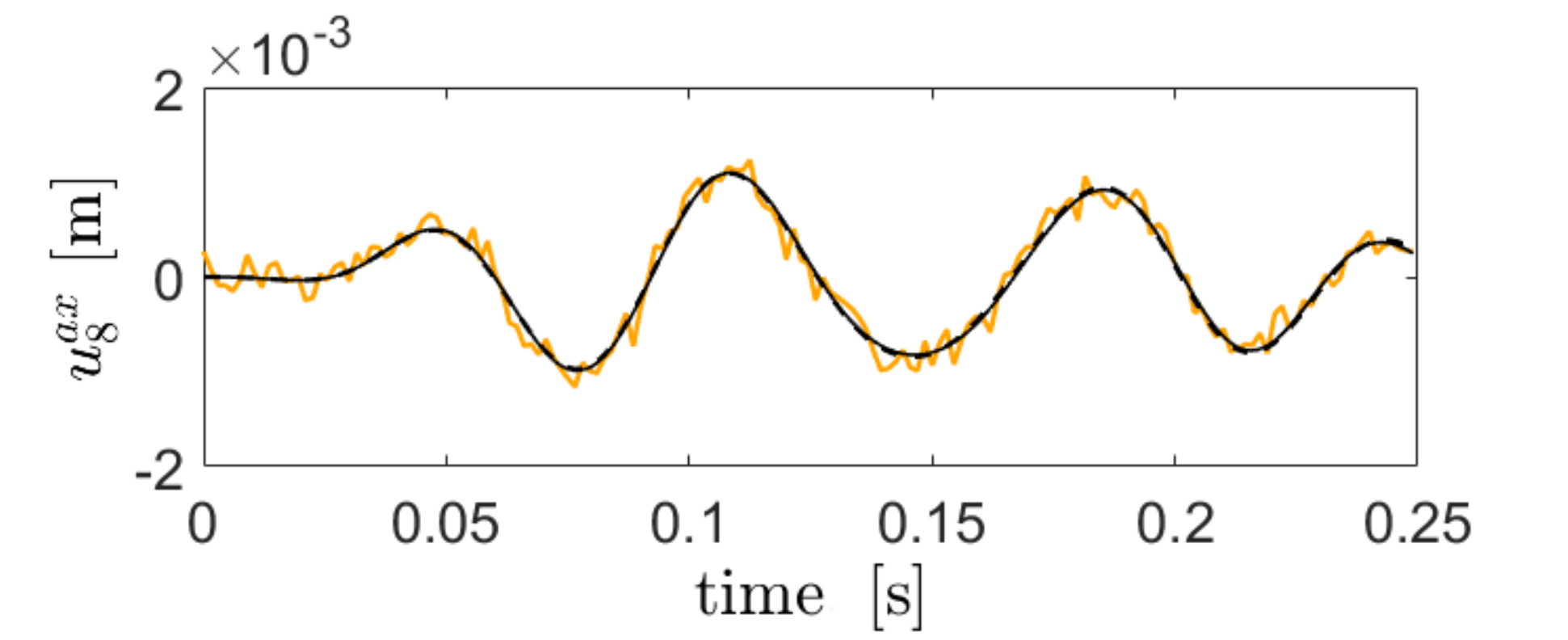}} $~$ \subfloat[{damaged scenario 8\label{fig:comp_signal_44_8_8_axial_enlarg}}]{\includegraphics[scale=0.32]{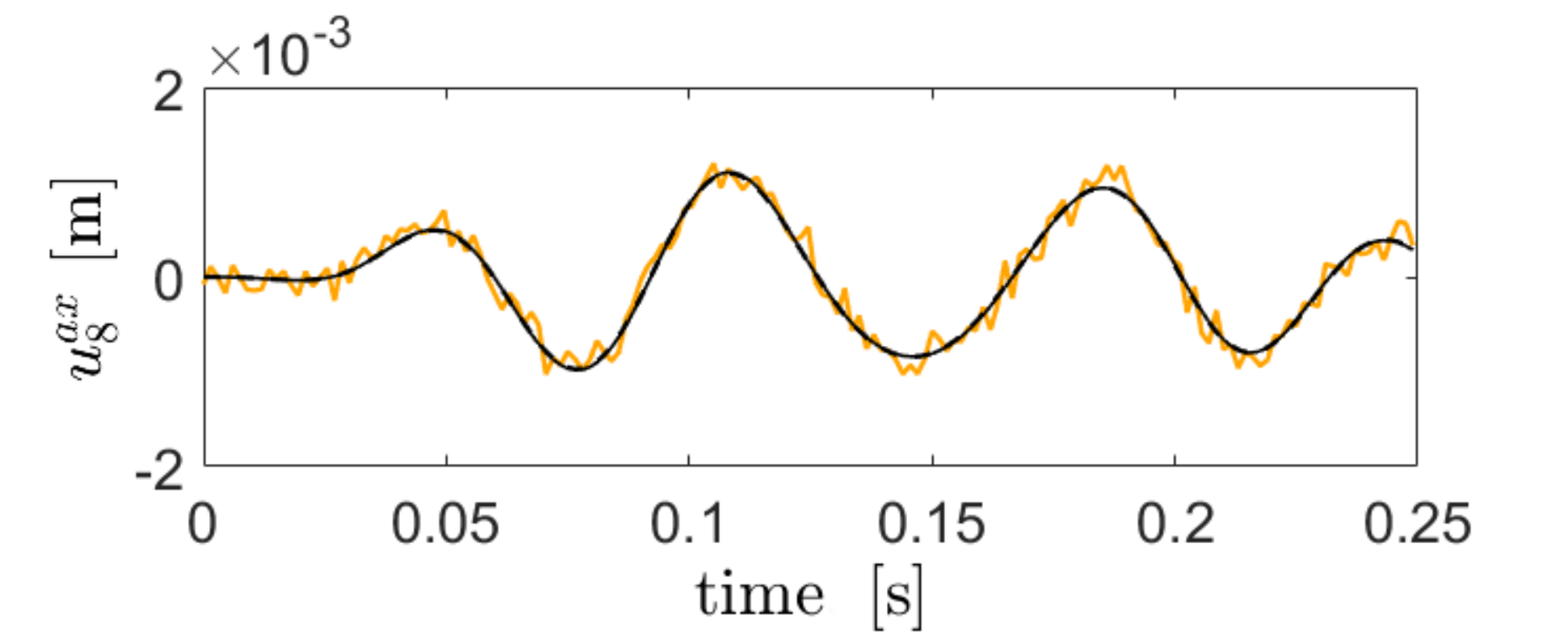}} \vspace{-0.15cm}
\caption{Examples of time evolutions of displacements in the $z$ direction of the $8$-th story for SNR$=10$ dB, with $f_{1,2}^{ax}=\left(15.5, 22.0 \right)$, $\gamma_{1,2}^{ax}=\left(1.133,-1.140\right)$, in the undamaged scenario (\ref{fig:signal_44_0_8_axial_enlarg_2}) and all possible damage scenarios (\ref{fig:comp_signal_44_1_8_axial_enlarg}-\ref{fig:comp_signal_44_8_8_axial_enlarg}). Orange lines represent $\boldsymbol{u}$, whereas black lines stand for $\boldsymbol{r}$, according to Eq.~\eqref{eq:add_noise}. To show the effects of damage on the structural dynamics, the black dotted lines in \ref{fig:comp_signal_44_1_8_axial_enlarg}-\ref{fig:comp_signal_44_8_8_axial_enlarg} report the noise-free structural dynamics related to the undamage scenario.\label{fig:damaged_signal_axial}}
\vspace{-0.25cm}
\end{figure}

To build the dataset required for the NN training, the procedure described so far has been adopted for all the damage scenarios. Fig.~\ref{fig:damaged_signal_flex} and Fig.~\ref{fig:damaged_signal_axial} respectively show the effects of damage on ${u}^{sh}_8$ and ${u}^{ax}_8$, highlighting the sensitivity of this output to the handled damage state. To better highlight this sensitivity, the time evolutions in Fig.~\ref{fig:damaged_signal_flex} and Fig.~\ref{fig:damaged_signal_axial} are provided for $I= \left[0,2.5\right]$s and $I= \left[0,0.25\right]$s only, even though $I= \left[0,10\right]$s and $I=\left[0,1\right]$s for the NN training. Drifts from the responses relevant to the undamaged case can be observed when the damage scenarios refer to the stiffness reduction of the lowest stories; however, it looks nearly impossible, in general, to perform any classification of the damage scenarios without any effectively trained classifier.

\subsubsection{Case 2 (white noise load case)}
\label{sec:load_case_2}

\begin{figure}[h!]
\captionsetup[subfigure]{justification=centering}
\centering
\subfloat[{\label{fig:signal_15_17_time_flex}}]{\includegraphics[scale=0.3]{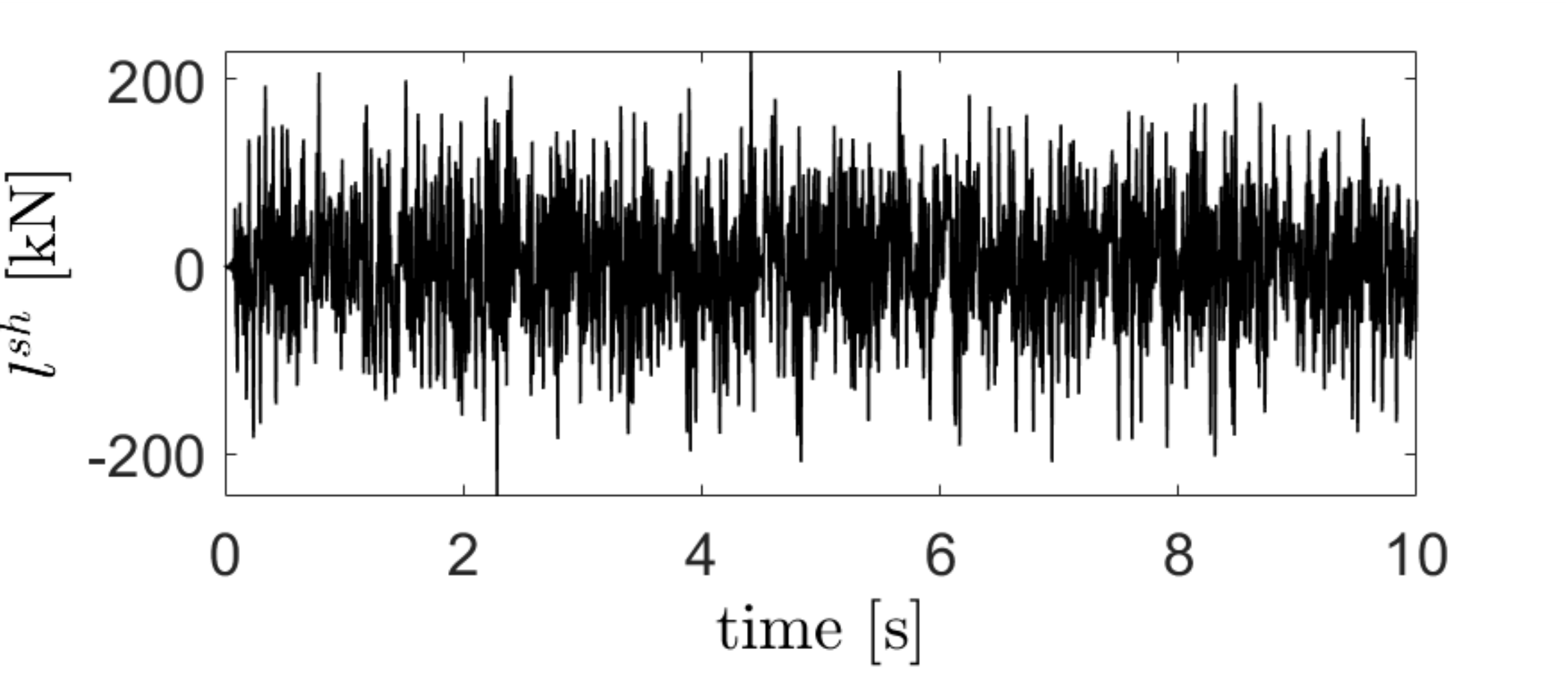}} $~$
\subfloat[{\label{fig:signal_15_17_power_flex}}]{\includegraphics[scale=0.3]{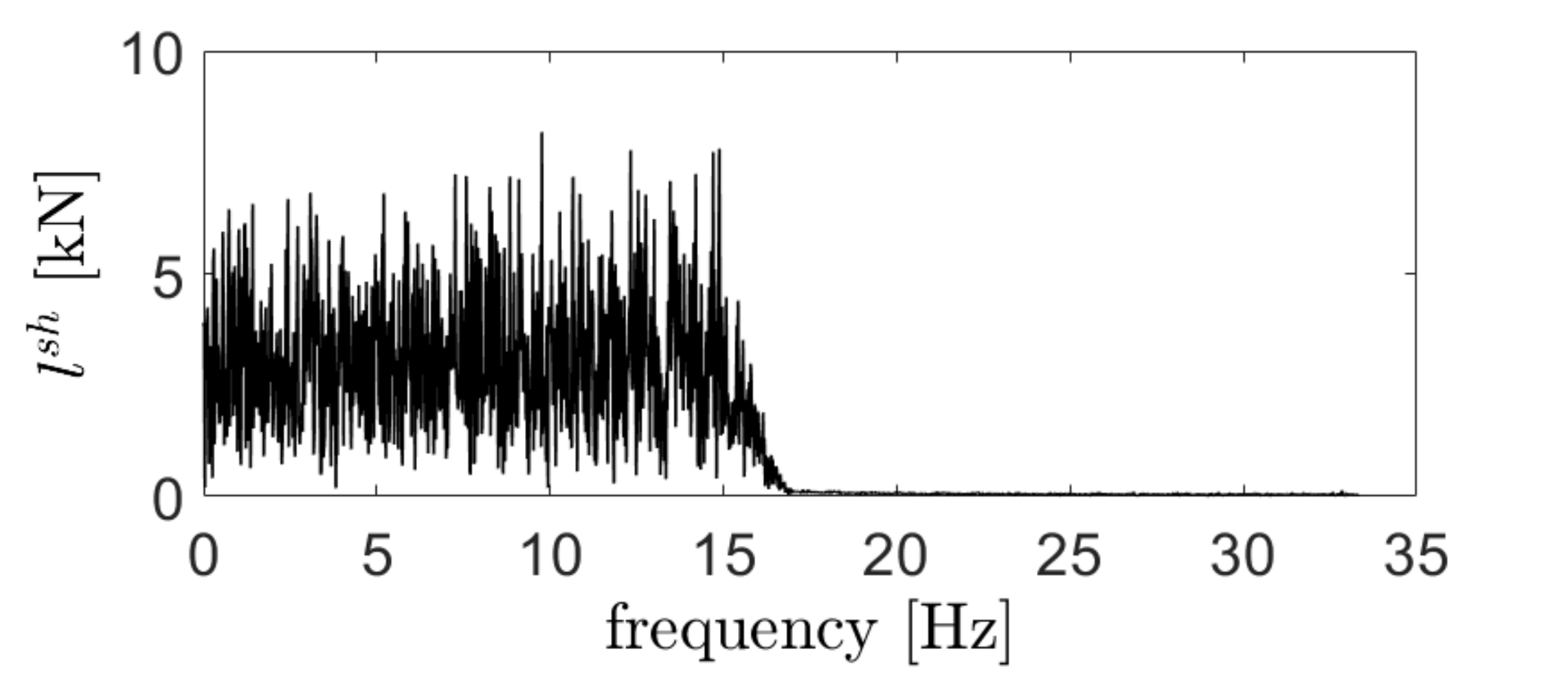}} \\
\subfloat[{\label{fig:signal_15_17_time_axial}}]{\includegraphics[scale=0.3]{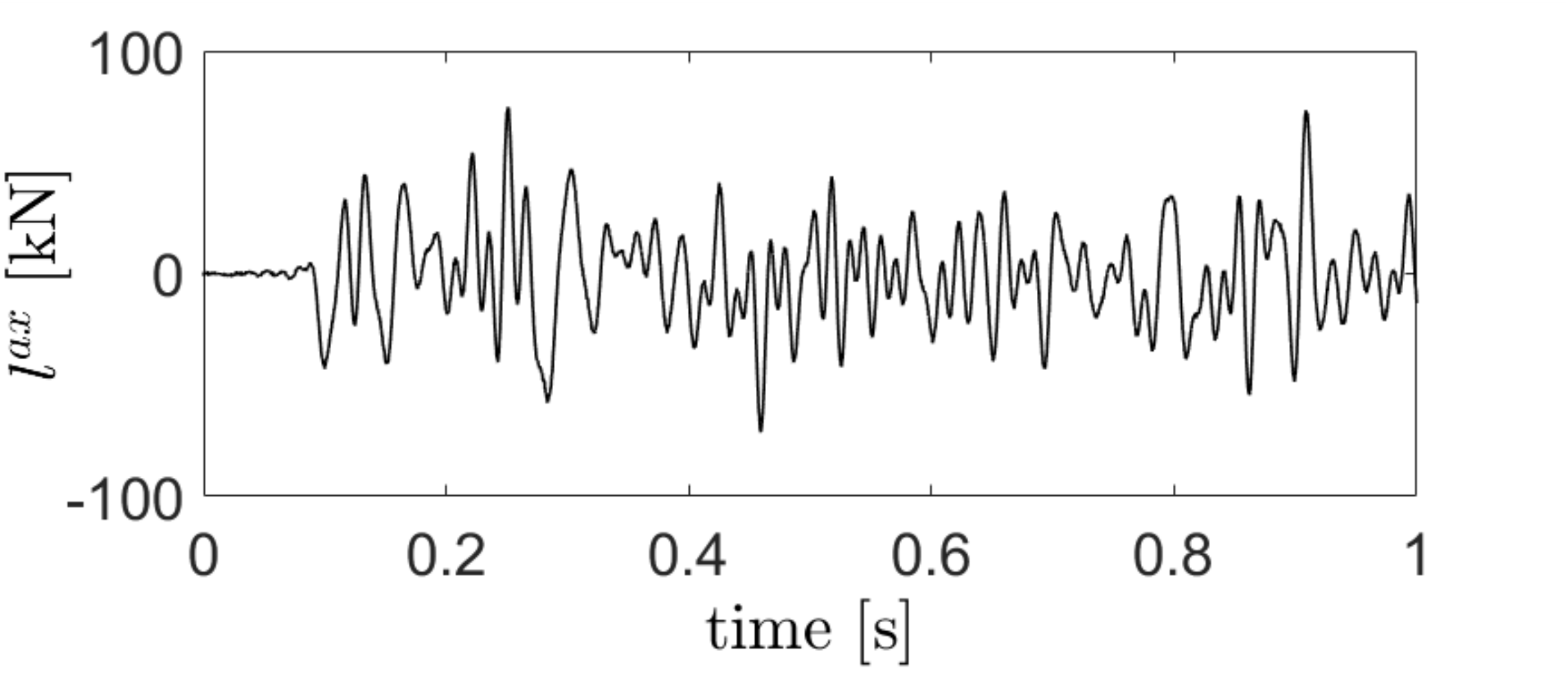}} $~$ \subfloat[{\label{fig:15_17_power_axial}}]{\includegraphics[scale=0.3]{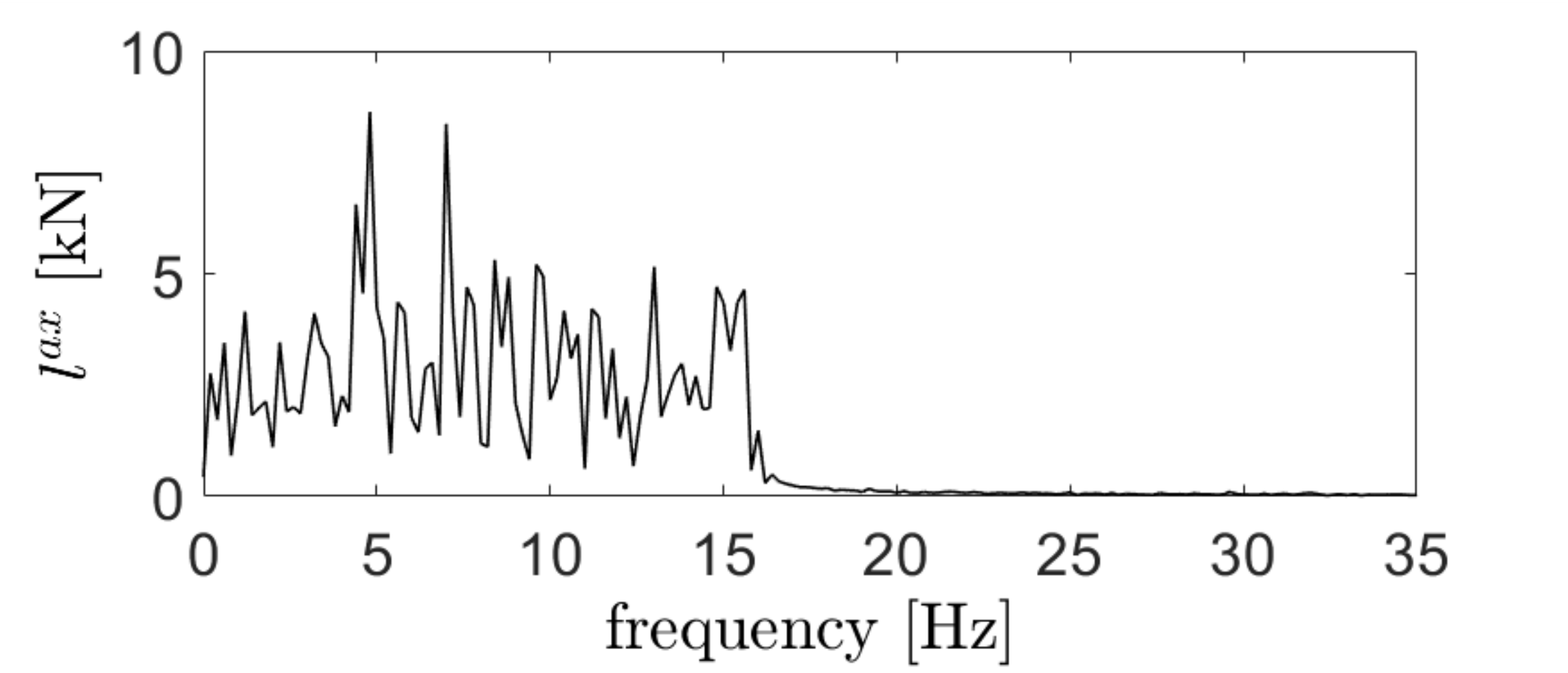}} \\
\caption{White noise load case, $f_{min}=15$ and $f_{max}=17$ Hz. Time evolutions (left column) and Power Spectral Density (right column) of the forces applied to all the building stories in $x$ (first row) and $z$ direction (second row).\label{fig:vibrations_15_17}}
\end{figure}

\begin{figure}[h!]
\captionsetup[subfigure]{justification=centering}
\centering
\subfloat[{\label{fig:signal_5_7_time_flex}}]{\includegraphics[scale=0.3]{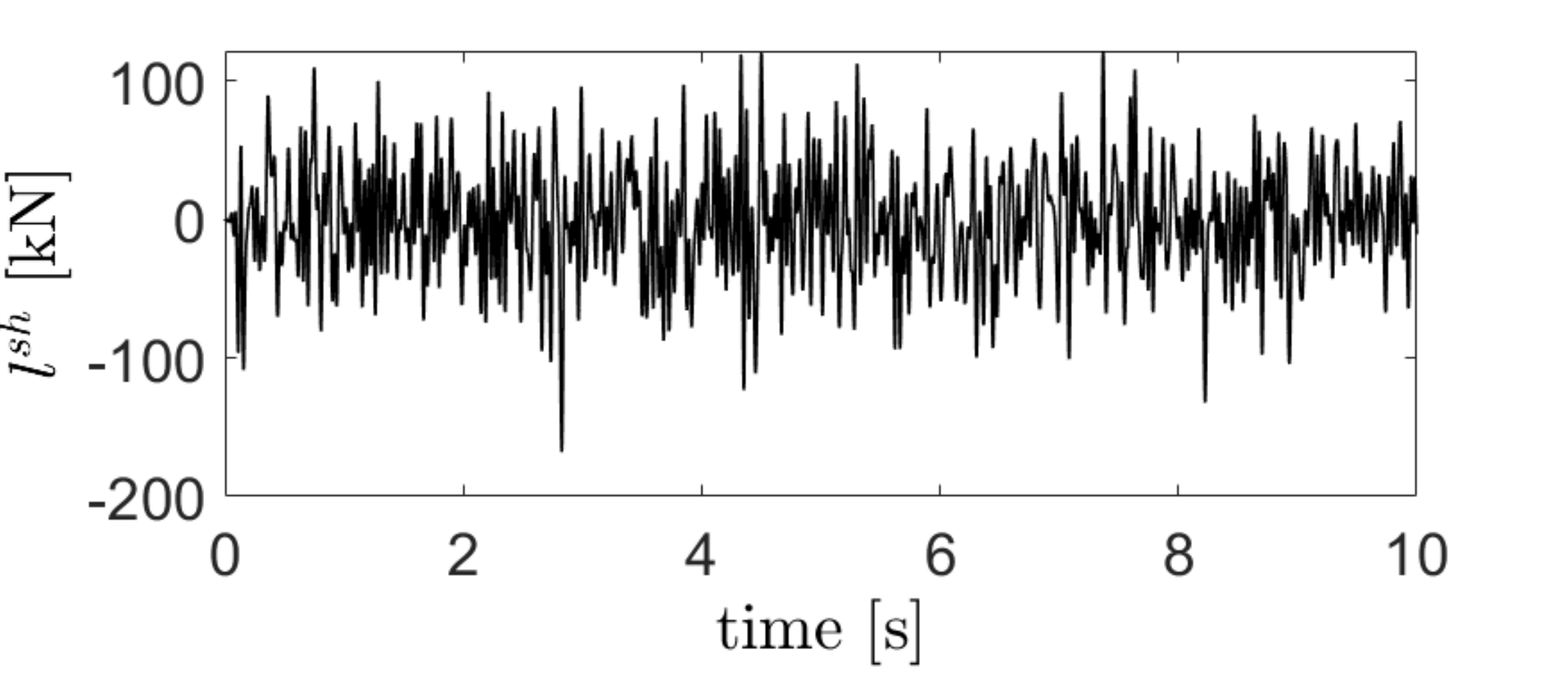}} $~$
\subfloat[{\label{fig:signal_5_7_power_flex}}]{\includegraphics[scale=0.3]{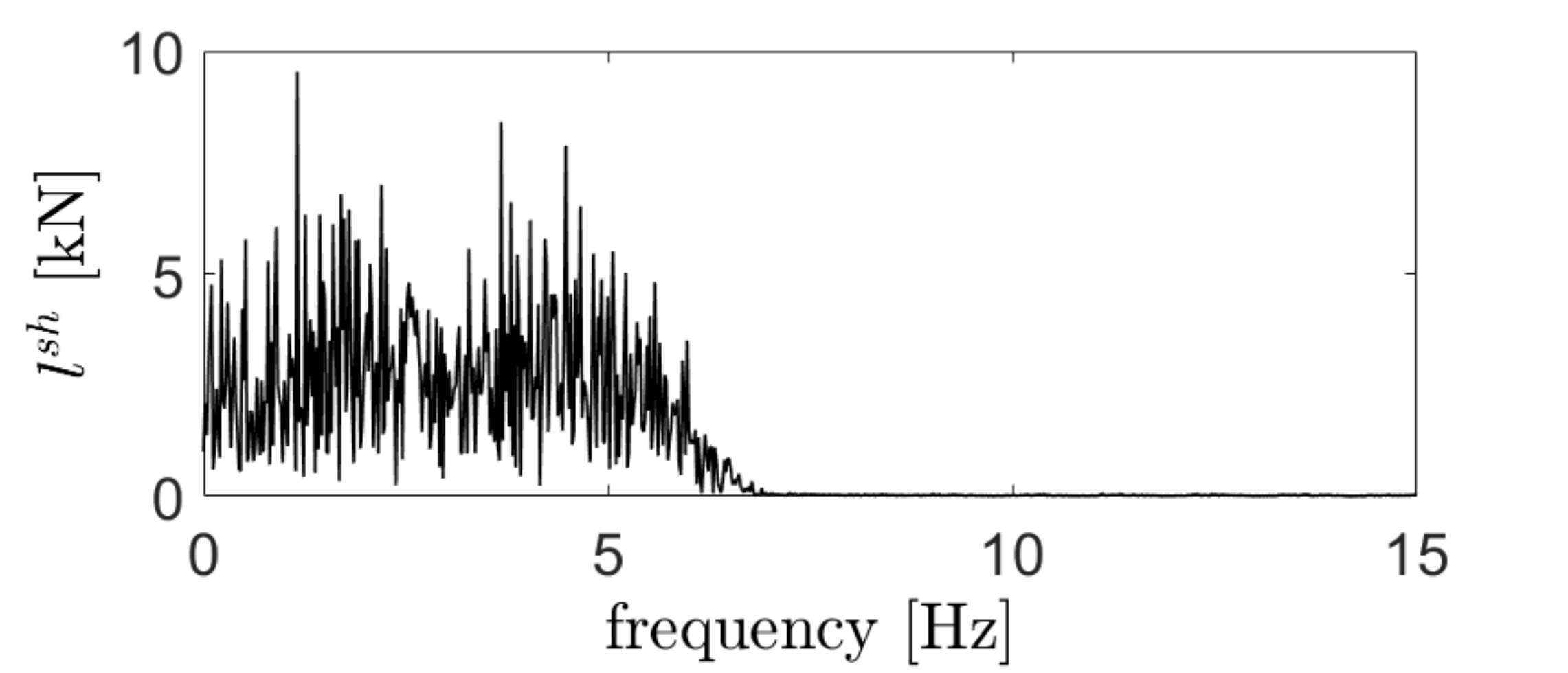}} \\
\subfloat[{\label{fig:signal_5_7_time_axial}}]{\includegraphics[scale=0.3]{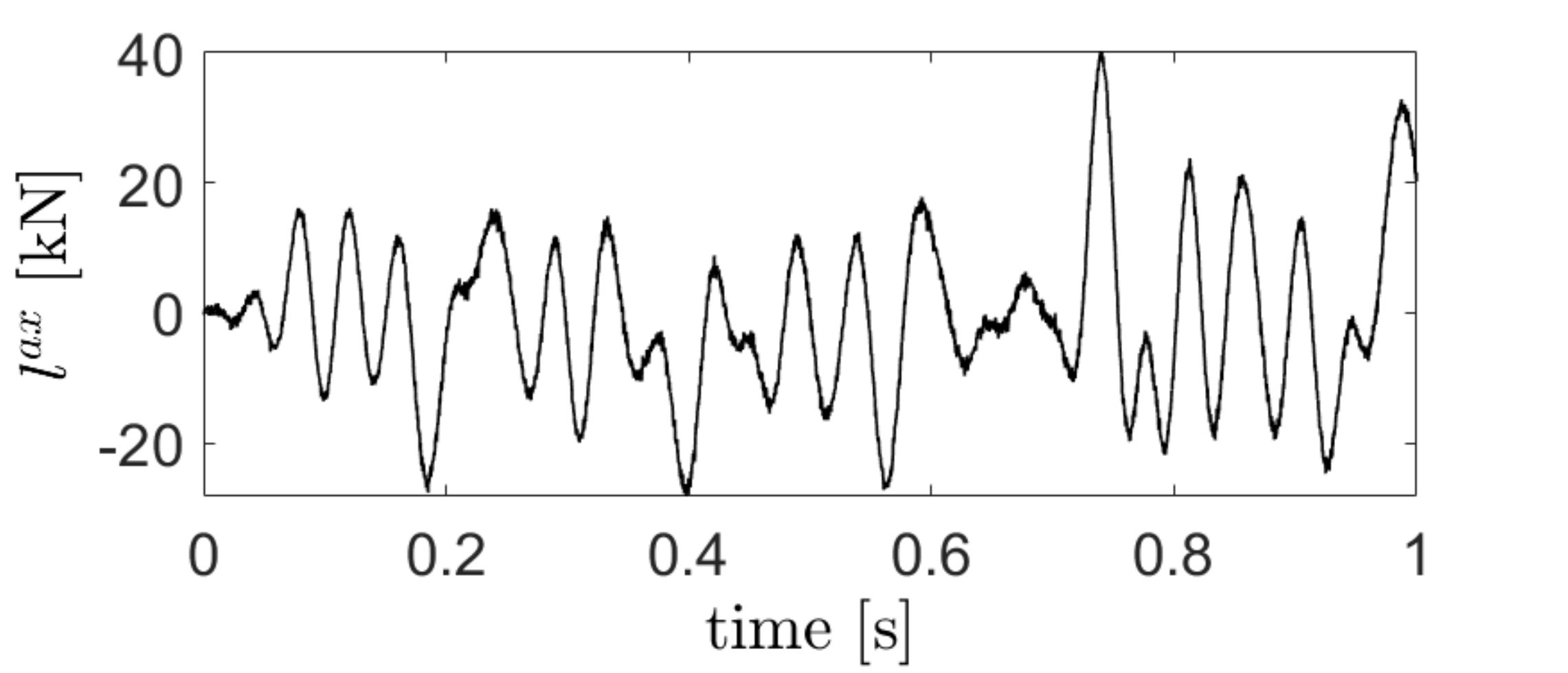}} $~$ \subfloat[{\label{fig:5_7_power_axial}}]{\includegraphics[scale=0.3]{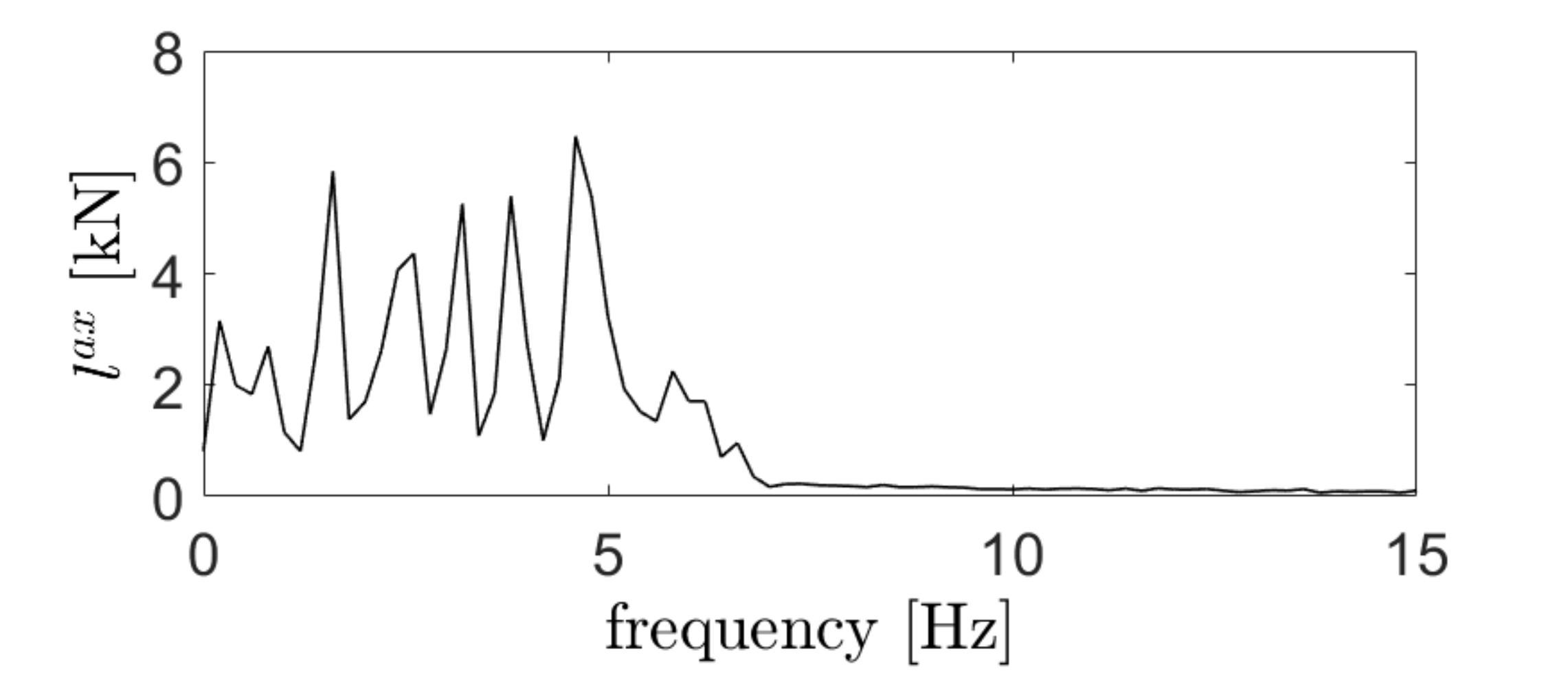}} \\
\caption{White noise load case, $f_{min} = 5$ and $f_{min} = 7$ Hz. Time evolutions (left column) and Power Spectral Density (right column) of the forces enforced to all the building stories in $x$ (first row) and $z$ direction (second row)\label{fig:vibrations_5_7}.}
\end{figure}

In the second load case we have accounted for random vibrations caused e.g. by low-energy seismicity \cite{art:AVT}. The applied loads $\boldsymbol{l} = [\boldsymbol{l}^{sh}, \boldsymbol{l}^{ax}]$, with $i=1,\ldots,8$, at each floor and each time instants are obtained by first sampling out the values from a normal distribution $\mathcal{N}\left( 0, 10^4 \right)$ and then low-pass filtering them with a ``roll-off" set between frequencies $f_{min}$ and $f_{max}$. Two different scenarios have been considered for the frequency range of the applied excitations: $f_{min}=15$ and $f_{max}=17$ Hz; $f_{min}=5$ and $f_{max}=7$ Hz. In the first case all the shear modes and the first axial mode have been excited; in the second case, just the first three shear modes and no axial frequencies have been excited, see Tab.~\ref{tab:eigen}. Fig.~\ref{fig:vibrations_15_17} and Fig.~\ref{fig:vibrations_5_7} respectively provide an overview of the simulated forces for the two cases.

\subsubsection{Dataset composition and NN training}
\label{sec:dataset_composition}

We now detail the construction of the employed datasets and the NN training phase. Each of the two classifiers has been trained on a different dataset, made by instances generated  by evaluating the physics-based model for different loading and damage conditions.  Each instance is made up by $N_0 = 16$  time series recordings of displacements (in two directions, for each of the $8$ floors)  of length $L_0 = 667$. Two global datasets $\mathbb{D}^d$ and $\mathbb{D}^l$ made by  $V=4608$ instances each have been generated, and then split onto a training, a validation and a testing set, thus yielding $\mathbb{D}^d = \mathbb{D}^d_{train} \cup \mathbb{D}^d_{val} \cup \mathbb{D}^d_{test}$ and $\mathbb{D}^l = \mathbb{D}^l_{train} \cup \mathbb{D}^l_{val} \cup \mathbb{D}^d_{test}$, with $V = V^{train} + V^{val} + V^{test}$ in both cases.


For the splitting of the dataset $\mathbb{D}^d$ into training $\mathbb{D}^d_{train}$, validation $\mathbb{D}^d_{val}$ and test $\mathbb{D}^d_{test}$ sets, no specific rules are available, and only some heuristics can be used -- see, e.g., \cite{book:Haykin}. We have thus employed $75\%$ of $V$ to train and validate the NN ($V^{train}$ and $V^{val}$), and the remaining $25\%$ ($V^{test}$) to test it. Within the first subset, $75\%$ of the instances have been in turn allocated for training, and the remaining $25\%$ for validation. The final dataset subdivision then reads: $V^{train}=56.25\% V$, $V^{val}=18.75\% V$, and $V^{test}=25\%V$. The splitting of $\mathbb{D}^l$ has been done identically. The large number of instances employed for validation and test has allowed us to perform a robust assessment of the NN generalization capabilities. This has been done without limiting the information content that can be employed for the NN training; in fact, the dataset dimensions can be arbitrarily enlarged, if necessary, through a synthetic generation of the new instances, still keeping the same subdivision.

During the training, an equal number of instances $V_g^{train} = V^{train} / G$ related to each damage scenario $g=0,\ldots, 8$ (the undamaged case has been considered, too, in addition to the $ G=8$ possible cases of damage) have been provided to the NN, to avoid the construction of a biased dataset $\mathbb{D}^d_{train}$; the same has been done for $\mathbb{D}^l_{train}$. In this way, we indeed prevent the NN to be prone to return the class labels that  have been more frequently presented in the training stage.

There are no specific rules to set $V_g^{train}$ (and, therefore, the overall dimension $V_g = V/G$ of simulated cases for each damage scenario) a priori. Only few theoretical studies provide some recommendations for specific cases, see, e.g., \cite{art:dataset_size}; however, they are not applicable to FCNs. In general, the problem complexity and the employed NN architecture must be taken into account on a case-by-case basis. For this reason, we have evaluated the $\mathcal{G}_{d}$ and $\mathcal{G}_{l}$ classifiers accuracies $A_d$ and $A_{l}$ on the validation set $\mathbb{D}^d_{train}$ and $\mathbb{D}^l_{train}$, and the training time at varying $V^{train}_g$. We have then chosen the best dataset size according to a tradeoff between the two aforementioned indicators, and keeping in mind that the time required to generate a dataset and to train the NN both scale linearly with $V^{train}_g$. 
The $\mathcal{G}_{d}$ classifier accuracy is defined as the ratio $A_d = {V_{\star}^{val}}/{V^{val}}$, where 
$V_{\star}^{val}$ is the number of instances of $\mathbb{D}^l_{val}$ which are correctly classified by $\mathcal{G}_{d}$; the $\mathcal{G}_{l}$ classifier accuracy $A_l$ is defined in a similar way. 

\begin{figure}[h!]
 \centerline{
\includegraphics[scale=0.3]{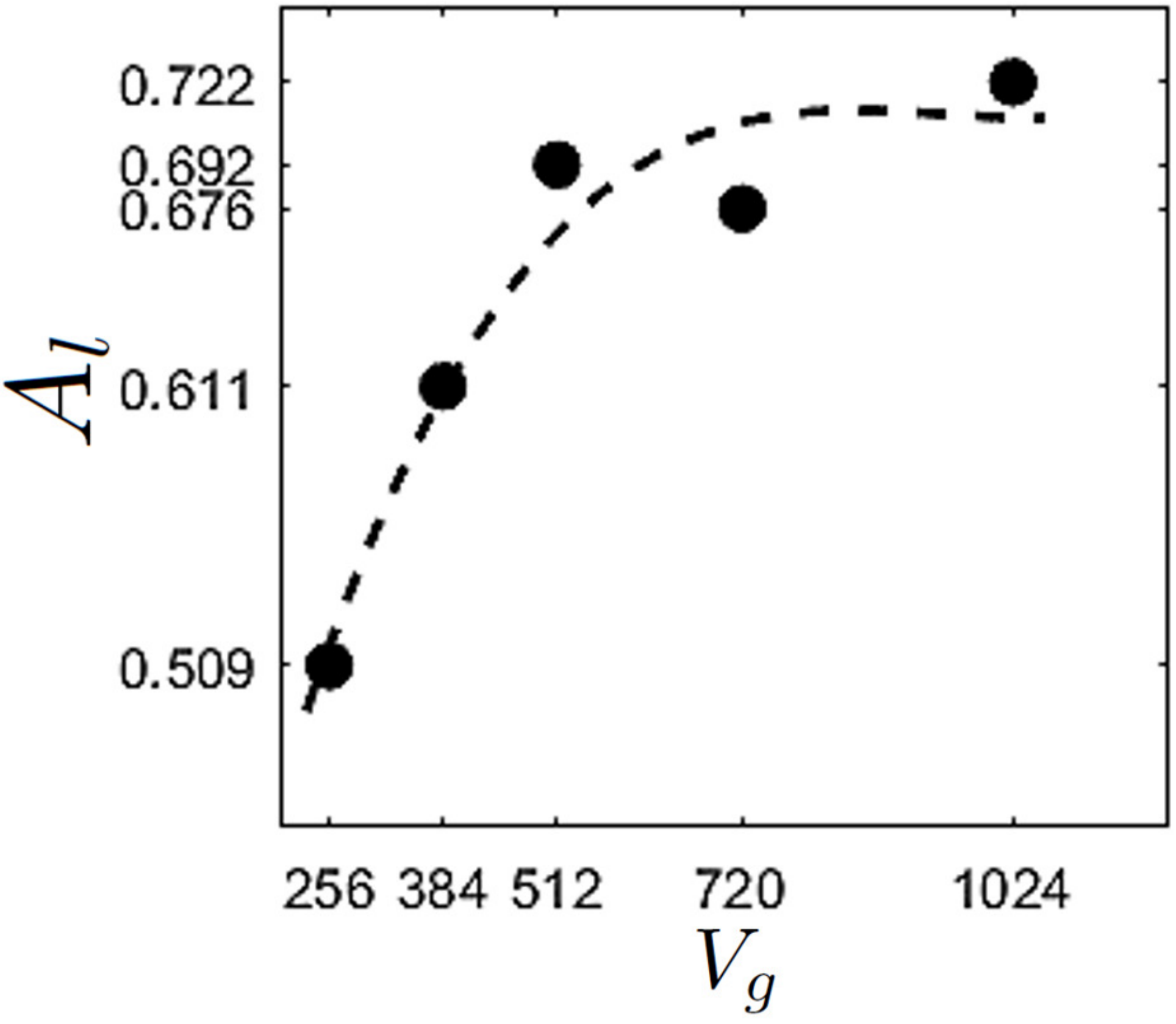}
}
\caption{Damage localization, case 1. Dependence on $V_g$ of the accuracy $A_l$ of the classifier $\mathcal{G}_l$.\label{fig:param_dataset}}
 \vspace{-0.15cm}
\end{figure}

Let us now see how we have determined the overall dataset size $V$ by applying the heuristic approach previously discussed. In Fig.~\ref{fig:param_dataset}, the accuracy $A_{l}$ at varying values of $V_g$ is reported, by considering the local case 1. 

By increasing $V_g$ from $256$ to $384$, $A_{l}$ is highly affected, while a further increasing  yields a smaller gain in accuracy. The non-monothonic variation of $A_{l}$ with respect to $V_g$ is due to the randomness of the procedure, and in particular to the initialization of the weights 
 of the convolutional filters. For the above reasons, we have adopted $V_g = 512$ during the training phase.

Treating the damage detection task for case 1, a total number of $V = 9216$ instances have been generated. Half of the instances refers to the undamaged conditions, half to damaged conditions. Each damage scenario is equally represented ($V_g = 512$ instances each). Regarding instead the damage localization task, $V = 4608$ and $V_g = 512$ (including the undamaged case $g=0$). 

Still adopting the discussed heuristic criterion for the determination of the overall dataset dimension, $V = 4096$ has been used for the damage detection task when the white noise load case is treated. Once again, half of the instances refers to the undamaged conditions, half to the damage condition. Each damage scenario is equally represented ($V_g = 128$ instances each). Regarding the damage localization task, $V = 4608$ and $V_g = 128$ (including the undamaged case $g=0$).

\subsection{Classification outcomes}

We now report the numerical results obtained for the two load cases, and for the two required tasks of damage detection and damage localization. The obtained classification outcomes are affected by the NN architecture, either with one  or two convolutional branches, depending on whether the horizontal and vertical sensing are both considered or not. In particular, when treating the damage localization task in presence of the white noise load condition, we will also try to assess the impact of each input channel $\mathcal{F}^{n}_0$ on the overall NN accuracy.

Useful indications about the goodness of the training can be derived from the behavior of the loss functions $J_d\left(\boldsymbol{Y},\boldsymbol{p}\right)$ and $J_l\left(\boldsymbol{Y},\boldsymbol{p}\right)$ -- see Eq.~\eqref{eq:cross_entropy} -- of $\mathcal{G}_d$ and $\mathcal{G}_l$, and of the accuracies $A_d$ and $A_l$ on the training and validation sets ($\mathbb{D}^t_{train}$ and $\mathbb{D}^t_{val}$ for $\mathcal{G}_d$; $\mathbb{D}_{train}$ and $\mathbb{D}_{val}$ for $\mathcal{G}_l$) as a function of the number of iterations. This latter depends on both the number of epochs and the minibatch size chosen for the training\footnote{\footnotesize In other words, if the dataset is composed by $100$ instances and a minibatch size of $10$ instances is adopted, after the first epoch the iteration number is equal to $10$.}.

To evaluate the NN performances, the adopted indices are still $A_d$ and $A_l$, yet evaluated on $\mathbb{D}_{test}^d$ and $\mathbb{D}^l_{test}$. These indices are always compared against the ones produced by a random guess, equal to $0.5$ for $\mathcal{G}_d$, and to $1/9=0.111$ for $\mathcal{G}_{l}$. For the damage localization case, the misclassification is measured by a confusion matrix in which the rows correspond to the target classes and the columns to the NN predictions.

\subsubsection{Damage detection and localization in case 1 - sinusoidal load case}
\label{sec:dam_det_sin_load}

In Tab.~\ref{tab:dam_det_sin_load_tab} the accuracies $A_d$ of $\mathcal{G}_d$ on $\mathbb{D}^d_{test}$ for the two considered noise levels (SNR$=15$ dB and SNR$=10$ dB) are reported. NN architectures with both one and two convolutional branches have been tested.
\begin{table}[h!]
\centering
\begin{tabular}{*3c}
\hline
\parbox{2 cm}{SNR (dB)} & \parbox{3cm}{\centering $\lbrace\mathcal{F}_{*} \rbrace$} & $A_{d}$ \\
\hline
$15$ & $\{\boldsymbol{u}^{sh}_i\}_{i=1}^8$ & $0.814$ \\
$15$ & $\{\boldsymbol{u}^{ax}_i\}_{i=1}^8$ & $0.850$ \\
$15$ & $\{\boldsymbol{u}^{sh}_i\}_{i=1}^8$ and $\{\boldsymbol{u}^{ax}_i\}_{i=1}^8$ & $0.879$ \\
$10$ & $\{\boldsymbol{u}^{sh}_i\}_{i=1}^8$ & $0.768$ \\
$10$ & $\{\boldsymbol{u}^{ax}_i \}_{i=1}^8$ & $0.775$ \\
$10$ & $\{\boldsymbol{u}^{sh}_i\}_{i=1}^8$ and $\{\boldsymbol{u}^{ax}_i\}_{i=1}^8$ & $0.765$ \\
\bottomrule
\end{tabular}
\caption{Damage detection, case 1. Accuracy $A_{d}$ of the classifier $\mathcal{G}_d$ evaluated on $\mathbb{D}^d_{test}$. \label{tab:dam_det_sin_load_tab}}
\end{table}

The classifier $\mathcal{G}_d$ reaches $A_d = 0.879$ for SNR$=15$ dB and $A_d = 0.775$ for SNR$=10$ dB. These outcomes obtained on high-noise datasets show the potentialities of the proposed approach in view of facing real engineering applications. Indeed, noise effect is a principal concern especially when pervasive and low-cost microelectromechanical systems (MEMS) sensor networks are employed \cite{art:low-cost_MEMS}, so that the possibility to handle it through FCNs may enhance the application of MEMS networks. Moreover, thanks to our procedure, we have been able to avoid the data pre-processing required by any ML approach based on problem specific features.

\begin{figure}[h!]
\captionsetup[subfigure]{justification=centering}
\centering
\subfloat[SNR=15 dB\label{fig:loss_sin_det_15_dB_34_1}]{\includegraphics[scale=0.5]{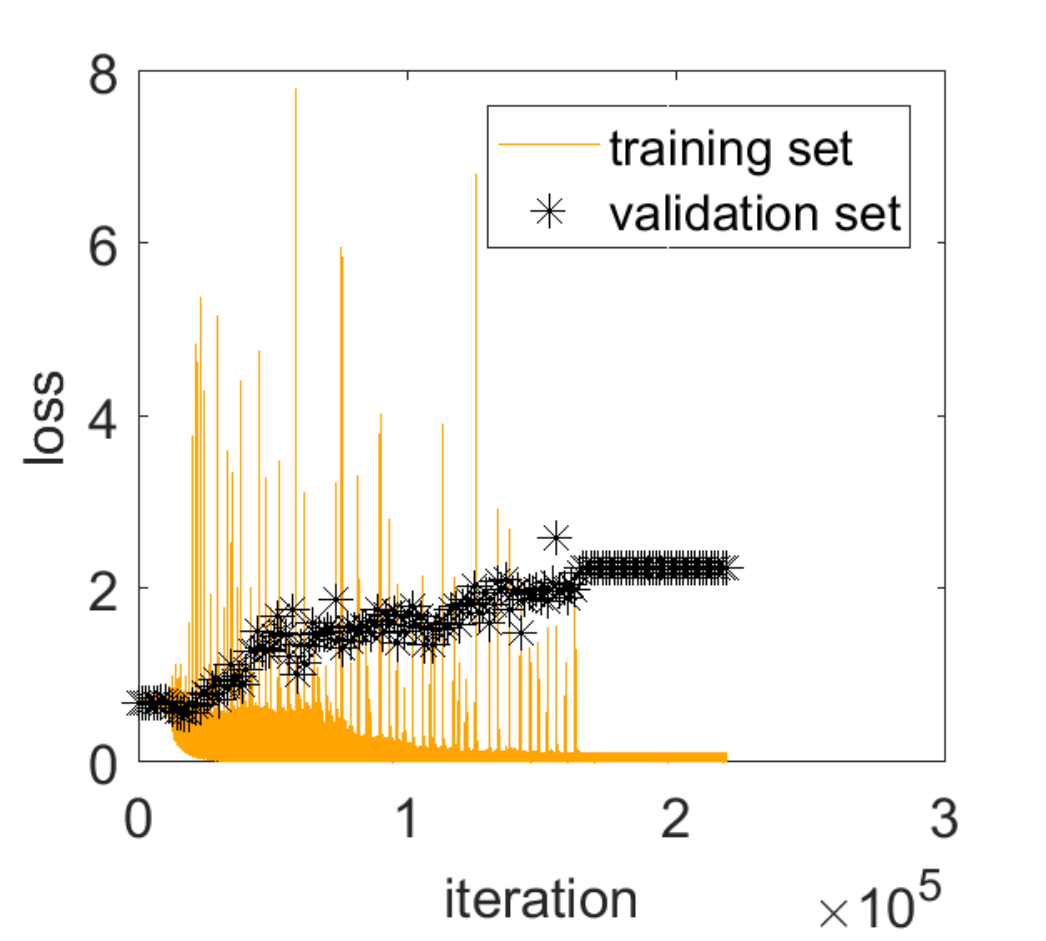}}
\subfloat[SNR=15 dB\label{fig:acc_sin_det_15_dB_34_1}]{\includegraphics[scale=0.5]{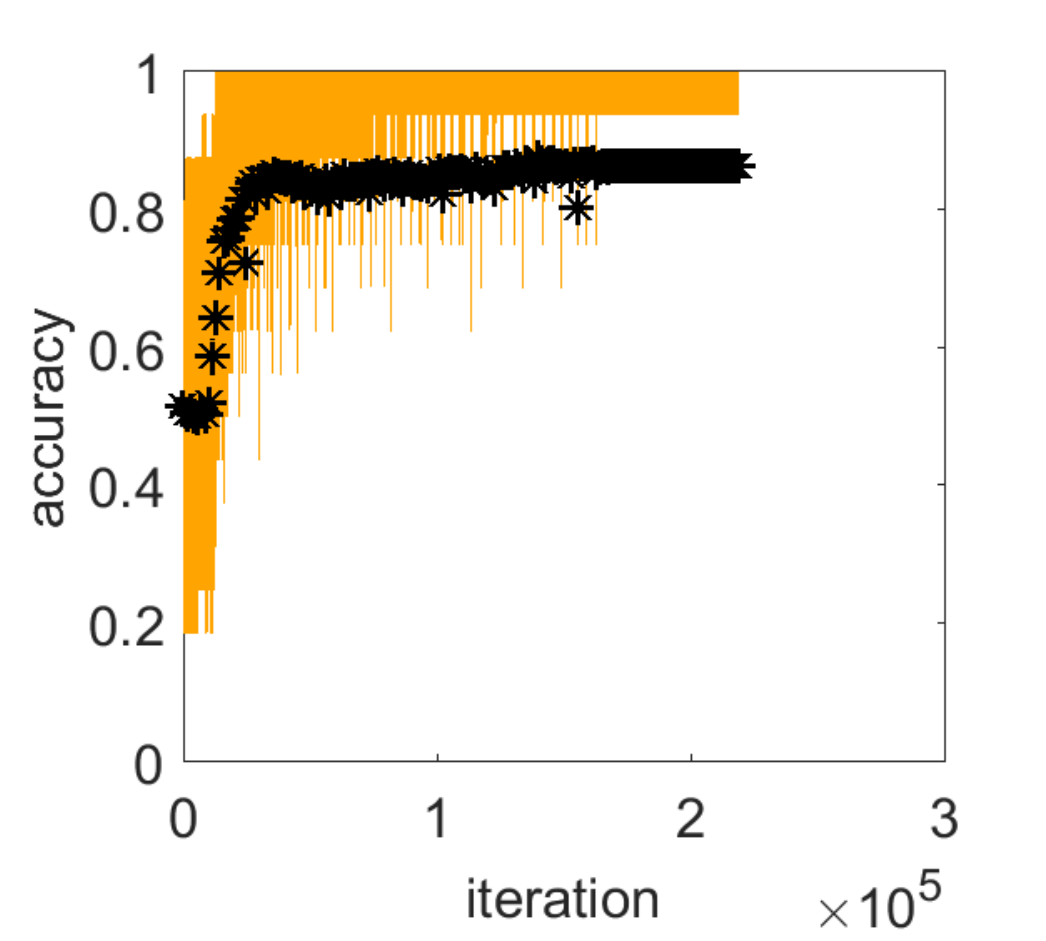}} \vspace{-0.2cm}  \\
\subfloat[SNR=10 dB\label{fig:loss_sin_det_10_dB_33_1}]{\includegraphics[scale=0.5]{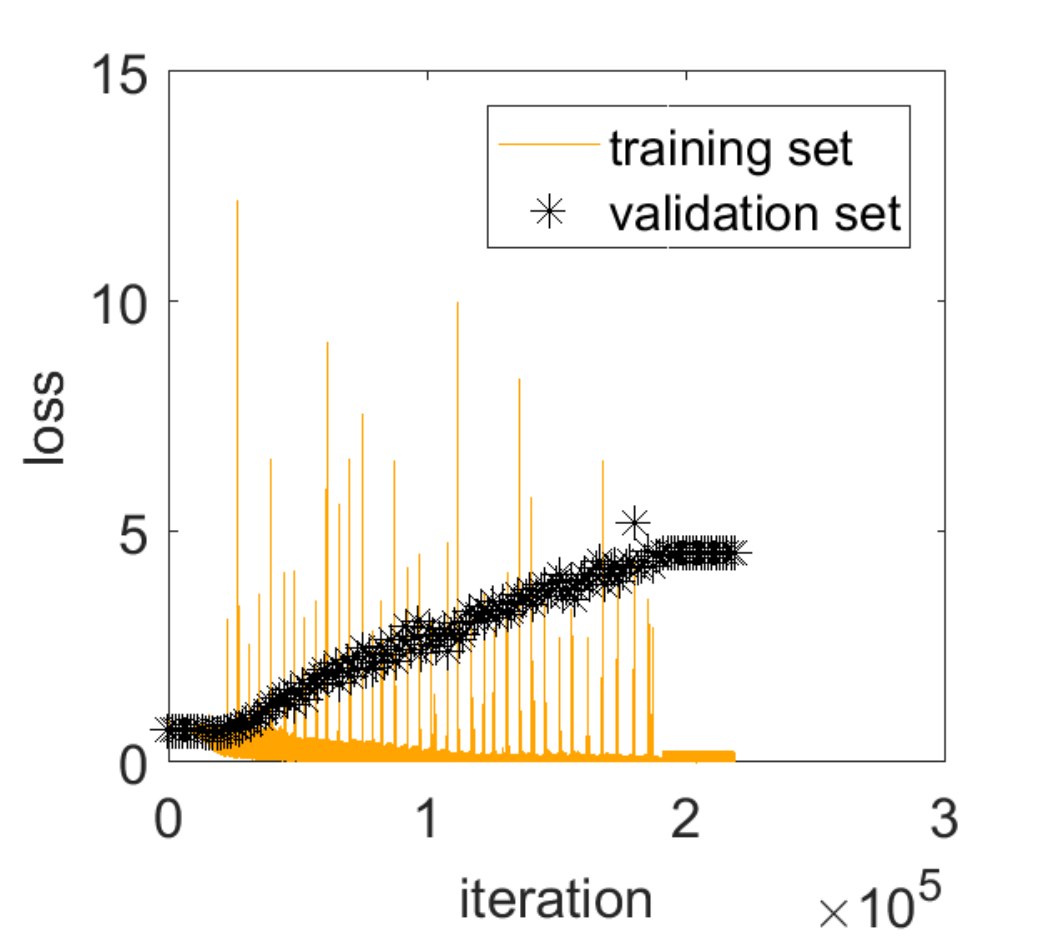}}
\subfloat[SNR=10 dB\label{fig:acc_sin_det_10_dB_33_1}]{\includegraphics[scale=0.5]{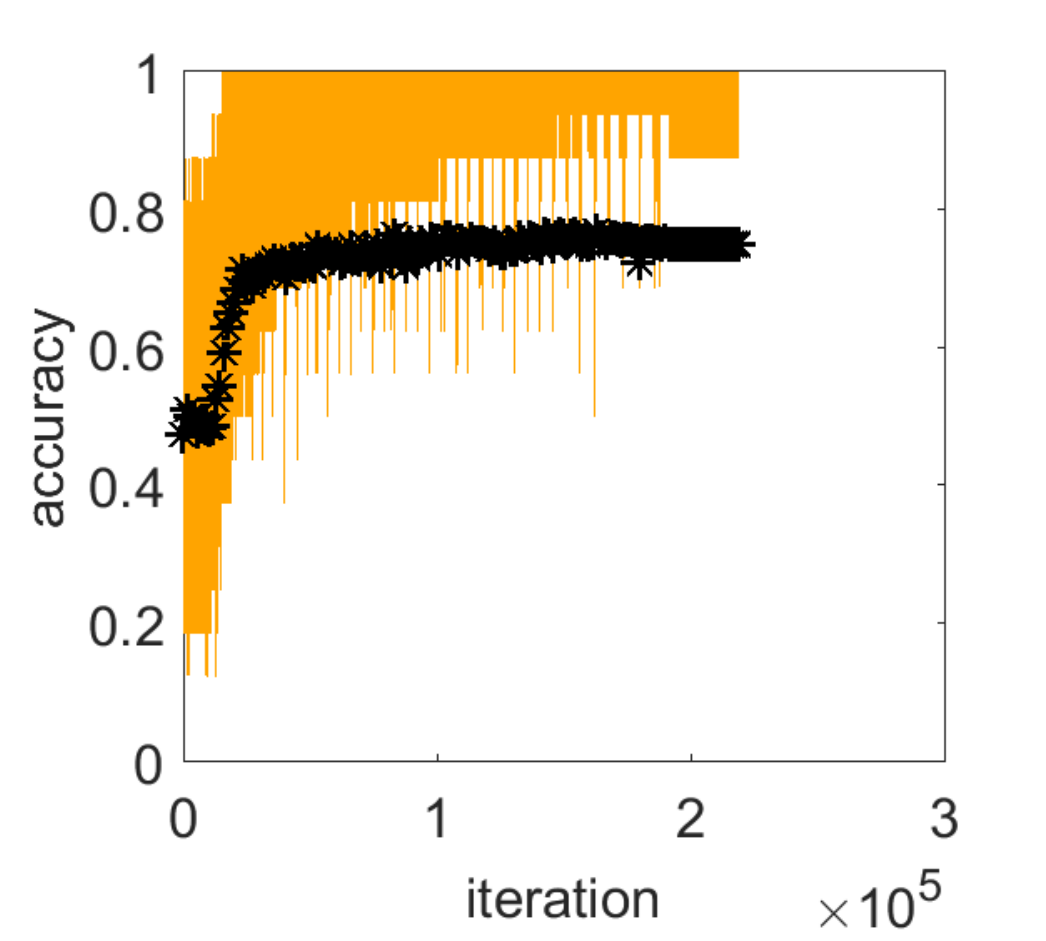}} \vspace{-0.1cm} 
\caption{Damage detection, case 1. Training and validation of the two branches convolutional architecture: evolution of the loss  $J_d \left(\boldsymbol{Y}, \boldsymbol{p} \right)$ on $\mathbb{D}^d_{train}$ and $\mathbb{D}^d_{val}$ (left column), and of $\mathcal{G}_d$ accuracy $A_d$ (right column) on $\mathbb{D}^d_{train}$ and on $\mathbb{D}^d_{val}$, both for the SNR$=15$ dB case (top row) and for the SNR$=10$ dB case (bottom row). \label{fig:sin_det}} \vspace{-0.1cm} 
\end{figure}

Fig.~\ref{fig:sin_det} reports the evolution of the training and validation loss for the dataset with SNR$=15$ dB and SNR$=10$ dB. The iteration number accounts for the number of times the NN weights 
are modified during the training process. The depicted training and validation loss functions refer to the case in which a two branches convolutional architecture has been employed to detect damage. The several spikes observed both in the loss and accuracy graphs are due to the stochastic nature of the training algorithm. During the early stages of the training, the NN displays the most significative gains in terms of classification accuracy, while further increasing the number of iterations only yields a limited effect on the generalization capabilities of the NN. Due to the lack of improvements, the early-stopping criterion has finally stopped the training.

Moving to the damage localization task, Tab.~\ref{tab:dam_id_sin_load_tab} collects the results related to the outcomes of $\mathcal{G}_l$ on $\mathbb{D}^l_{test}$ obtained for two different noise levels.

\begin{table}[h!]
\centering
\begin{tabular}{*3c}
\hline
\parbox{2 cm}{SNR (dB)} & \parbox{3cm}{\centering $\lbrace\mathcal{F}_{*} \rbrace$} & $A_{l}$ \\
\hline
$15$ & $\{\boldsymbol{u}^{sh}_i\}_{i=1}^8$ & $0.768$ \\
$15$ & $\{\boldsymbol{u}^{ax}_i\}_{i=1}^8$ & $0.769$ \\
$15$ & $\{\boldsymbol{u}^{sh}_i\}_{i=1}^8$ and $\{\boldsymbol{u}^{ax}_i\}_{i=1}^8$ & $0.812$ \\
$10$ & $\{\boldsymbol{u}^{sh}_i\}_{i=1}^8$ & $0.654$ \\
$10$ & $\{\boldsymbol{u}^{ax}_i\}_{i=1}^8 $ & $0.642$ \\
$10$ & $\{\boldsymbol{u}^{sh}_i\}_{i=1}^8$ and $\{\boldsymbol{u}^{ax}_i\}_{i=1}^8$ & $0.707$ \\
\bottomrule
\end{tabular}
\caption{Damage localization, case 1. Accuracy $\mathbb{A}_{l}$ of the classifier $\mathcal{G}_l$  evaluated on $\mathbb{D}^l_{test}$. \label{tab:dam_id_sin_load_tab}} \vspace{-0.25cm} 
\end{table}
The results show that the NN performances benefit from the employment of a two branches architectures: $A_{l}$ increases, compared to the best outcome of the single convolutional layer architecture, from $0.769$ to $0.812$ for the SNR$=15$ dB case, and from $0.654$ to $0.707$ for the SNR$=10$ dB case. This means that the NN has succeeded in performing a data fusion of the extracted information for the sake of classification.

The values of $A_{d}$ and $A_{l}$ are quite close, despite of the greater complexity of the damage localization problem; this might be due to the intrinsic capability of the FCN to detect correlations between different sensor recordings, allowing us to perform a correct damage localization. 

\begin{figure}[h!]
\captionsetup[subfigure]{justification=centering}\centering
\subfloat[SNR=15 dB\label{fig:loss_sin_id_15_dB_49_1}]{\includegraphics[scale=0.5]{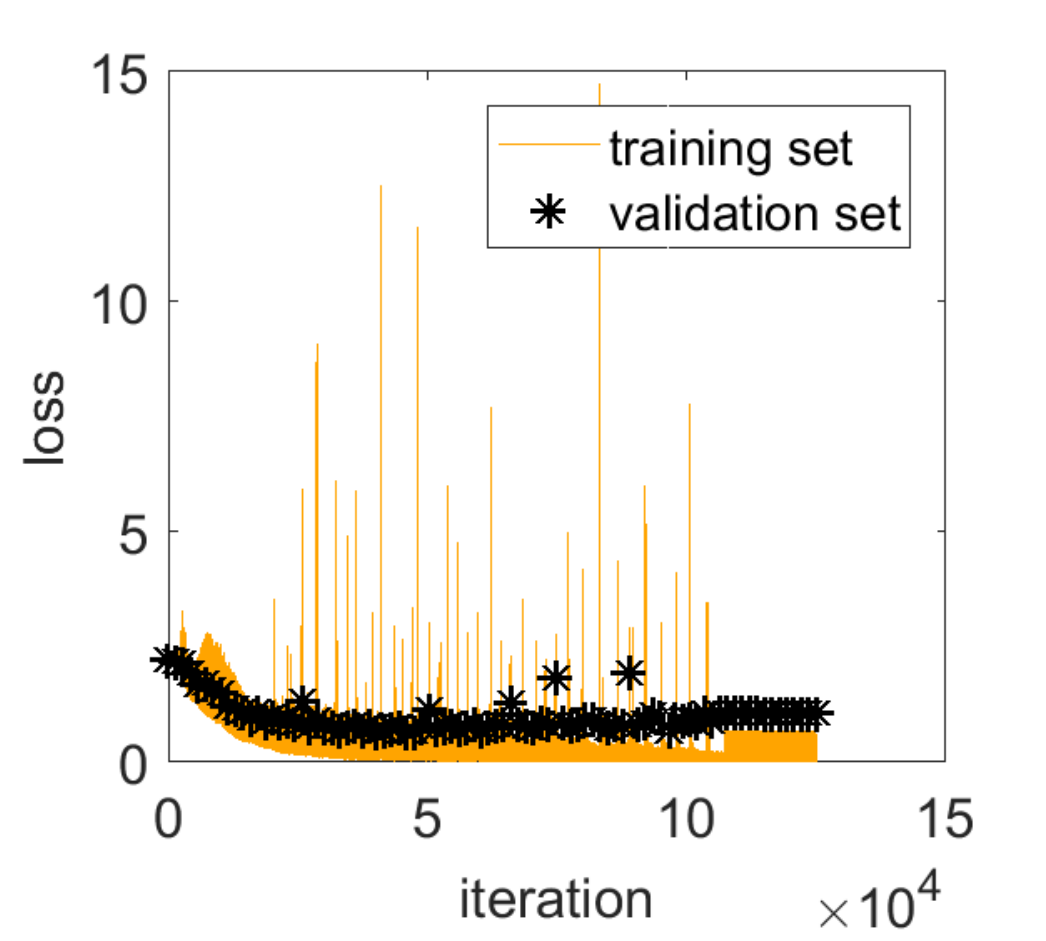}}
\subfloat[SNR=15 dB\label{fig:acc_sin_id_15_dB_49_1}]{\includegraphics[scale=0.5]{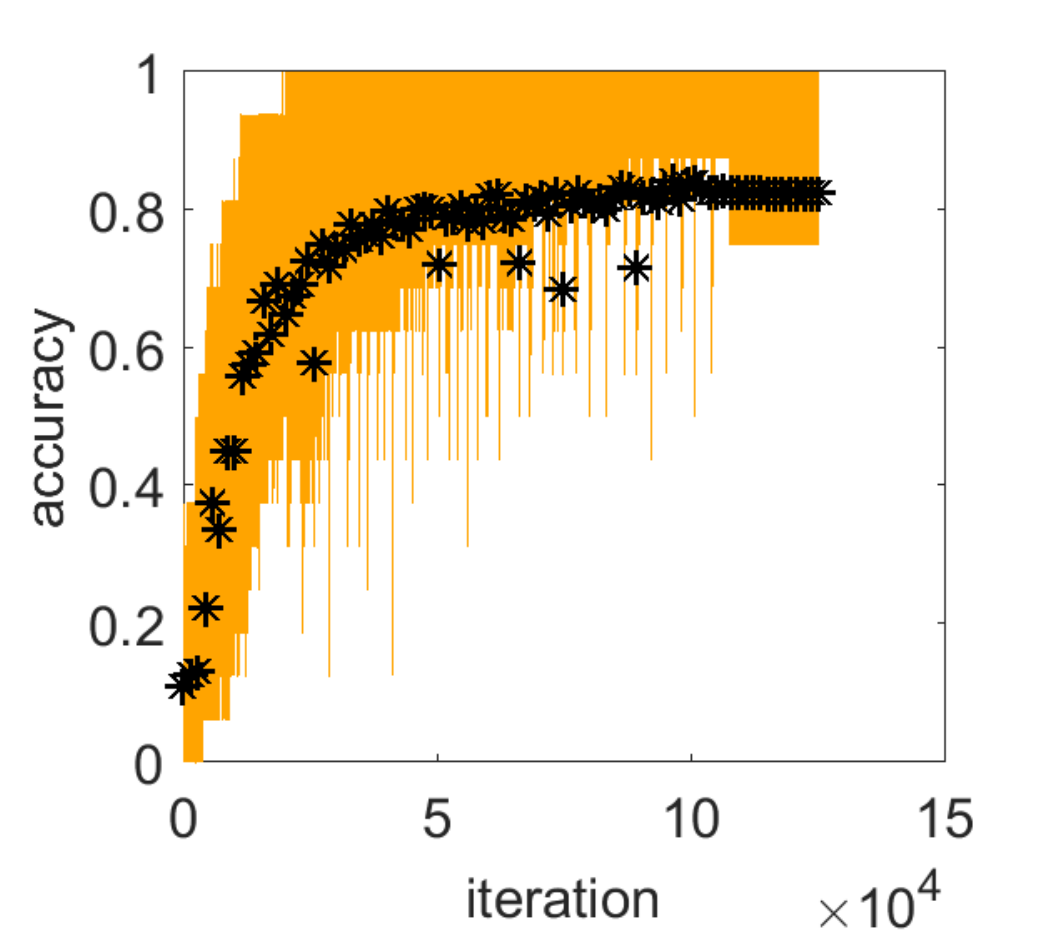}}  \vspace{-0.25cm} \\
\subfloat[SNR=10 dB\label{fig:loss_sin_id_10_dB_50_1}]{\includegraphics[scale=0.5]{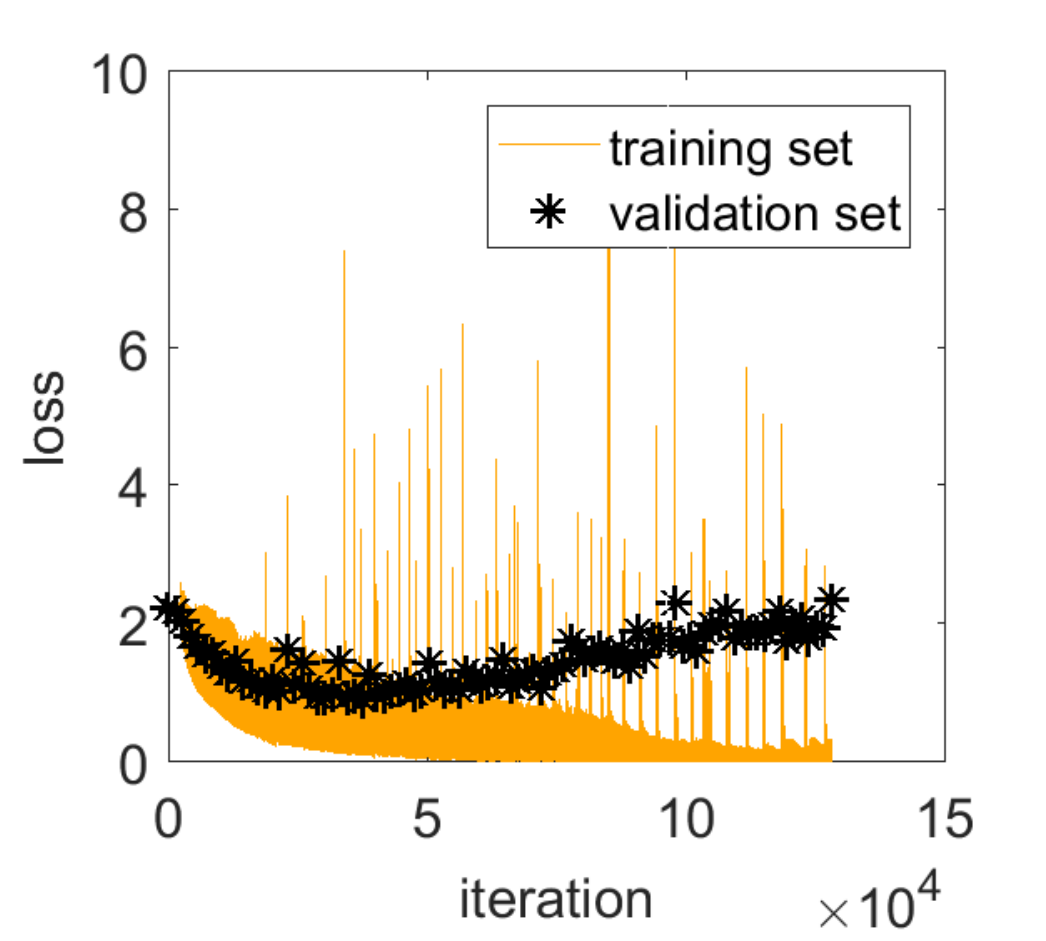}}
\subfloat[SNR=10 dB\label{fig:acc_sin_id_10_dB_50_1}]{\includegraphics[scale=0.5]{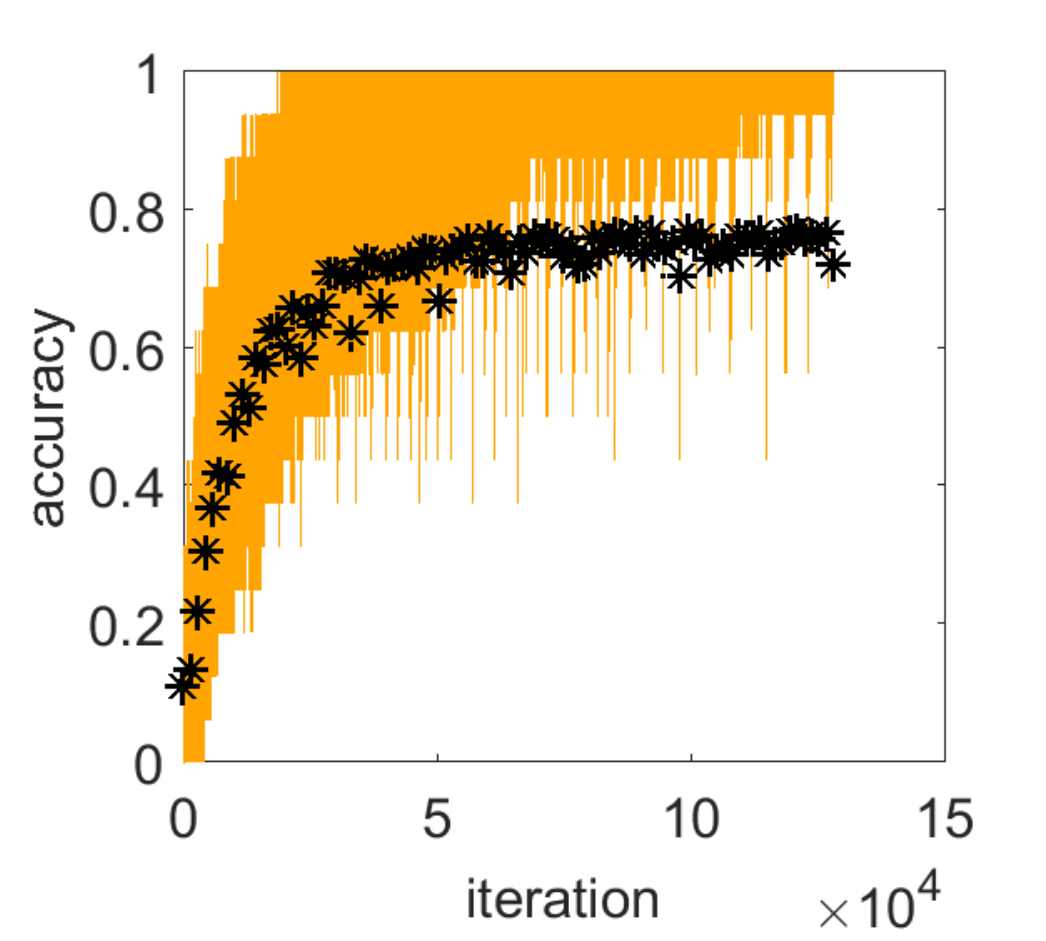}} \vspace{-0.1cm} 
\caption{Damage localization, case 1. Training and validation of the two branches convolutional architecture: evolution of the  loss $J_l \left(\boldsymbol{Y}, \boldsymbol{p} \right)$ on $\mathbb{D}^l_{train}$ and on $\mathbb{D}^l_{val}$ (left column), and of $\mathcal{G}_l$ accuracy $A_l$ (right column) on $\mathbb{D}^l_{train}$ and on $\mathbb{D}^l_{val}$, both for the SNR$=15$ dB case (top row) and for the SNR$=10$ dB case (bottom row).\label{fig:sin_id}}\vspace{-0.25cm} 
\end{figure}

Fig.~\ref{fig:sin_id} reports the evolution of the training and validation loss functions on $\mathbb{D}^l_{val}$ and $\mathbb{D}^l_{test}$ for the datasets with SNR$=15$ dB and SNR$=10$ dB, in the case where a two branches convolutional architecture has been employed. Compared with Fig.~\ref{fig:sin_det}, a smaller difference in terms of loss and accuracy can be highlighted. This is due to the greater complexity of the damage localization task, that requires to exploit the computational resources of the NN entirely. Indeed, the same number of filters $N_1$, $N_2$ and $N_3$ has been used for both the classification tasks, despite of their different complexity. On the other hand, we expect that $A_d$ on $\mathbb{D}^d_{test}$, reported in Tab.~\ref{tab:dam_det_sin_load_tab}, would not be affected by reducing the number of filters. This conclusion can be reached by looking at Fig.~\ref{fig:sin_det} and observing that, during the last stages of the training, $A_d$ on $\mathbb{D}^d_{train}$ is shown to be always greater than the one obtained on $\mathbb{D}^d_{val}$.

\begin{figure}[t!]
\vspace{-0.25cm} 
\captionsetup[subfigure]{justification=centering}
\centering
\subfloat[SNR = 15 dB\label{fig:conf_sin_id_15_dB_49_1}]{\includegraphics[scale=0.4]{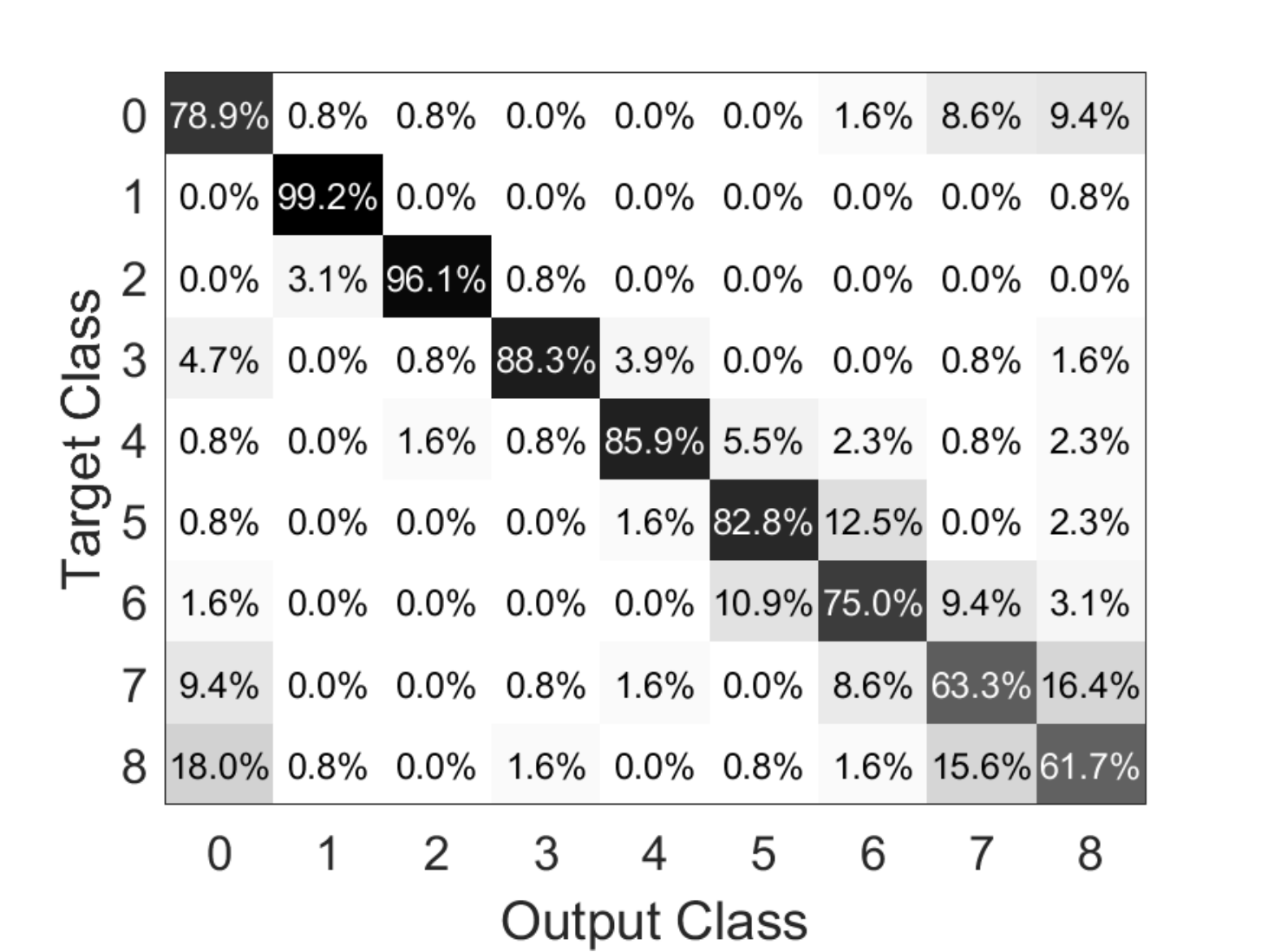}}
\subfloat[SNR = 10 dB\label{fig:conf_sin_id_10_dB_50_1}]{\includegraphics[scale=0.4]{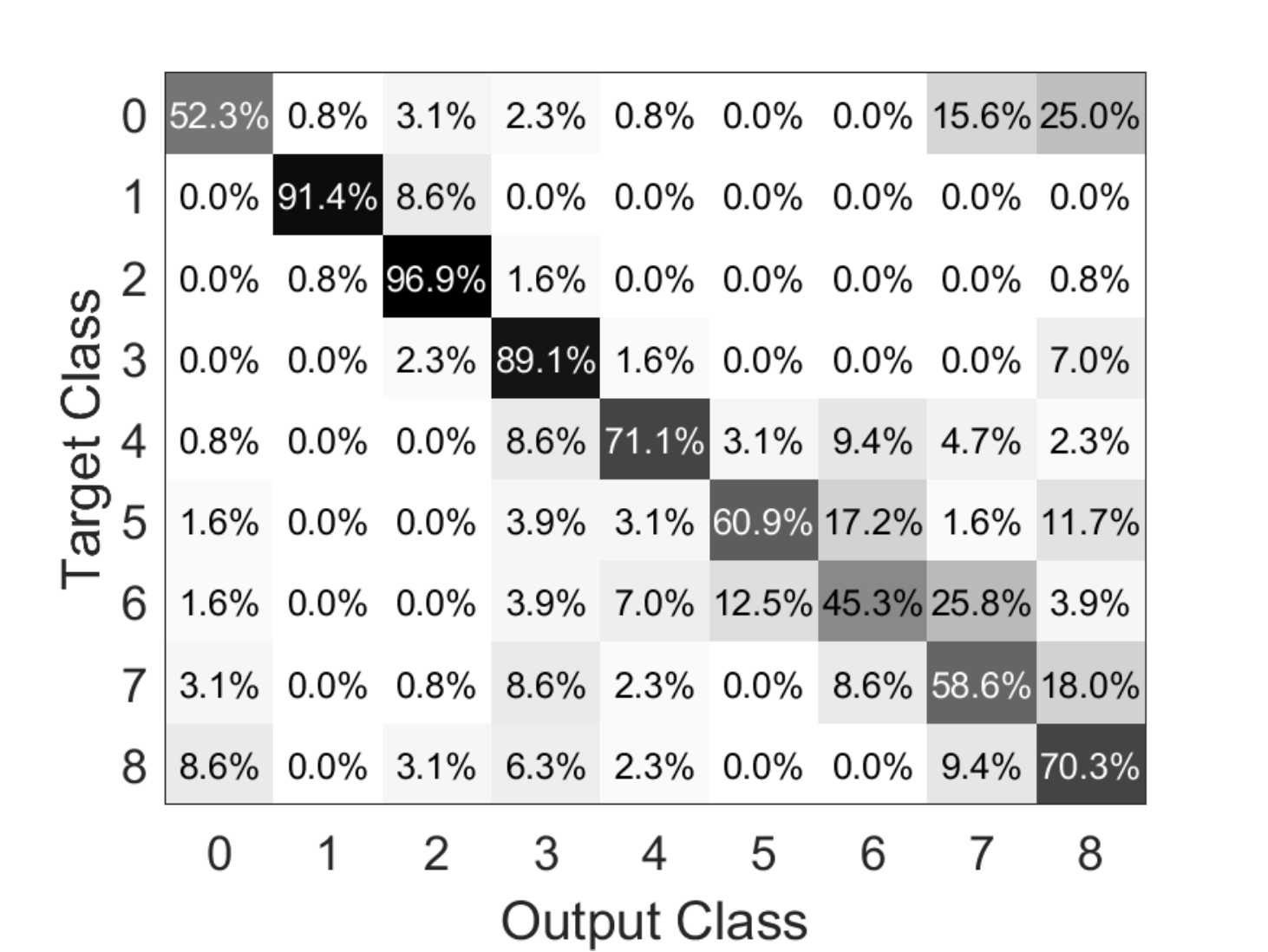}} \vspace{-0.2cm} 
\caption{ Damage localization, case 1. Confusion matrices, case 1, $15$ dB (left picture) and $10$ dB (right picture) SNR datasets.\label{fig:conf_sin_id}} \vspace{-0.1cm} 
\end{figure}

In Fig.~\ref{fig:conf_sin_id} the confusion matrices related to the two datasets (SNR$=15$ dB and SNR$=10$ dB) are reported.
Most of the errors concern the classification of the damage scenarios in which the inter-story stiffness of the highest floors has been reduced, as shown by the entries of the $7$-th and $8$-th rows and columns of the matrices. This outcome is not surprising if we consider that these damage scenarios only induce small variations in the shear frequencies. Moreover, by looking at Figs.~\ref{fig:damaged_signal_flex} and \ref{fig:damaged_signal_axial}, we can remark that the time evolution of the structural motions under these damage scenarios cannot be easily distinguished from the undamaged case.

\subsubsection{Damage detection and localization in case 2}
We now consider the outcomes of the trained classifiers in the case where a random disturbance is applied to the structural system. Regarding the damage detection task, with this type of excitation the NN is able to distinguish between undamaged and damaged instances almost perfectly (see Tab.~\ref{tab:dam_det_noise_load_tab}). Indeed, $A_d=0.999$ and $A_d=0.998$ have been reached by the two convolutional branches architecture when $f_{min}=15$ and $f_{max}=17$ Hz, or $f_{min}=5$ and $f_{max}=7$ Hz, have been selected as frequency ranges for the applied lateral and vertical forces.

\begin{table}[h!]
\centering
\begin{tabular}{*3c}
\hline
\parbox{2cm}{\centering $f_{min}$ - $f_{max}$ (Hz)} & \parbox{3cm}{\centering $\lbrace\mathcal{F}_{*} \rbrace$} & $A^d$ \\
\hline
$15-17$ & $\{\boldsymbol{u}^{sh}_i\}_{i=1}^8$ & $0.998$ \\
$15-17$ & $\{\boldsymbol{u}^{ax}_i\}_{i=1}^8$ & $0.997$ \\
$15-17$ & $\{\boldsymbol{u}^{sh}_i\}_{i=1}^8$ and $\{\boldsymbol{u}^{ax}_i\}_{i=1}^8$ & $0.999$ \\
$5-7$ & $\{\boldsymbol{u}^{sh}_i\}_{i=1}^8$ & $0.996$ \\
$5-7$ & $\{\boldsymbol{u}^{ax}_i\}_{i=1}^8$ & $0.892$ \\
$5-7$ & $\{\boldsymbol{u}^{sh}_i\}_{i=1}^8$ and $\{\boldsymbol{u}^{ax}_i\}_{i=1}^8$ & $0.998$ \\
\bottomrule
\end{tabular}
\caption{Damage detection, case 2. Accuracy $\mathbb{A}_{d}$ of the classifier $\mathcal{G}_d$  evaluated on $\mathbb{D}^d_{test}$. \label{tab:dam_det_noise_load_tab}}
\end{table}

We next consider the NN outcomes for the damage localization task. With this type of excitation, the NN is able to accomplish an extremely accurate classification of the damaged scenarios, reaching $A_{l}=0.986$ and $A_{l}=0.993$ when $f_{min}=15$ and $f_{max}=17$ Hz or $f_{min}=5$ and $f_{max}=7$ Hz have been used, respectively. In the former case, the best classification performances have been obtained by the two convolutional branches architecture, as shown in Tab.~\ref{tab:dam_id_noise_load_tab}. For the latter case, the NN employing as input $\mathcal{F}_{*} = \{{u}^{sh}_i\}_{i=1}^8$ provides the best classification result. The better performances of the NN employing $\mathcal{F}_{*} = \{{u}^{sh}_i\}_{i=1}^8$ rather than $\mathcal{F}_{*} = \{{u}^{ax}_i\}_{i=1}^8$ is likely due to the fact in this latter case no axial frequencies have been excited by the applied load, as remarked in \nameref{sec:load_case_2}. However, this fact also shows that the data fusion operated by the two convolutional branches architecture has been only partially able to select the most important information required for the damage localization task. Nevertheless, very good results have been reached by also employing $\mathcal{F}_{*} = \{{u}^{ax}_i\}_{i=1}^8$ (see Tab.~\ref{tab:dam_id_noise_load_tab}).

\begin{figure}[h!]
\captionsetup[subfigure]{justification=centering}
\centering
\vspace{-0.2 cm}
\subfloat[$N_0=1$; $i=8$\label{fig:conf_noise_id_5_7_42_5_flex_1_ch}]{\includegraphics[scale=0.4]{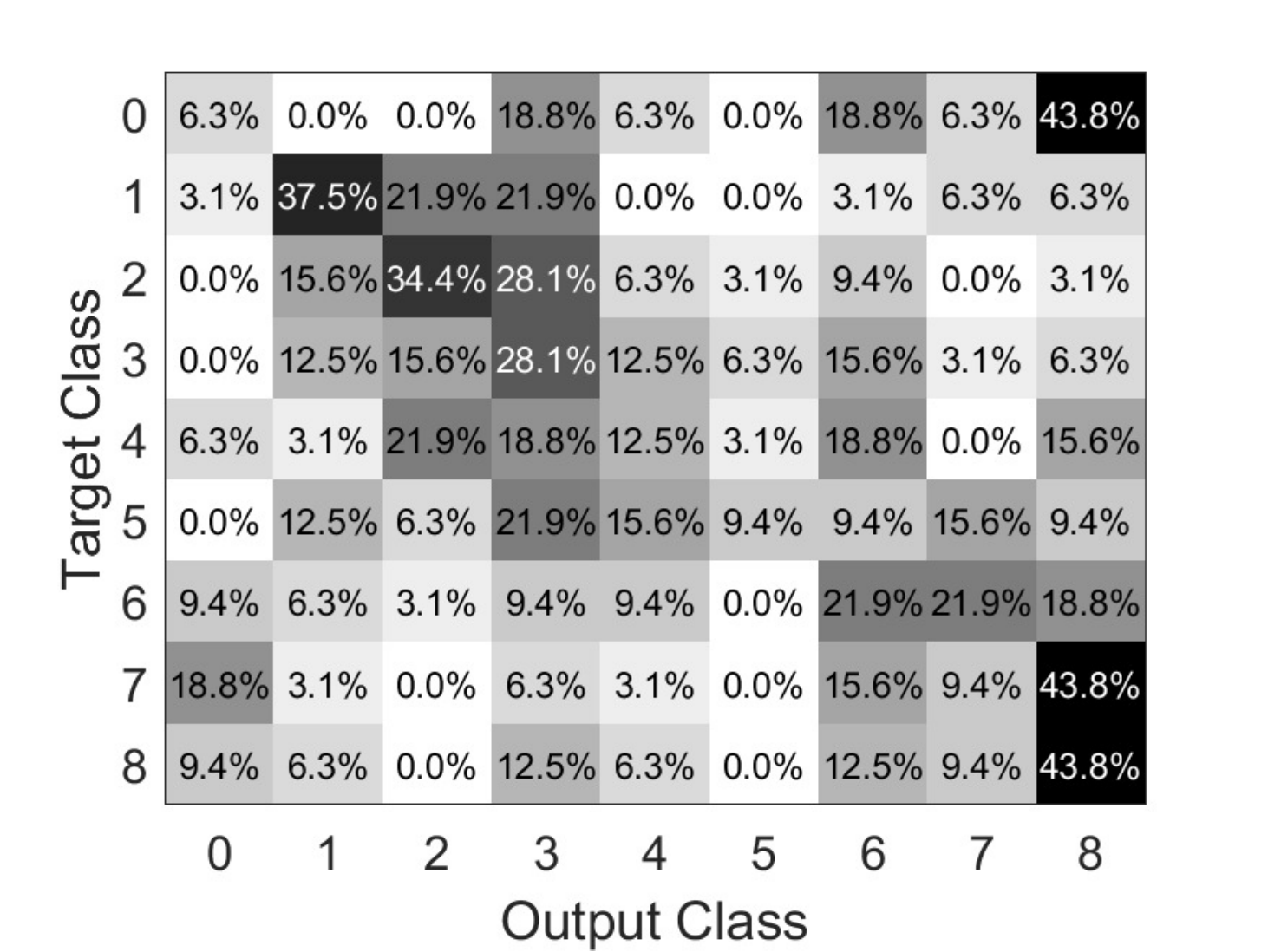}}
\subfloat[$N_0=2$; $i=$($4,8$)\label{fig:conf_noise_id_5_7_42_6_flex_2_ch}]{\includegraphics[scale=0.4]{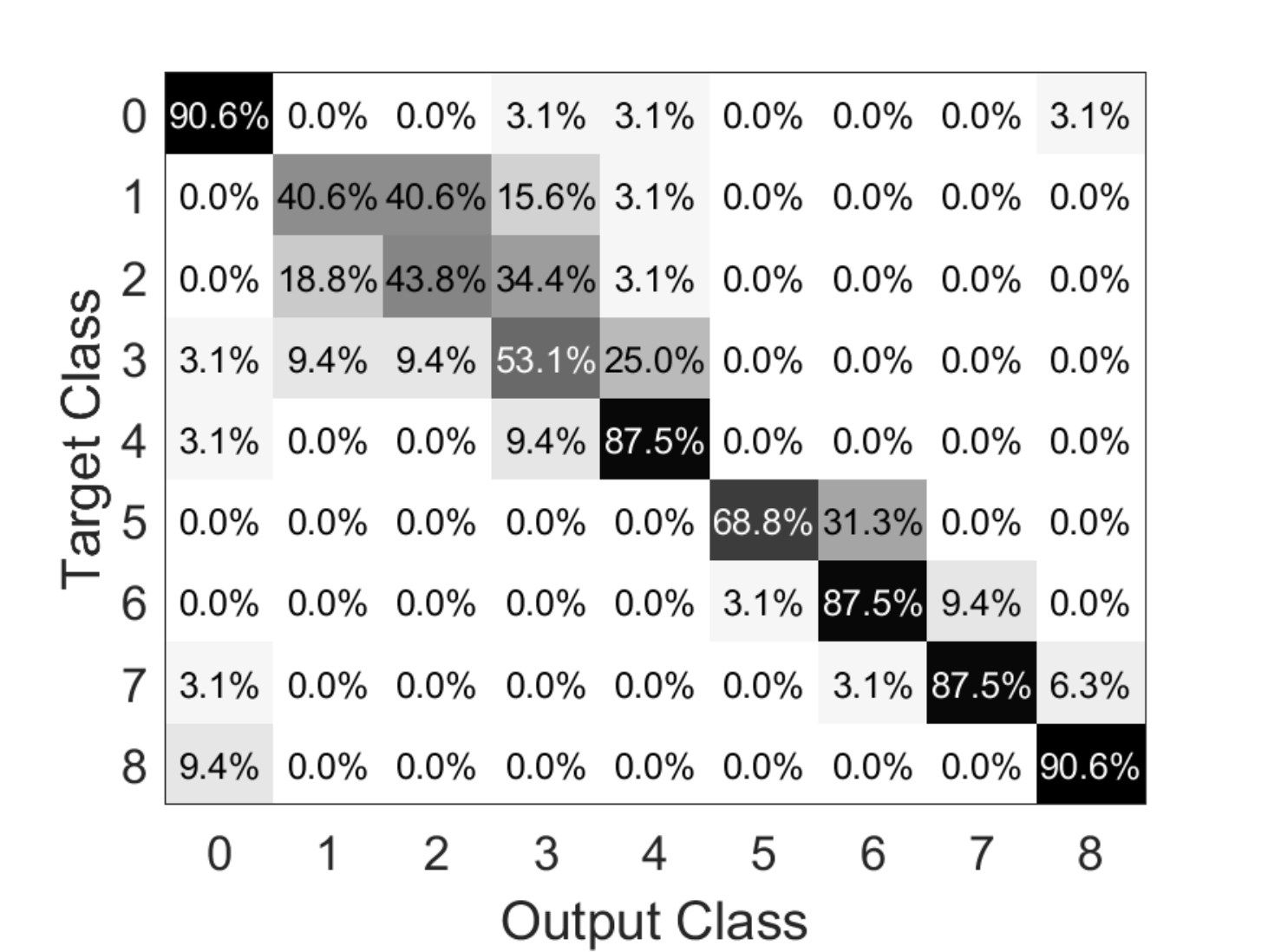}} \\
\vspace{-0.4cm}
\subfloat[$N_0=3$; $i=$($2,4,8$)\label{fig:conf_noise_id_5_7_42_7_flex_3_ch}]{\includegraphics[scale=0.4]{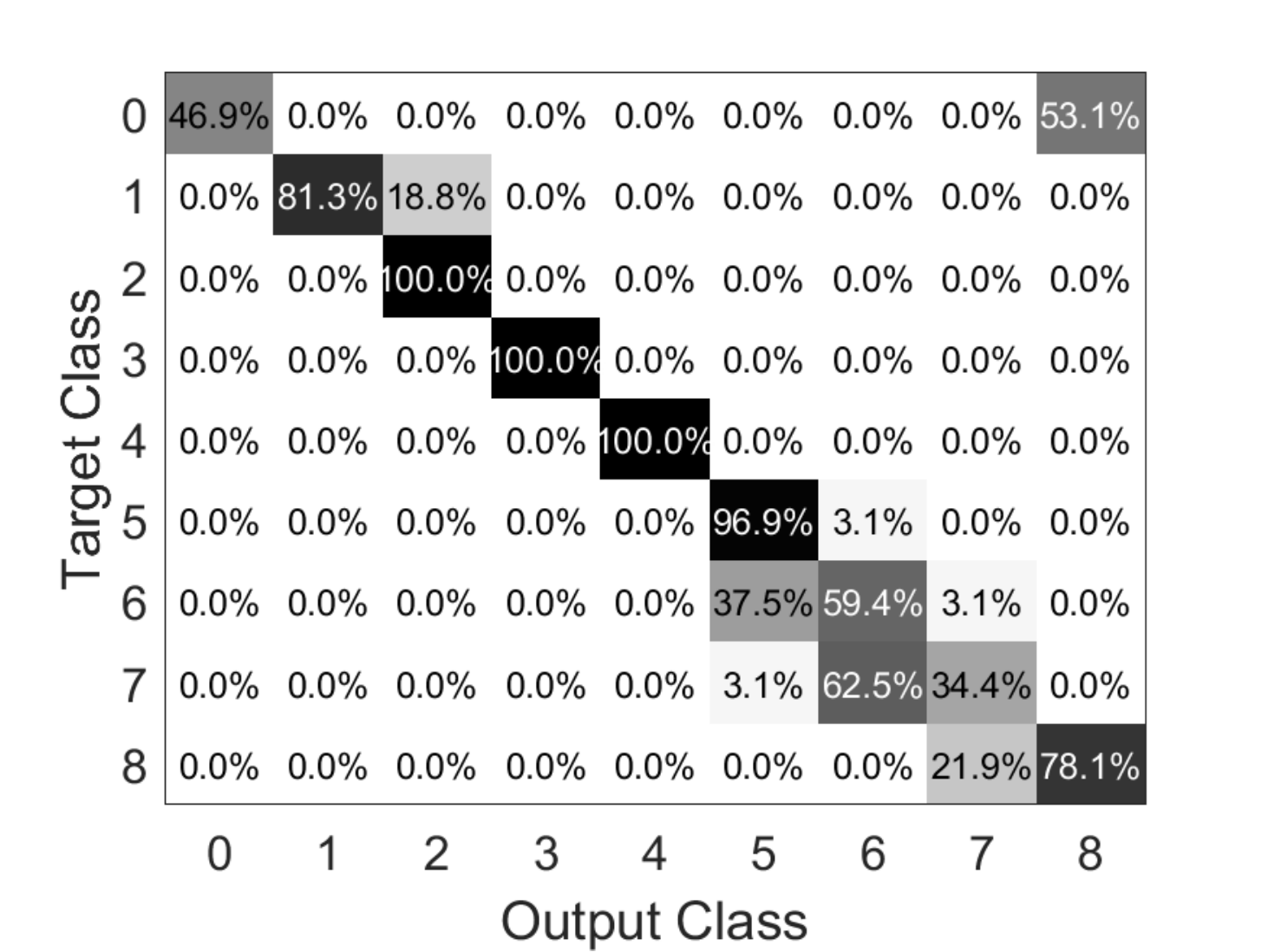}}
\subfloat[$N_0=4$; $i=$($2,4,6,8$)\label{fig:conf_noise_id_5_7_42_4_flex_4_ch}]{\includegraphics[scale=0.4]{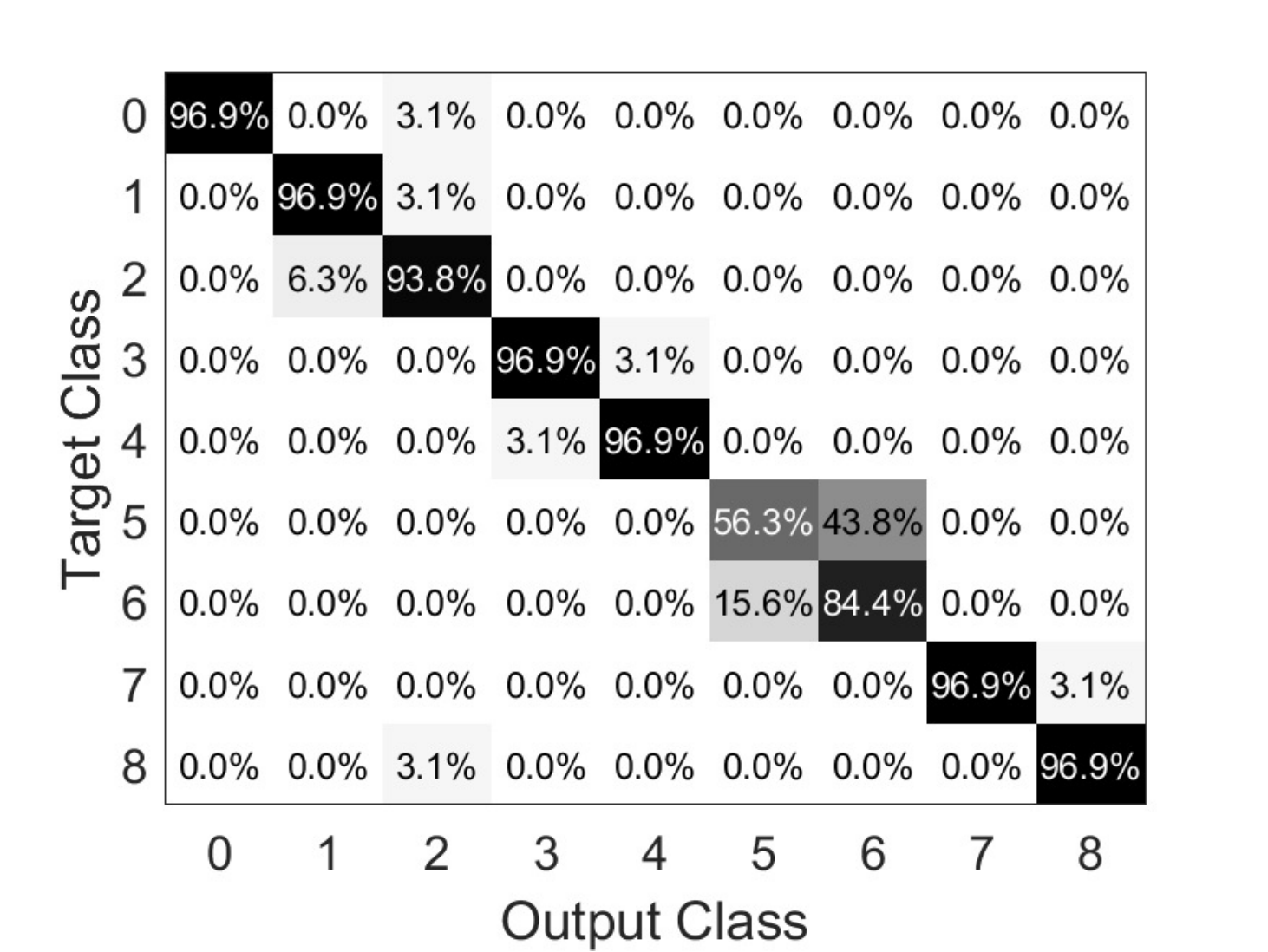}} \\
\vspace{-0.4cm}
\subfloat[$N_0=5$; $i=$($2,4,6,7,8$)\label{fig:conf_noise_id_5_7_42_8_flex_5_ch}]{\includegraphics[scale=0.4]{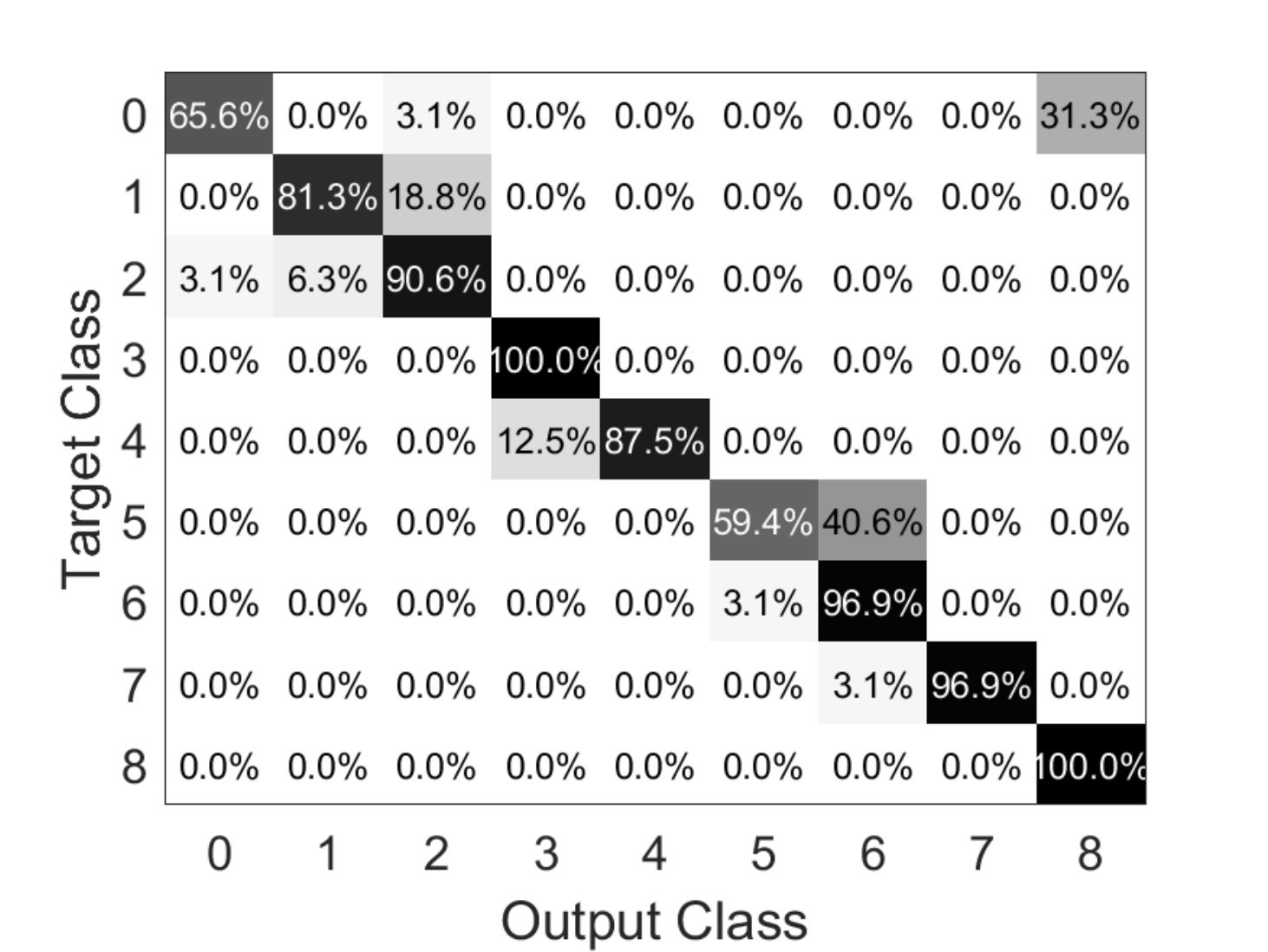}}
\subfloat[$N_0=6$; $i=$($2,4,5,6,7,8$)\label{fig:conf_noise_id_5_7_42_9_flex_6_ch}]{\includegraphics[scale=0.4]{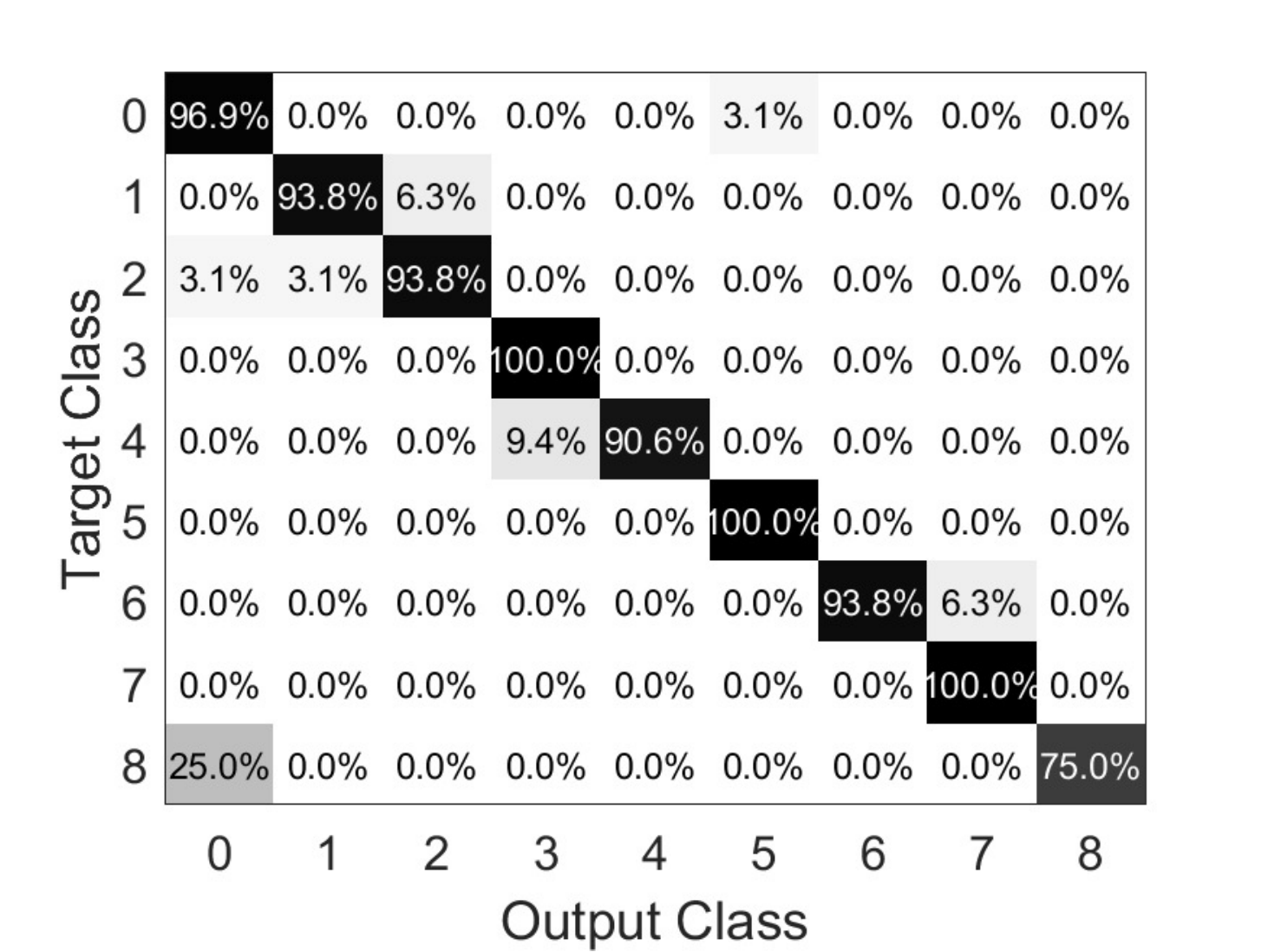}} \\
\vspace{-0.4cm}
\subfloat[$N_0=7$; $i=$($2,3,4,5,6,7,8$)\label{fig:conf_noise_id_5_7_42_10_flex_7_ch}]{\includegraphics[scale=0.4]{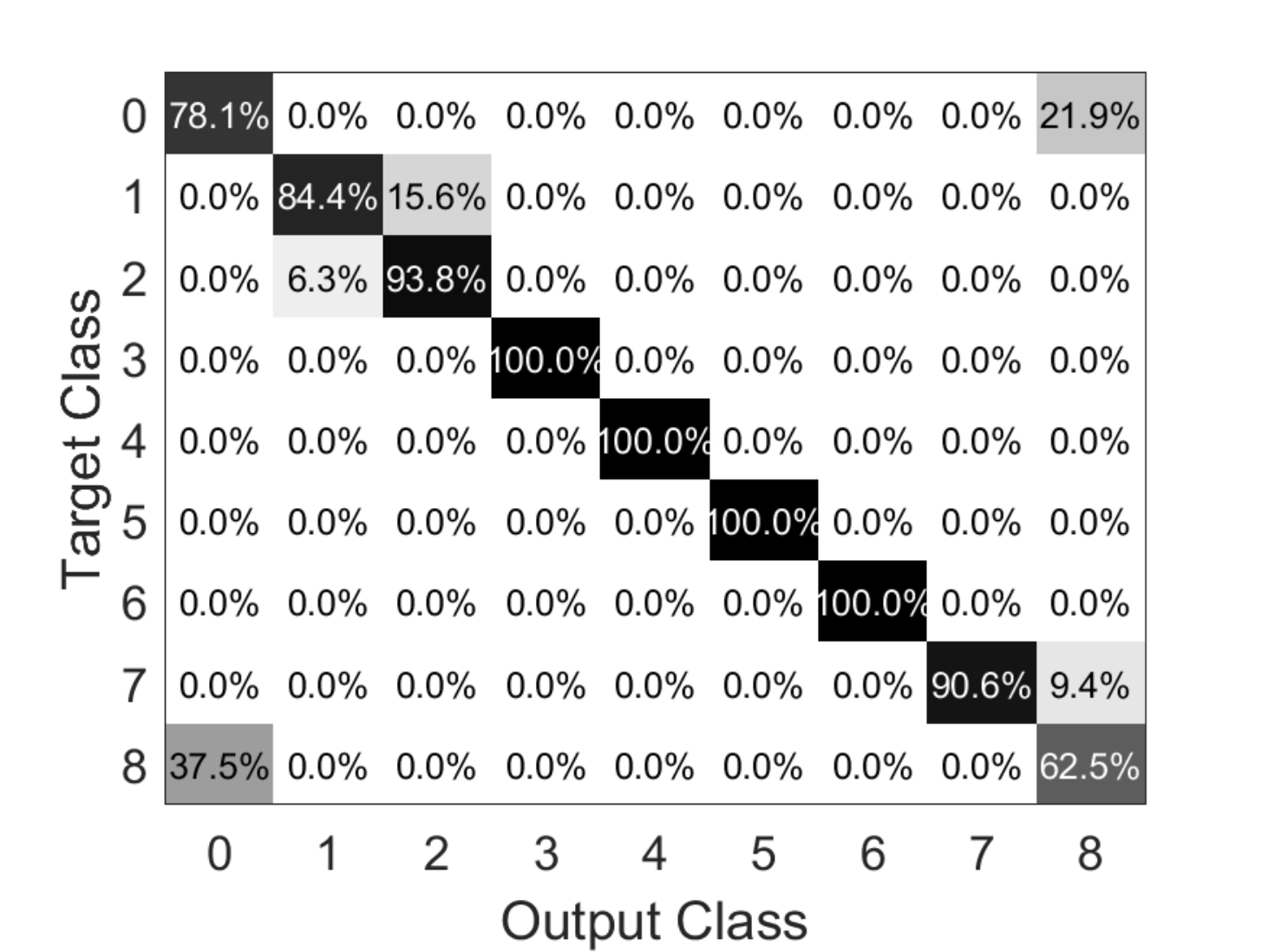}}
\subfloat[$N_0=8$; $i=$($1,2,3,4,5,6,7,8$)\label{fig:conf_noise_id_5_7_42_3_flex}]{\includegraphics[scale=0.4]{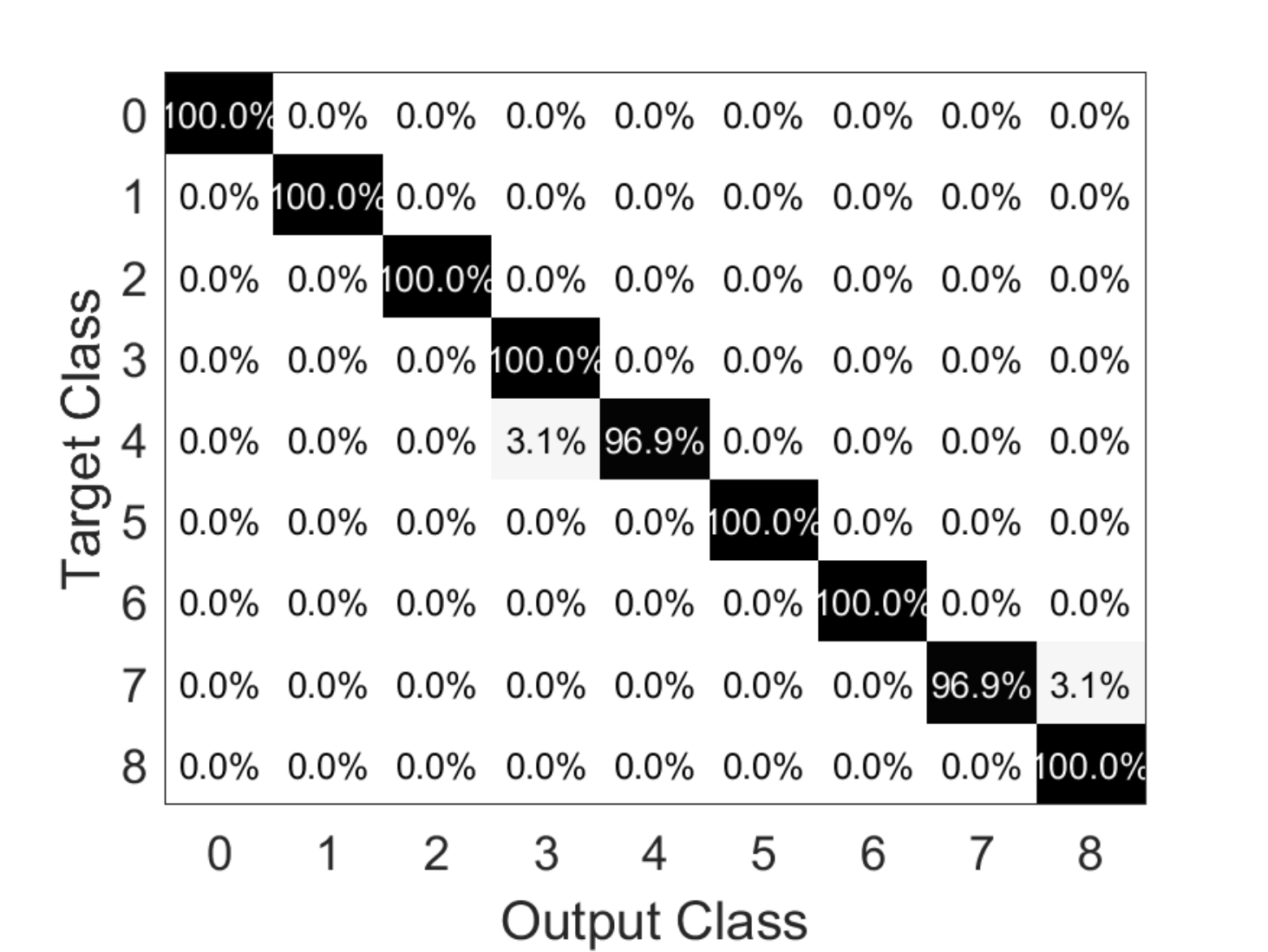}} \\
\caption{$\mathcal{G}_{l}$ confusion matrices for different number $N_0$ of input channels $\mathcal{F}^{i}_{*}$. Case 2, $f_{min}=5$ and $f_{max}=7$ Hz.\label{fig:conf_noise_id}}
\end{figure}

\begin{table}[h!]
\centering
\vspace{-0.3cm} 
\begin{tabular}{*3c}
\hline
\parbox{2cm}{\centering $f_{min}$ - $f_{max}$ (Hz)} & \parbox{3cm}{\centering $\lbrace\mathcal{F}_{*} \rbrace$} & $A^{l}$ \\
\hline
$15-17$ & $\{\boldsymbol{u}^{sh}_i\}_{i=1}^8$ & $0.972$ \\
$15-17$ & $\{\boldsymbol{u}^{ax}_i\}_{i=1}^8$ & $0.972$ \\
$15-17$ & $\{\boldsymbol{u}^{sh}_i\}_{i=1}^8$ and $\{\boldsymbol{u}^{ax}_i\}_{i=1}^8$ & $0.986$ \\
$5-7$ & $\{\boldsymbol{u}^{sh}_i\}_{i=1}^8$ & $0.993$ \\
$5-7$ & $\{\boldsymbol{u}^{ax}_i\}_{i=1}^8$ & $0.892$ \\
$5-7$ & $\{\boldsymbol{u}^{sh}_i\}_{i=1}^8$ and $\{\boldsymbol{u}^{ax}_i\}_{i=1}^8$ & $0.972$ \\
\bottomrule
\end{tabular}
\vspace{-0.1cm}
\caption{Damage localization, case 2. Accuracy $\mathbb{A}_{l}$ of the classifier $\mathcal{G}_l$  evaluated on $\mathbb{D}^l_{test}$.  \label{tab:dam_id_noise_load_tab}}
\end{table}

We highlight the effect of each incoming signal 
on the classification outcomes (see Tab.~\ref{tab:dam_id_noise_load_ch_tab}), since the accuracy $A^{l}$ on $\mathbb{D}^l_{test}$ changes for different numbers of input signals $N_0$. The results refer to the case in which only some of the displacements ${u}^{sh}_i$, $i=1,\ldots,8$ have been considered, and $f_{min}=5$ and $f_{max}=7$ Hz. The corresponding confusion matrices are sketched in Fig.~\ref{fig:conf_noise_id}, showing that the classification error related to a damage scenario $g$ is reduced when the corresponding ${u}^{sh}_g$, that is the signal acquired on the floor whose inter-story stiffness has been reduced, is used as input for the NN.

\begin{table}[h!]
\centering
\begin{tabular}{*3c}
\hline
\parbox{1cm}{\centering $N_0$} & \parbox{3cm}{\centering $\lbrace\mathcal{F}_{*} \rbrace$} & $A_l$ \\
\hline
$1$ & $i=8$ & $0.226$ \\
$2$ & $i=$($4,8$) & $0.722$ \\
$3$ & $i=$($2,4,8$) & $0.774$ \\
$4$ & $i=$($2,4,6,8$) & $0.906$ \\
$5$ & $i=$($2,4,6,7,8$) & $0.865$ \\
$6$ & $i=$($2,4,5,6,7,8$) & $0.937$ \\
$7$ & $i=$($2,3,4,5,6,7,8$) & $0.899$ \\
$8$ & $i=$($1,2,3,4,5,6,7,8$) & $0.993$ \\
\bottomrule
\end{tabular}
\vspace{-0.1cm}
\caption{Damage localization, case 2. Accuracy $\mathbb{A}_{l}$ of the classifier $\mathcal{G}_l$  evaluated on $\mathbb{D}^l_{test}$.  Different numbers $N_0$ of input channels $\mathcal{F}_{*}$, related to ${u}^{sh}_i$, are employed. Here, $f_{min}=5$ and $f_{min}=7$ Hz.\label{tab:dam_id_noise_load_ch_tab}}
\vspace{-0.1cm}
\end{table}

\section{Conclusions}
\label{sec:conclusions}

In this paper we have investigated a new strategy for real-time structural health monitoring, treating damage detection and localization as classification tasks \cite{art:Farrar_1}, and framing the proposed procedure in the family of SBC approaches \cite{art:Taddei_Patera}. For the first time in this field, we have proposed to employ fully convolutional networks to analyse time series coming from a set of sensors. Fully convolutional networks architectures differing for the number of convolutional branches have been exploited to deal with datasets including time signals of different length and sampling rate. Convolutional layers have been shown to enable the automatic extraction of features to be used for the classification task at hand. The neural network architecture has been trained in a supervised manner on data generated through the numerical solution of a physics-based model of the monitored structure under different damage scenarios.

In the considered numerical benchmarks, we have obtained extremely good performances concerning both damage detection and damage localization, even in presence of noise, when   the applied loads   can be characterized either $(i)$ in terms of a few (a priori, random) frequencies, or $(ii)$ by a higher number of frequencies, within a given range. Especially in the second case, the outcomes of the NN classifier have shown the potentialities of the proposed procedure in view of the application to real-life cases.

In future works we aim to employ the proposed architecture to deal with data coming from real monitoring systems, tackling the main limit of the proposed procedure concerning the adherence of the simulated dataset to the real structural response. This is a well-known problem in the machine learning community \cite{art:domain_adaptation}. By coupling recurrent layers branches to the proposed convolutional ones, we expect to further increase the NN performances. As further steps, we will try to exploit model order reduction techniques for the dataset construction, extending the proposed methodology to more complex structural configurations and damage scenarios, and to design the set of sensors according to a Bayesian optimization technique \cite{art:Capellari_1,proc:Capellari_1,art:Capellari_3}.



%
%

\section*{Acknowledgments}
The authors thank Andrea Opreni (Politecnico di Milano) for fruitful discussions about DL architectures. LR, SM and AC gratefully acknowledge the financial support from MIUR Project PRIN 15-2015LYYXA 8 ``Multi-scale mechanical models for the design and optimization of microstructured smart materials and metamaterials".
\bibliography{biblio}

\end{document}